\documentclass[journal]{IEEEtran}
\usepackage{multirow}
\usepackage[caption=false]{subfig}

\usepackage{algorithmic}
\usepackage{algorithm}
\usepackage{babel}

\usepackage{comment}
\usepackage{url}

\usepackage{ifpdf}
 \ifpdf
 \else
 \fi

\ifCLASSINFOpdf
   \usepackage[pdftex]{graphicx}
   \graphicspath{{../eps/}{../jpg/}}
   \DeclareGraphicsExtensions{.eps,.jpg,.png}
\else
   \usepackage[dvips]{graphicx}
   \graphicspath{{../eps/}}
   \DeclareGraphicsExtensions{.eps}
\fi

\usepackage{amsmath}
\usepackage{amssymb}
\usepackage{xcolor}

\begin{document}

\title{\textcolor{black}{An Improved Discriminative Optimization \\for 3D Rigid Point Cloud Registration} }
\author{Jia~Wang, Ping~Wang, Biao~Li, Ruigang~Fu, and Junzheng~Wu
\thanks{Manuscript received XXX XX, 2021; revised XXX XX, 2021 and XXX XX, 2021; accepted XXX XX, 2021.
\textit{(Corresponding author: Ping Wang.)}}
\thanks{J. Wang, P. Wang, B. Li, R. Fu and J. Wu are with the Key Laboratory of ATR, College of Electronic Science and Technology, National University of Defense Technology, Changsha, 410073, China. (e-mail: wangj\_cs04@sina.com; 760521407@qq.com; libiao\_cn@163.com; ruigangfu@nudt.edu.cn; wujunzheng@nint.ac.cn).}
}

\markboth{Journal of \LaTeX\ Class Files,~Vol.~14, No.~8, November~2020}%
{Shell \MakeLowercase{\textit{et al.}}: Bare Demo of IEEEtran.cls for IEEE Journals}

\maketitle

\begin{abstract}
\textcolor{black}{The Discriminative Optimization (DO) algorithm has been proved much successful in 3D point cloud registration. In the original DO, the feature (descriptor) of two point cloud was defined as a histogram, and the element of histogram indicates the weights of scene points in "front" or "back" side of a model point. In this paper, we extended the histogram which indicate the sides from "front-back" to "front-back", "up-down", and "clockwise-anticlockwise". In addition, we reweighted the extended histogram according to the model points' distribution. We evaluated the proposed Improved DO on the Stanford Bunny and Oxford SensatUrban dataset, and compared it with six classical State-Of-The-Art point cloud registration algorithms. The experimental result demonstrates our algorithm achieves comparable performance in point registration accuracy and root-mean-sqart-error.}

\end{abstract}

\begin{IEEEkeywords}
\textcolor{black}{Discriminative Optimization, Linear Regression, Point Cloud Registration, Optimization, Supervised Sequential Mapping.}
\end{IEEEkeywords}

\IEEEpeerreviewmaketitle

\section{Introduction}
\IEEEPARstart{T}{he} mathematical optimization plays a key role in most of computer vision tasks, such as point cloud registration. An important step of optimization is to select a proper penalty function (loss function), which is used for the judgement of optimum. But sometimes it is difficult because that we can not ideally know the distribution of dataset and its distortion. Therefore, the penalty selection free method is preferable than the manually selection mehtod. Thus the learning based optimization algorithms has been studied in the past decades. The typical learning based optimization is the Supervised Sequential Updating (SSU) algorithms, in this architecture, the parameters to be estimated are updated iteratively through a sequential of maps. The maps are trained in a supervised learning method, such as the Least Square Method. The Cascaded Pose Regression (CPR) algorithm \cite{P2010Cascaded} is a representative SSU method, it progressively refined the initialized parameter to the optimum, and each fine-tuning step is regressed by a linear map. The Discriminative Optimization algorithm \cite{vongkulbhisal2017discriminative} proposed by Jayakorn is also a typical SSU method, which achieved an outstanding performance in 3D point cloud rigid registration. But there remains some shortcoming in the original DO. Firstly, for each model point, the scene point cloud was only separated into "front" or "back" side of that model point. Therefore, they omitted the rotational relationships between the 3D piont pairwise. Secondly, the feature (descriptor) is designed as a Gaussian weighted histogram, which would be a sparse vector when the scene point cloud is far away from the model point cloud. So that the original DO will be failed in this case. To improve the performance of the original DO, we proposed an improved DO algorithm, which introduced two extra relative sides between the two point cloud, and they are "up-down" and "clockwise-anticlockwise" side. In addition, we re-weighted the histograms according to a known prior that is the model points' distribution. After these two steps, we improved the original DO's performance in 3D point cloud registration.

\section{Previous Work}
During the last decades, a mass of point registration algorithms have been proposed. These algorithms can be simply classified into two categories: 1) traditional approaches, which applied a handcrafted features to pre-match the correspondence, then searching the optimal registration parameters through a gradient based (first-order, second-order or higher-order) or derivation-free method (heuristic algorithms). In contrast, 2) the learning based approaches simultaneously predicted the registration parameters and point correspondence through the trained maps \cite{P2010Cascaded}, \cite{2013Superviseddescentmethod} or neural networks \cite{2020guo3Dclouds}.

\subsection{Traditional Methods}
\textcolor{black}{The representative traditional approach is Iterative Closest Point (ICP) \cite{1992ICP} and its variants \cite{rusinkiewicz2001ICP},\cite{fitzgibbon2003robust}. The ICP algorithm solved the point correspondence and transformation parameters alternatively until convergence. The drawback of ICP is that it is sensitive to the initialized paramters, so that the ICP is often used in the fine tuning  step. To avoid the influence of initialization, the global interaction between two point cloud must be established. The Coherent Point Drift (CPD) \cite{2010CPD} treats the scene point cloud as the sampling point from the mixed-Gaussian probability distribution, where the mixed-Gaussian is generated from the model point cloud. The CPD method registered the point cloud by maximizing the posterior probability of the scene point cloud. More recently, a Bayesian Coherent Point Drift (BCPD) algorithm was proposed, it discribes the coherent drift in the variational Bayesian inference thoery, and the experimental result demonstrated that the BCPD outperformed the original CPD in all cases. The Kernel Correlation (KC) method \cite{2004KC_kernel} aligns the densities of two point cloud through a "kernel-trick", it register the points pair by maximizing their correlation. Robust Point Matching (RPM) \cite{RPM1998} adopted a soft assignment rather than direct assignment, so that it was robust to degradations than strict correlation KC method. Iteratively Reweighted Least Squares (IRLS) \cite{bergstrom2014IRLS} applied various loss functions to overcome the outliers and large rotation angle. Gaussian Mixture Model Registration method \cite{GMMReg} reformulated the alignment as a statistical discrepency minimizing, and the L2 distance was adopted since it was differentiable and effecient. \cite{2016supportVecotrRegress} constructed the point cloud by a learned one-class support vector machine, then regressed the parameters by minimizing L2 error between the support vectors. \cite{2016Gravitational} treated the point cloud as particles with universal gravitation, then registered them through a simulation of movement in universal gravitation field. Despite the classical methods performed well, there remain some drawbacks, for example, 1) the traditional methods usually lack of generalization capability; 2) for a complicated transformation, the loss function may be non-convex and indifferentiable.}

\subsection{Learning-based Methods}
The learning based method can predicts the optimal registration parameters in a sequential or direct manner. The learning based method can be classified into two categories: 1) Supervised Sequential Update (SSU) based method and Deep Neural Network (DNN) based method.
\subsubsection{Supervised Sequential Update}
In the SSU architecture, the registration parameters can be updated through a sequential of regressions. The regressors are learned from the training samples which is the point cloud pair and the optimal registration parameter. The trained regressor maps the point cloud to an updating vector that points to the direction of global optimum. The first SSU method is a single regressor proposed by Cootes et al.\cite{1998AAM} for facial image alignment. They introduced an active appearance model which is learned from the shape variation and texture variation of image, and the image's residual was minimized through a iteratively regression. Jurie and Dhome \cite{HyperplaneTempMatch} utilized boost logistic regression to estimate the current displacement of image features to register images. Cristinacce and Cootes \cite{2007GentleBoost} introduced a GentleBoost technology to iteratively update the registration parameters until the terminated condition meets. Bayro-Corrochano and Ortegon-Aguilar \cite{2004Lietemptracking} \cite{2007Lie3Dtracking} applied Lie algebra thoery to describe the image projection and affine transformation, so that the expression of projection and affine can be described in a linear vector space. Thus the deriviation of loss function to rotation matrix becomes simple. Simultaneously, Tuzel et al. \cite{2008HOGfeatureLie} combined Lie algebra with the HOG feature to learn regressors. The methods mentioned above only have one regressor, and it would restrict the algorithm's performance. So, the approaches with a sequential maps are proposed. Saragih et al. \cite{2006IEBM} present the Iterative Error Bound Minimisation (IEBM) method, they utilized V-support vector regression \cite{2002Vsupport} to learn a sequence of regressors, and it performed well on the non-rigid registration task. Dollar et al. \cite{P2010Cascaded} present Cascade Poses Regression to solve the object pose estimation, and the learned maps not only can be applied in summation updating rules, but also other inversible composition rules. Cao et al. \cite{cao2014faceboost} also learned a sequence of regressors to minimized the residual of parameters in the boosting technology. Sun et al. \cite{2013DeepCascadefacial} proposed a neural networks to learn the feature extraction and parameters regression which is also a SSU method. Xiong et al. \cite{2013Superviseddescentmethod} \cite{xiong2015global} present Supervised Descent Method, which learned the gradient descent directions that minimizes a non-linear least-squares function in parameter space. The Supervised Descent Method minimizes the loss function without calculation of Jacobian matrix. Rather than proposing a fixed sequential procedure for the optimizaiton, Zimmermann et al. selected the regressors from the pre-trained regressors, and concatenate them into a new sequence \cite{2009OptimalSequence}. The DO algorithm \cite{vongkulbhisal2017discriminative} also inspired by the Supervised Descent Method, specially, it used the last learned regressor to infer the parameters until termination condition occurs.

\subsubsection{DNN-based Methods}
The renaissance of Nerual Network attracted much attention of computer vison researchers. A gread deal of DNN-based point cloud registration approaches have been propsed. Because the point clouds is unordered and the deep neural network requires a regular input data, deep learning was firstly applied in points classification and segmentation. The representative architecture are PointNet \cite{Qi2017PointNet} and PointNet++ \cite{Qi2017PointNetPP}, which is simply consist of Multi-Layer Perceptron, Feature Transform Block, and Max Pooling Layers. After that, the PointNet was adopted to point cloud registration. Yasuhiro Aoki \cite{aoki2019pointnetlk} combined the Lucas \& Kanade (LK) algorithm with PointNet for point cloud registration. In original PointNet, the "T-net" block is computed at each iteration. To improve the computational efficency, the authors replaced it with the Lucas \& Kanade algorithm, and the the differential executation was reduced to only once. X. Huang et al. \cite{huang2020feature} proposed a Feature Metric Registration network, they trained an encoder to extract features and a decoder to compute metric, then minimizing the feature metric projection error to register point clouds. Comparing to the traditional geometrical error, the special feature metric projection error is robust to points degradation. Sheng Ao et al. proposed a conceptually simple but sufficiently informative SpinNet \cite{ao2020SpinNet}. By projecting the point cloud into a cylindrical space, they designed a cylindrical convolutional neural layers to extract a rotational invariant local feature. The experimental result shows that the SpinNet achieved the best generalization ability across different benchmarks. Choy et al. present deep global registration \cite{choy2020deep}, which is consist of three differentiable modules: 1) a Unet-based network to estimate a inlier likelihood as weight for each correspondence; 2) a weighted Procrustes \cite{gower1975generalized} algorithm to predict the registration parameters; 3) a robust gradient-based $SE(3)$ optimizer to refine parameters. Gil Elbaz et al. \cite{elbaz20173d} present super-points descriptor extracted from local cluster of points, then predict the coarse to fine correspondence between descriptors. Experimental result reveals that the super-points descriptor aligned the large-scale and close-proximity point clouds successfully. Gojcic et al. \cite{gojcic2020learning} also present a coarse to fine method for multiview 3D point cloud registration. They proposed a novel Overlap Pooling Layer as the similarity metric estimation block to predict transformation parameters and utilized the Iterative Reweighted Least Squares (IRLS) to refine the point correspondence and transformation parameters alternatively. Lu et al. \cite{lu2019deepvcp} present DeepVCP which consist of deep feature extraction layer, weighting layer, deep feature embedding layer, and corresponding point generation layer. It is a typical End-to-End point cloud registration network.

\section{Proposed Approach}
\subsection{Original Discriminative Optimization}
The mission of point cloud registration optimization algorithm is to find the optimal parameters as rapid and precise as possible. The points registration can be described as:
\begin{equation}
	{\bf{x }}^ *   = {\arg } \mathop {\min}\limits_{\bf{x }} \left\{ {  \varphi \left( {{\bf T}{\rm{(}}{\bf{U}},{\bf{x }}),{\bf{V}}} \right) + \lambda \left\| {\bf{T}} \right\|} \right\}
	\label{eq:registrationformula}
\end{equation}
Where ${\bf{S}} = ({\bf{s}}_1 ,...,{\bf{s}}_S )^T  \in \mathbb{R}^{S \times 3} $ is the scene point clouds, and $\mathbf{M}=\left( \mathbf{m}_1,...,\mathbf{m}_M \right) ^T\in \mathbb{R}^{M\times 3}$ is the model point clouds. $\varphi:\mathbb{R}^{M \times 2}  \times \mathbb{R}^{S \times 2}  \to \mathbb{R}$ is a similarity metric function, and $\bf{\lambda }$ is a parameter that controls the tradeoff. A first order derivation optimizing step is:
\begin{equation}
{\bf{x}}_{k + 1}  = {\bf{x}}_k  - \left. {\mu \frac{{\partial \varphi ({\bf{x}})}}{{\partial {\bf{x}}}}} \right|_{{\bf{x}}_k } 
	\label{eq:updataformula}
\end{equation}
here, the parameter is updated in the opposite direction of the gradient. In practice, the derivation of $\varphi$ usually computational or undifferentialbe, and sometimes $\varphi$ is non-convex. So that numerical optimization methods may be inefficient, especially, in a high dimension search space, the handcrafted optimization algorithm may trap into local minimum.

The Discriminative Optimization (DO) algorithm predicts the registration parameter in a sequential updating manner, which is proved to be a excellent method. Here, we recall the original discriminative optimization briefly. In the DO the current estimation of parameter is refined by the trained map ${\bf D}_{k+1}$, it is expressed in Eq.\ref{eq:update}.
\begin{equation}
{\bf{x}}_{k+1} = {\bf{x}}_k - {\bf{D}}_{k+1} {\bf{h}}({\bf{x}}_k)
	\label{eq:update}
\end{equation}
Where ${\bf{x}}_k\in\mathbb{R}^p$ is the $k{\rm{-th}}$ estimation of parameters. $ {\bf{h}}({\bf{x}}_k )$ extracts feature from input data with the $t{\rm{-th}}$ estimated parameters $ {\bf{x}}_k$. Exactly, the full expression of $\bf{h}$ is:
\begin{equation}
{\bf{h}}({\bf{x}}_k ,{\bf M},{\bf S}^{(i)}):\mathbb{R}^{3 \times N_M }  \times \mathbb{R}^{3 \times N_{S^{(i)}} }  \times \mathbb{R}^p  \to \mathbb{R}^f 
	\label{eq:hexpress}
\end{equation}
but we abbreviate it as ${\bf{h}}({\bf{x}}_k )$, where ${\bf M}\in\mathbb{R}^{3\times N_M} $ is a fixed model point cloud and ${\bf S}^{(i)}\in \mathbb{R}^{3\times N_{S^{(i)}} } $ is the $i-\rm th$ scene point cloud. $N_M$, $N_{S^{(i)}}$ is the number of model point cloud and scene point cloud respectively. $ {\bf{D}}_{k + 1}  \in \mathbb{R}^{d \times f} $ is one of the learned maps (${\bf{D}}_{k+1}:\mathbb{R}^f \to \mathbb{R}^d $ ) which regress the feature ${\bf{h}}({\bf{x}}_k)$ to an updating vector $\Delta{\bf{x}}$. Therefore, the DO is a linear regeression method, and the updating continues until the terminant condition is met.

Given a training dataset $ \{ ({\bf{x}}_0^{(i)} ,{\bf{x}}_ * ^{(i)} ,{\bf{h}}^{(i)} )\} _{i = 1}^n $, where $i$ denotes the $i{\rm{-th}}$ instance, which includes the point cloud feature ${\bf{h}}^{(i)}$, the initialization ${\bf{x}}_0^{(i)}$, and the ground truth ${\bf{x}}_ * ^{(i)}$. $ {\bf{D}}_{k + 1} $ is trained by all of the training samples through the following formulation.
\begin{equation}
{\bf{D}}_{k + 1}  = \arg \mathop {\min }\limits_{{\bf{\tilde D}}} \frac{1}{N}\sum\limits_{i = 1}^N {\left\| {{\bf{x}}_ * ^{(i)}  - {\bf{x}}_k^{(i)}  + {\bf{\tilde Dh}}^{(i)} ({\bf{x}}_k^{(i)} )} \right\|_2^2  + \lambda \left\| {{\bf{\tilde D}}} \right\|_F^2 } 
	\label{eq:Dt_1argmin}
\end{equation}
The ${\bf{D}}_{t + 1}$ can be solved as follows: 
\begin{equation}
	{\bf{D}}_{k + 1} = \frac{1}{N}\sum\limits_{i = 1}^N {\frac{{({\bf{x}}_ * ^{(i)}  - {\bf{x}}_k^{(i)} ) \cdot {\bf{h}}^{(i)} ({\bf{x}}_k^{(i)} )^T }}{{\lambda  + {\bf{h}}^{(i)} ({\bf{x}}_k^{(i)} )^T  \cdot {\bf{h}}^{(i)} ({\bf{x}}_k^{(i)} )}}}
	\label{eq:D_hat}
\end{equation}
For all $i$ , ${\bf{x}}_k^{(i)} $  is updated to ${\bf{x}}_{k+1}^{(i)} $ through Eq.\eqref{eq:updateX}
\begin{equation}
{\bf{x}}_{k + 1}={\bf{x}}_k - {\bf{D}}_{k + 1} {\bf{h}}\left( {{\bf{x}}_k } \right)
\label{eq:updateX}
\end{equation}

To register the $\bf M$ and $\bf S$, the discriminative optimization algorithm can be summarized in the following three steps. 
\begin{itemize}
	\item Firstly, for every model point ${\bf m}_i$, fitting a local-plane at each point ${\bf m}_i$ upon its sixth nearest neighbor points.
	\item Secondly, for every model point ${\bf m}_i$, the scene points ${\bf S}^{(i)}$ was seperated into the "\textcolor{red}{front}" or "\textcolor{blue}{back}" side of the model point. 
	\item Thirdly, for every model point ${\bf m}_i$, the seperated two part of scene points, which is on "\textcolor{red}{front}" or "\textcolor{blue}{back}" side of ${\bf m}_i$, are summerized (with Gaussian weighting) into the histogram ${\bf{h}}_a$ and ${\bf{h}}_{a+N_{M}}$ respectively, here ${\bf{h}}\in \mathbb{R}^{2N_M}$, $a=1,...,N_M$.
\end{itemize}

For model point ${\bf m}_a$, ${\bf{h}}_a$ and ${\bf{h}}_{a+N_M}$ is the corresponding Gaussian weighted voting results that indicate the scene points on the front and back side of ${\bf m}_a$. The histogram $\bf h$, which is a descriptor, is formulated as follows.
\begin{equation}
	S_a^ +   = \{ s_b |n_a^T (\mathcal{T}(s_b ,x) - m_a ) > 0\} 
\end{equation}
\begin{equation}
	S_a^ -   = \{ s_b |n_a^T (T(s_b ,{\bf{x}}) - m_a ) \le 0\} 
\end{equation}
\begin{equation}
[{\bf{h}}({\bf{x}};S)]_a  = \frac{{1}}{z}(\sum\limits_{s_b  \in S_a^ +  } {\exp ( - \frac{1}{{\sigma ^2 }}\left\| {T\left( {s_b ;{\bf{x}}} \right) - m_a } \right\|_2^2 )} ) 
\end{equation}
\begin{equation}
[{\bf{h}}({\bf{x}};S)]_{a + N_M }  = \frac{{1}}{z}(\sum\limits_{s_b  \in S_a^ -  } {\exp ( - \frac{1}{{\sigma ^2 }}\left\| {T\left( {s_b ;{\bf{x}}} \right) - m_a } \right\|_2^2 )} )
\end{equation}

\subsection{Improved Discriminative optimization}
\begin{figure}[htpb]
	\centering
	\subfloat[front-back side]{\includegraphics[width=0.7\linewidth,height=0.5\linewidth]{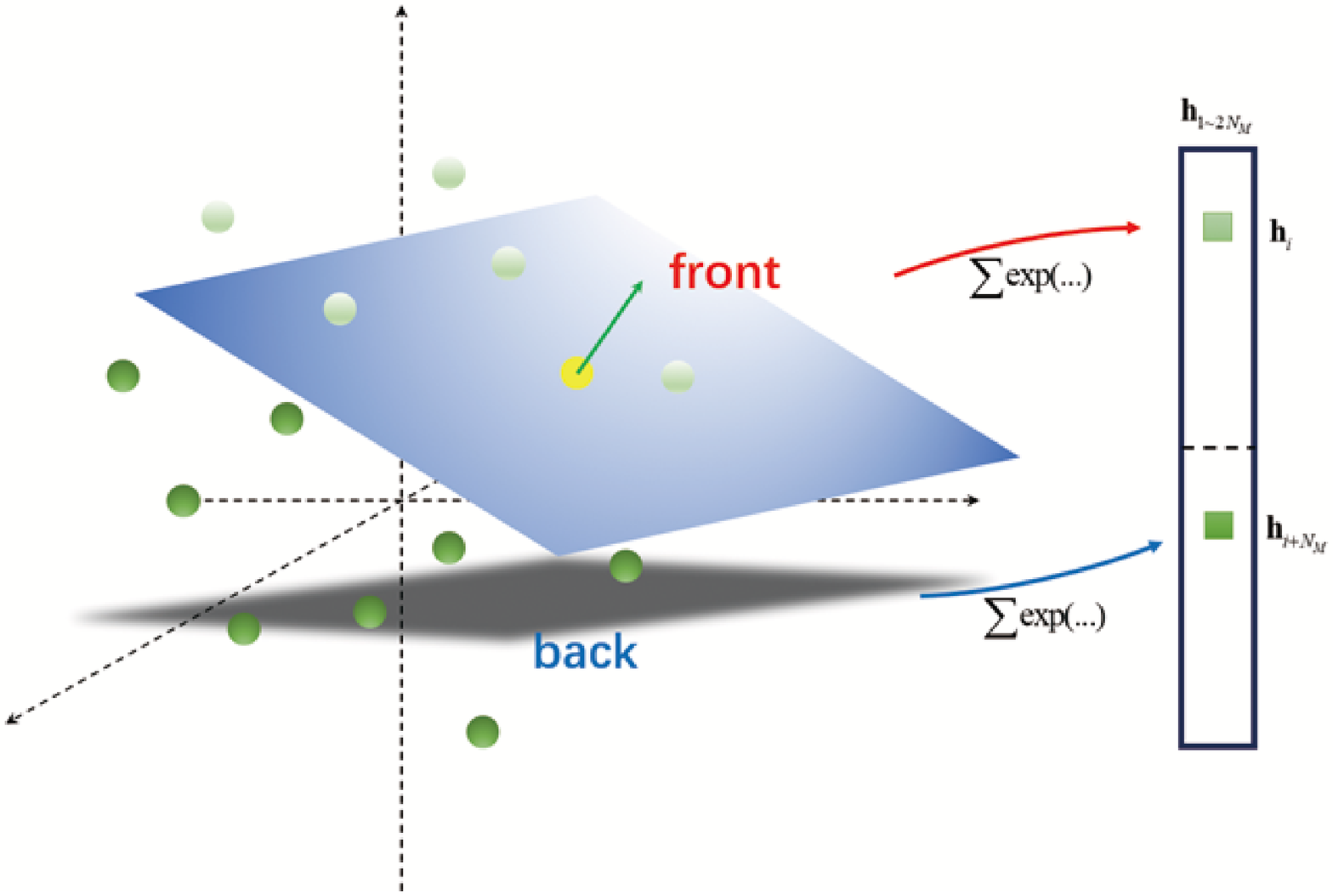}}
	\label{subfig:front-back}
	\hspace{1em}
	\subfloat[up-down side]{\includegraphics[width=0.7\linewidth,height=0.5\linewidth]{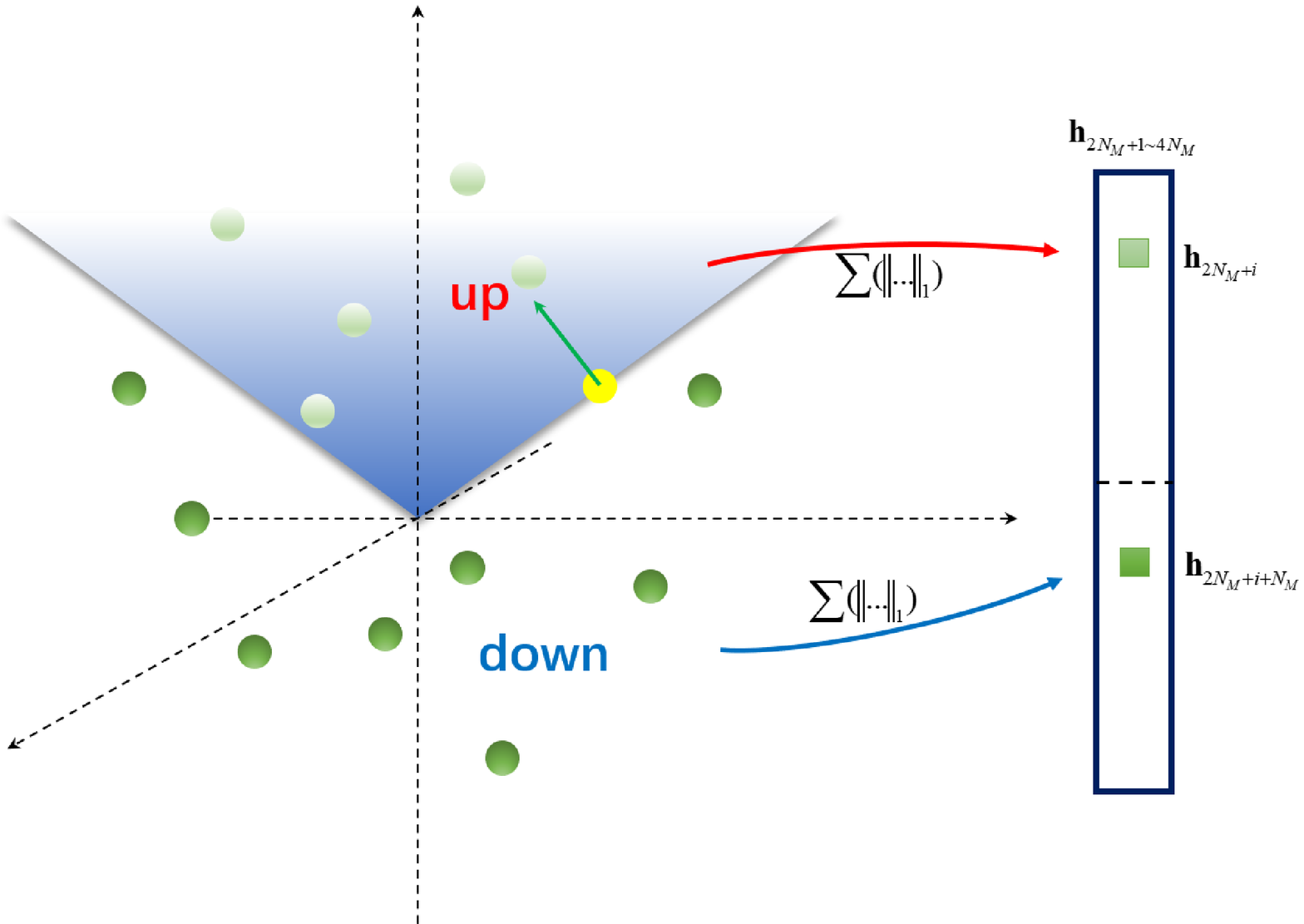}}
	\label{subfig:up-down}
	\hspace{1em}
	\subfloat[clockwise-anticlockwise side]{\includegraphics[width=0.7\linewidth,height=0.5\linewidth]{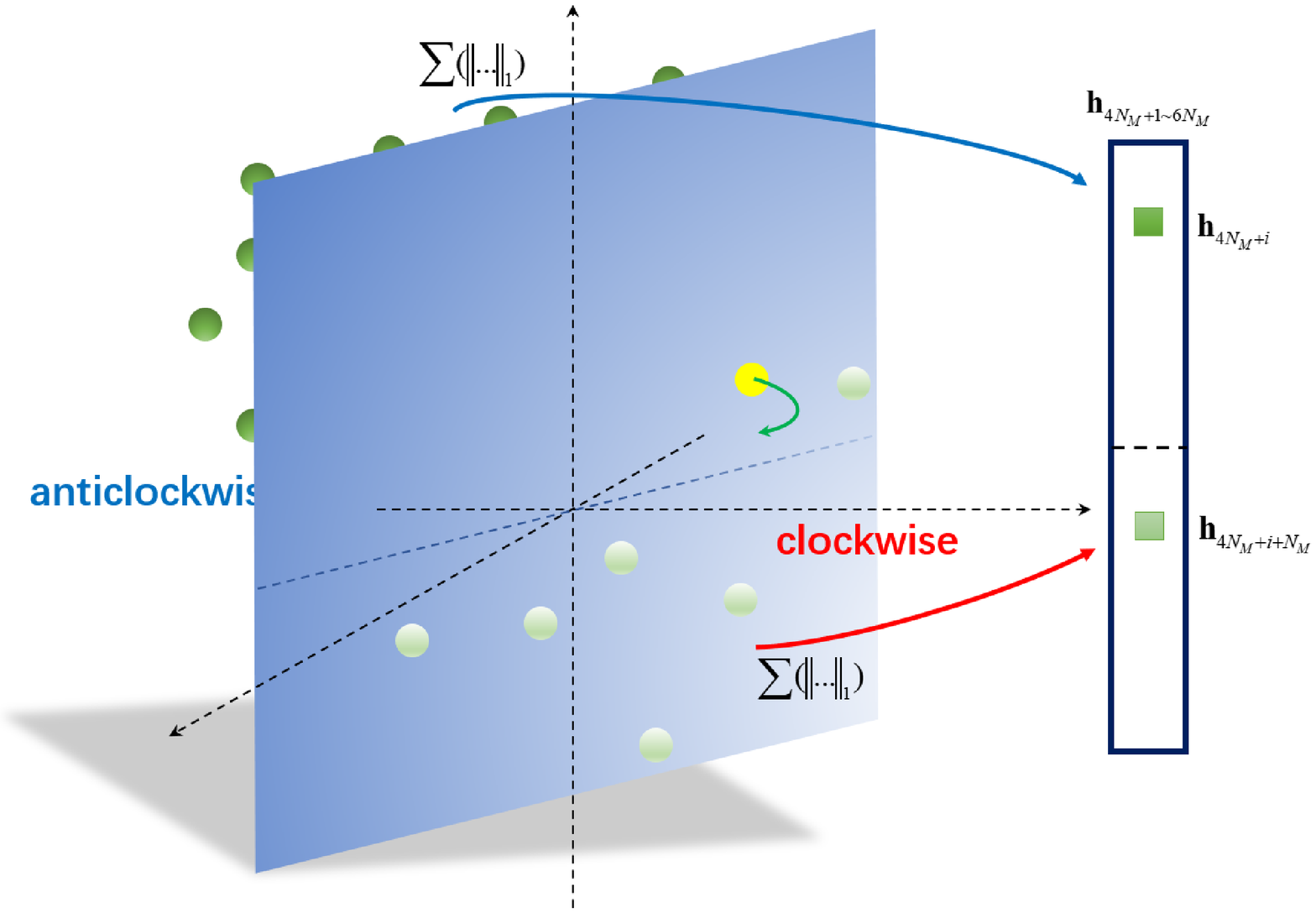}}
	\label{subfig:clockwise-anticlockwise}
	\caption{\textcolor{black}{the "triple-binary" diagram}}
	\label{multirelationship}
\end{figure}

The original Discriminative Optimization method does not embed enough information of model points in the descriptor, which restricts its performance in tough degradation datasets. To utilize the relationships between model points and scene points (shown in Fig.\ref{multirelationship}), we extend the descriptor $\bf{h}$ from the "\textcolor{red}{front} (${S_a^+}$) - \textcolor{blue}{back} (${S_a^-}$)" side, that we termed it as "single-binary", to "\textcolor{red}{front} (${S_a^+}$) - \textcolor{blue}{back} (${S_a^-}$)", "\textcolor{red}{up} (${S_a^ \uparrow}$) - \textcolor{blue}{down} (${S_a^ \downarrow}$)" and "\textcolor{red}{clockwise} (${S_a^ \to}$) - \textcolor{blue}{anticlockwise} (${S_a^ \leftarrow}$)" side, that we termed it as "triple-binary". In Fig.\ref{multirelationship}, a model point ${\bf M}_a$ is depicted in yellow dot, and we paint those scene points  pale green which are in "front" or "up" or "clockwise" of ${\bf M}_a$, and those in the opposite side are painted dark green. The "up-down" and "clockwise-anticlockwise" sides are described in Eq.\eqref{up} to Eq.\eqref{anticlock}.
\begin{equation}
S_a^ \uparrow   = \{ {\bf s}_b |\theta _{T({\bf s}_b ,{\bf{x}})}  > \theta _{{\bf m}_a } \} 
\label{up}
\end{equation} 
\begin{equation}
S_a^ \downarrow   = \{ {\bf s}_b |\theta _{T({\bf s}_b ,{\bf{x}})}  \le \theta _{{\bf m}_a } \} 	
\label{down}
\end{equation}
\begin{equation}
	S_a^ \to   = \{ {\bf s}_b |\omega _{T({\bf s}_b ,{\bf{x}})}  > \omega _{{\bf m}_a } \}
	\label{clock}
\end{equation}
\begin{equation}
	S_a^ \leftarrow   = \{ {\bf s}_b |\omega _{T({\bf s}_b ,{\bf{x}})}  > \omega _{{\bf m}_a } \} 
	\label{anticlock}
\end{equation}
Where $\theta$, $\omega$ are elevation angle and azimuth angle of a 3D point, respectively. Here, we rewrite the whole histogram as follows.
\begin{equation}
[{\bf{h}}({\bf{x}};S)]_a  = \frac{{\alpha _a }}{z}(\sum\limits_{{\bf s}_b  \in S_a^ +  } {\exp ( - \frac{1}{{\sigma ^2 }}\left\| {T\left( {{\bf s}_b ;{\bf{x}}} \right) - {\bf m}_a } \right\|_2^2 )} )	
\end{equation}
\begin{equation}
[{\bf{h}}({\bf{x}};S)]_{a + N_M }  = \frac{{1 - \alpha _a }}{z}(\sum\limits_{{\bf s}_b  \in S_a^ -  } {\exp ( - \frac{1}{{\sigma ^2 }}\left\| {T\left( {{\bf s}_b ;{\bf{x}}} \right) - {\bf m}_a } \right\|_2^2 )} )	
\end{equation}
\begin{equation}
[{\bf{h}}({\bf{x}};S)]_{a + 2N_M }  = \frac{{\beta _a }}{{z'}}(\sum\limits_{{\bf s}_b  \in S_a^ \uparrow  } {\left\| {(\theta _{T({\bf s}_b ;{\bf{x}})}  - \theta _{{\bf m}_a } ) > 0} \right\|_1 } )	
\end{equation}
\begin{equation}
[{\bf{h}}({\bf{x}};S)]_{a + 3N_M }  = \frac{{1 - \beta _a }}{{z'}}(\sum\limits_{{\bf s}_b  \in S_a^ \downarrow  } {\left\| {(\theta _{T({\bf s}_b ;{\bf{x}})}  - \theta _{{\bf m}_a } ) < 0} \right\|_1 } )	
\end{equation}
\begin{equation}
[{\bf{h}}({\bf{x}};S)]_{a + 4N_M }  = \frac{{\gamma _a }}{{z''}}(\sum\limits_{{\bf s}_b  \in S_a^ \to  } {\left\| {(\omega _{T({\bf s}_b ;{\bf{x}})}  - \omega _{{\bf m}_a } ) > 0} \right\|_1 } )	
\end{equation}
\begin{equation}
[{\bf{h}}({\bf{x}};S)]_{a + 5N_M }  = \frac{{1 - \gamma _a }}{{z''}}(\sum\limits_{{\bf s}_b  \in S_a^ \leftarrow  } {\left\| {(\omega _{T({\bf s}_b ;{\bf{x}})}  - \omega _{{\bf m}_a } ) < 0} \right\|_1 } )	
\end{equation}
Here, $z$, $z'$, $z''$ is the normalization term, $\alpha _a$, $\beta _a$, $\gamma _a$ are weights as determined by the model point cloud, as the following equations.
\begin{equation}
\alpha _a  = {{N_a^ +  } \mathord{\left/
		{\vphantom {{N_a^ +  } {N_M }}} \right.
		\kern-\nulldelimiterspace} {N_M }}	
\end{equation}
\begin{equation}
\beta _a  = N_a^ \uparrow  /N_M 	
\end{equation}
\begin{equation}
	\gamma _a  = {{N_a^ \to  } \mathord{\left/
			{\vphantom {{N_a^ \to  } {N_M }}} \right.
			\kern-\nulldelimiterspace} {N_M }}
\end{equation}
Where, $N_a^ + $ represents the number of subset of model point, whose position are in "front" of the model point ${\bf m}_a$. Similarly, $N_a^ \uparrow $ represents the number of subset of model point, whose elevation angle are larger than $\theta _{{\bf m}_a }$, and $N_a^ \to $ means the subset of model points' number, the subset's azimuth angle are larger then $\gamma _{{\bf m}_a }$.

Since we have encoded the elevation angle and azimuth angle into the re-designed histogram, the information of descriptor has been enriched and would improving the training efficiency and testing accuracy. Exactly, $\bf{h}$ is a descriptor which encodes the distribution of the scene point cloud on the field of model points. It is noteworthy that we only add two "binary" to divide the scene point cloud eventhough the points in a 3D space. Considering an arbitrary rotation in the 3D space, the rotated points can be restored after two steps as follows. 1)  rotate along with a proper elevation angle, which aligned the axis deviated $Z'$ with the original axis $Z$. 2) rotate along with a proper azimuth angle, which aligns the deviated $X'-Y'$ axises with the original $X-Y$ axises. As a consequencewe, we seperate the additional two "binary" according to elevation angle and azimuth angle.

\subsection{Lie groups and Lie algebras application}
The rigid transformation contains a rotation matrix $\bf R$ and a translation vector $\bf t$, since the Discriminative Optimization method is a linear regressor, but the rigid transformation ${\bf T}=[{\bf R},{\bf t}]$ described in Euclidian space is a nonlinear tranformation model, so that optimizing the rigid transformation matrix would be complexed. Thanks to the Lie groups and Lie algebras theory \cite{2015LiegroupAlgebras}, which merges the rotation and translation operations into a linear vector space. Here we briefly introduced the Lie theory and utilize it to parametrize the rigid transformation. In Lie group theory, the rotation matrix ${\bf{R}} \in \mathbb{R}^{3\times 3}$ forms a special orthogonal group $SO(3)$, while the rigid transformation matrix $\bf{T}$ forms a special Euclidian group $SE(3)$.
\begin{equation}
SO(3) = \left\{ {{\bf{R}} \in \mathbb{R}^{3 \times 3} |{\bf{RR}}^T  = {\bf{I}},\det ({\bf{R}}) = 1} \right\}
\end{equation}
\begin{equation}
SE(3) = \left\{ {{\bf{T}} = \left[ {\begin{array}{*{20}c}
			{\bf{R}} & {\bf{t}}  \\
			0 & 1  \\
	\end{array}} \right] \in \mathbb{R}^{4 \times 4} |{\bf{R}} \in SO(3),{\bf{t}} \in \mathbb{R}^3 } \right\}
\end{equation}
The Lie algebras $\mathfrak{so}(3)$, which is corresponding to the Lie group $SO(3)$, is defined as:
\begin{equation}
\mathfrak{so}(3) = \left\{ {\phi = \theta \bf{n}  \in \mathbb{R}^3 ,\phi ^ \wedge   \in \mathbb{R}^{3 \times 3} } \right\}
\end{equation}
where $\bf{n}$ is the rotation direction unit vector and $\theta$ is the rotation angle, and
\begin{equation}
	\phi ^ \wedge   = \bf{\Phi}  = \left[ {\begin{array}{*{20}c}
			0 & { - \phi _3 } & {\phi _2 }  \\
			{\phi _3 } & 0 & { - \phi _1 }  \\
			{ - \phi _2 } & {\phi _1 } & 0  \\
	\end{array}} \right], {\bf{\Phi}} ^ \vee   = \phi
\end{equation}
Then the relationship between ${\bf{R}} \in SO(3)$ and  $\phi \in \mathfrak{so}(3)$ is:
\begin{equation}
	{\bf{R}} = \exp (\phi ^ \wedge) = \exp (\theta {\bf{n}} ^ \wedge)
\end{equation}
\begin{equation}
	\phi  = \ln ({\bf{R}})^ \vee 
\end{equation}
A Lie algebras vector convert to the Lie group rotation matrix through the famous Rodrigues formula Eq.\eqref{Rodrigues}.
\begin{equation}
	{\bf{R}} = \exp (\theta {\bf{n}} ^ \wedge) = \cos \theta {\bf{I}} + (1 - \cos \theta ) {\bf{nn}}^T  + \sin \theta {\bf{n}}^ \wedge
	\label{Rodrigues}   
\end{equation}
and the inverse is:
\begin{equation}
	\theta  = \arccos \frac{{{\rm{tr}}({\bf{R}}) - 1}}{2}, {\bf{Rn}} = {\bf{n}}   
\end{equation}
The Lie algebras $\mathfrak{se}(3)$ is defined as Eq.\eqref{Liealgebras_se3}
\begin{equation}
	\mathfrak{se}(3) = \left\{ {{\bf{\xi }} = \left[ {\begin{array}{*{20}c}
				{\bf{\rho }}  \\
				\phi   \\
		\end{array}} \right] \in \mathbb{R}^6 ,{\bf{\rho }} \in \mathbb{R}^3 ,\phi  \in \mathfrak{so}(3)} \right\} 
	\label{Liealgebras_se3} 
\end{equation}
and the conversion between ${\bf{T}} \in SE(3)$ and $\xi \in \mathfrak{se}(3)$ are:
\begin{equation}
	{\bf{T}} = \exp ({\bf{\xi }}^ \Delta  )  
\end{equation}
\begin{equation}
{\bf{\xi }} = \ln ({\bf{T}})^ \nabla  
\end{equation}
Where,
\begin{equation}
	{\bf{\xi }}^ \Delta = {\bf{\Xi}} = \left[ {\begin{array}{*{20}c}
			{\phi ^ \wedge  } & {\bf{\rho }}  \\
			{{\bf{0}}^T } & 0  \\
	\end{array}} \right], {\bf{\Xi}} ^ \nabla = \bf{\xi } 
\end{equation}
the vector in $\mathfrak{se}(3)$ convert to Lie group matrix through the Eq.\eqref{eq_se3_to_SE3}.
\begin{equation}
	{\bf{T}} =  \exp ({\bf{\xi }}^ \Delta  ) = \left[ {\begin{array}{*{20}c}
			{\exp (\phi ^ \wedge  )} & {{\bf{J}}\rho }  \\
			{{\bf{0}}^T } & 1  \\
	\end{array}} \right]
	\label{eq_se3_to_SE3}
\end{equation}
where, $\bf J$ solved as follows:
\begin{equation}
{\bf{J}} = \frac{{\sin \theta }}{\theta }{\bf I} + \left( {1 - \frac{{\sin \theta }}{\theta }} \right){\bf{nn}}^T  + \frac{{1 - \cos \theta }}{\theta }{\bf{n}}^ \wedge
\end{equation}
and the conversion formula from $SE(3)$ to $\mathfrak{se}(3)$ is:
\begin{equation}
\theta  = \arccos \frac{{{\rm{tr}}({\bf{R}}) - 1}}{2},{\bf{Rn}} = {\bf{n}},{\bf{t}} = {\bf{J\rho }}
\end{equation}
Therefore, the rigid transformation can be parametrized from $ {\bf T} \in \mathbb{R}^{4 \times 4}$ to ${\bf \phi} \in \mathbb{R}^6$, and the parameter regression and updating becomes:
$ {\bf \phi}' =  {\bf \phi} + \Delta {\bf \phi} $, rather than a complicated formula: ${\bf T}' =  f({\bf T}, \Delta {\bf T}) $.

\subsection{Proof of Convergence}
A major question is that Does the sequential updating method converge to the optimum? Here, we provide detailed provement.

{\bf Theorem 1: Strict decrease in training error under a sequence of update maps.}

Given a training dataset,
\begin{equation}
\left\{ {\left( {{\bf {x}}_0^{(i)} ,{\bf {x}}_ * ^{(i)} ,{\bf {h}}^{(i)} } \right)} \right\}_{i = 1}^N
\end{equation}
If there exists a linear map ${\bf{\hat D}} \in \mathbb{R} ^{p \times f}$, such that, for each $i$, ${\bf{\hat Dh}}^{(i)}$ is strictly monotone at
 ${\bf{x}}_ *  ^{(i)}$, and if $\exists i:{\bf{x}}_k^{(i)}  \ne {\bf{x}}_ * ^{(i)} $, then the update rule:
\begin{equation}
{\bf x}_{k + 1}^{(i)}  = {\bf x}_k^{(i)}  - {\bf {D}}_{{{k + 1}}} {\bf {h}}^{(i)} ({\bf x}_k^{(i)} )
\end{equation}
With $ {\bf {D}}_{k + 1}  \in \mathbb{R}^{p \times f} $ obtained from: 
\begin{equation}
{\bf {D}}_{k + 1}  = \arg \mathop {\min }\limits_{\hat {\bf D}} \frac{1}{N}\sum\limits_{i = 1}^N {\left\| {{\bf {x}}_ * ^{(i)}  - {\bf {x}}_k^{(i)}  + {\bf {\hat Dh}}^{(i)} ({\bf {x}}_k^{(i)} )} \right\|^2 } 
\end{equation}

It guarantees that the training error strictly decreases in each iteration:   
\begin{equation}
\sum\limits_{i = 1}^N {\left\| {{\bf {x}}_ * ^{(i)}  - {\bf {x}}_{k + 1}^{(i)} } \right\|^2 }  < \sum\limits_{i = 1}^N {\left\| {{\bf {x}}_ * ^{(i)}  - {\bf {x}}_k^{(i)} } \right\|^2 } 
\end{equation}

{\bf{Proof:}} We assume that not all ${\bf{x}}_ * ^{(i)} {\bf{ = x}}_k^{(i)} $, otherwise all ${\bf{x}}_ * ^{(i)} 
$   are already at their optimal points. Thus, there exists an $i$ such that: 
\begin{equation}
{\bf {(x}}_k^{(i)}  - {\bf {x}}_ * ^{(i)} {\bf {)}}^T {\bf {\hat Dh}}^{(i)} ({\bf {x}}_k^{(i)} ) > 0
\end{equation}

We need to proof that:
\begin{equation}
\sum\limits_{i = 1}^N {\left\| {{\bf {x}}_ * ^{(i)}  - {\bf {x}}_{k + 1}^{(i)} } \right\|_2^2 }  < \sum\limits_{i = 1}^N {\left\| {{\bf {x}}_ * ^{(i)}  - {\bf {x}}_k^{(i)} } \right\|_2^2 } 
\end{equation}
This can be shown by letting:
\begin{equation}
{\bf {\bar D}} = \alpha {\bf {\hat D}}
\end{equation}
where:
\begin{equation}
\alpha  = \frac{\beta }{\gamma }
\end{equation}
\begin{equation}
\beta {=}\sum\limits_{i = 1}^N {{\bf {(x}}_k^{(i)}  - {\bf {x}}_ * ^{(i)} {\bf {)}}^T {\bf {\hat Dh}}^{(i)} ({\bf {x}}_k^{(i)} )} 
\end{equation}
\begin{equation}
\gamma {\bf { = }}\sum\limits_{i = 1}^N {\left\| {{\bf {\hat Dh}}^{(i)} ({\bf {x}}_k^{(i)} )} \right\|_2^2 } 
\end{equation}
Since there exists an $i$  such that ${\bf {(x}}_k^{(i)}  - {\bf {x}}_ * ^{(i)} {{)}}^{\bf {T}} {\bf {\hat Dh}}^{(i)} ({\bf {x}}_k^{(i)} ) > 0$, both $\beta $ and $\gamma$  are both positive, so that $\alpha$ is also positive. The training error decreases in each iteration as follows:
\begin{equation}
\begin{array}{l}
	\sum\limits_{i = 1}^N {\left\| {{\bf {x}}_ * ^{(i)}  - {\bf {x}}_{k + 1}^{(i)} } \right\|_2^2 }  = \sum\limits_{i = 1}^N {\left\| {{\bf {x}}_ * ^{(i)}  - {\bf {x}}_k^{(i)}  + {\bf {D}}_{k + 1} {\bf {h}}^{(i)} ({\bf {x}}_k^{(i)} )} \right\|_2^2 }  \\ 
	\begin{array}{*{20}c}
		{} & {\begin{array}{*{20}c}
				{} &  \le  & {\sum\limits_{i = 1}^N {\left\| {{\bf {x}}_ * ^{(i)}  - {\bf{x}}_k^{(i)}  + {\bf {\bar Dh}}^{(i)} ({\bf {x}}_k^{(i)} )} \right\|_2^2 } }  \\
		\end{array}}  \\
	\end{array} \\ 
\end{array}
\end{equation}
The inequality is because ${\rm{D}}_{k + 1}$ being the optimal matrix that minimizes the squared error:
\begin{equation}
{\bf {D}}_{k + 1}  = \arg \mathop {\min }\limits_{\bf {\hat D}} \frac{1}{N}\sum\limits_{i = 1}^N {\left\| {{\bf {x}}_ * ^{(i)}  - {\bf {x}}_k^{(i)}  + {\bf {\hat Dh}}^{(i)} ({\bf {x}}_k^{(i)} )} \right\|^2 }
\end{equation}
So that:
\begin{equation}
\begin{array}{l}
	\begin{array}{*{20}c}
		{} & {\sum\limits_{i = 1}^N {\left\| {{\bf {x}}_ * ^{(i)}  - {\bf {x}}_k^{(i)}  + {\bf {D}}_{k + 1} {\bf {h}}^{(i)} ({\bf {x}}_k^{(i)} )} \right\|^2 } }  \\
	\end{array} \\ 
	\begin{array}{*{20}c}
		\le  & {\sum\limits_{i = 1}^N {\left\| {{\bf {x}}_ * ^{(i)}  - {\bf {x}}_k^{(i)}  + {\bf {\hat Dh}}^{(i)} ({\bf {x}}_k^{(i)} )} \right\|^2 } }  \\
	\end{array} \\ 
\end{array}
\end{equation}
and by recalling the formula: ${\bf {\bar D}} = \alpha {\bf {\hat D}}$, We only need to proof: $\alpha  = \frac{\beta }{\gamma } > 1$, where,
\begin{equation}
\alpha  = \frac{\beta }{\gamma }{\bf { = }}\frac{{\sum\limits_{i = 1}^N {{\bf {(x}}_k^{(i)}  - {\bf {x}}_ * ^{(i)} {\bf {)}}^T {\bf {\hat Dh}}^{(i)} ({\bf {x}}_k^{(i)} )} }}{{\sum\limits_{i = 1}^N {\left\| {{\bf {\hat Dh}}^{(i)} ({\bf {x}}_k^{(i)} )} \right\|_2^2 } }}
\end{equation}

For the sake of proving $\alpha  > 1$, we only need to calculate the square of $L_2$ norm on the numerator and denominator respectively.
\begin{equation}
\begin{array}{l}
	\sum\limits_{i = 1}^N {\left\| {{\bf {(x}}_k^{(i)}  - {\bf {x}}_ * ^{(i)} {\bf {)}}^T {\bf {\hat Dh}}^{(i)} ({\bf {x}}_k^{(i)} )} \right\|_{{2}}^{{2}} }  \\ 
	= \sum\limits_{i = 1}^N {\left\| {{\bf {(x}}_k^{(i)}  - {\bf {x}}_ * ^{(i)} {\rm{)}}} \right\|_{{2}}^{{2}} }  + \sum\limits_{i = 1}^N {\left\| {{\bf {\hat Dh}}^{(i)} ({\bf {x}}_k^{(i)} )} \right\|_{{2}}^{{2}} }  \\ 
	\begin{array}{*{20}c}
		{} & { + \underbrace {\sum\limits_{i = 1}^N {{{2({\bf x}}}_k^{(i)}  - {\bf {x}}_ * ^{(i)} {{)}}^T {\bf {\hat Dh}}^{(i)} ({\bf {x}}_k^{(i)} )} }_{ > 0}}  \\
	\end{array} \\ 
	\ge \sum\limits_{i = 1}^N {\left\| {{\bf {\hat Dh}}^{(i)} ({\bf {x}}_k^{(i)} )} \right\|_2^2 }  \\ 
\end{array}
\label{numerator_denominator}
\end{equation}

Frome the Eq.\eqref{numerator_denominator}, we proved that $\alpha > 1$, and the {\bf Theorem 1} can be proved completely as follows.
\begin{equation}
\begin{array}{l}
	\begin{array}{*{20}c}
		{} & {\sum\limits_{i = 1}^N {\left\| {{\bf {x}}_ * ^{(i)}  - {\bf {x}}_k^{(i)}  + {\bf {\bar Dh}}^{(i)} ({\bf {x}}_k^{(i)} )} \right\|_2^2 } }  \\
	\end{array} \\ 
	\begin{array}{*{20}c}
		=  & {\sum\limits_{i = 1}^N {\left( {\left\| {{\bf {x}}_ * ^{(i)}  - {\bf {x}}_k^{(i)} } \right\|_2^2  + \left\| {{\bf {\bar Dh}}^{(i)} ({\bf {x}}_k^{(i)} )} \right\|_2^2 } \right.} }  \\
	\end{array} \\ 
	\begin{array}{*{20}c}
		{} & { + \left. {2({\bf {x}}_ * ^{(i)}  - {\bf {x}}_k^{(i)} )^{{T}} {\bf {\bar Dh}}^{(i)} ({\bf {x}}_k^{(i)} )} \right)}  \\
	\end{array} \\ 
	\begin{array}{*{20}c}
		=  & {\sum\limits_{i = 1}^N {\left( {\left\| {{\bf {x}}_ * ^{(i)}  - {\bf {x}}_k^{(i)} } \right\|_2^2 } \right) + \underbrace {\sum\limits_{i = 1}^N {\left( {\left\| {\alpha {\bf {\hat Dh}}^{(i)} ({\bf {x}}_k^{(i)} )} \right\|_2^2 } \right)} }_{ = \alpha ^2 \gamma }} }  \\
		{} & { + \underbrace {2\alpha \sum\limits_{i = 1}^N {\left( {({\bf {x}}_ * ^{(i)}  - {\bf {x}}_k^{(i)} )^{ {T}} {\bf {\hat Dh}}^{(i)} ({\bf {x}}_k^{(i)} )} \right)} }_{ =  - \beta }}  \\
	\end{array} \\ 
	\begin{array}{*{20}c}
		=  & {\sum\limits_{i = 1}^N {\left( {\left\| {{\bf {x}}_ * ^{(i)}  - {\bf {x}}_k^{(i)} } \right\|_2^2 } \right)}  + \alpha ^2 \gamma  - 2\alpha \beta }  \\
		=  & {\sum\limits_{i = 1}^N {\left( {\left\| {{\bf {x}}_ * ^{(i)}  - {\bf {x}}_k^{(i)} } \right\|_2^2 } \right)}  + \frac{{\beta ^2 }}{\gamma } - 2\frac{{\beta ^2 }}{\gamma }}  \\
	\end{array} \\ 
	\begin{array}{*{20}c}
		=  & {\sum\limits_{i = 1}^N {\left( {\left\| {{\bf {x}}_ * ^{(i)}  - {\bf {x}}_k^{(i)} } \right\|_2^2 } \right)}  - \underbrace {\frac{{\beta ^2 }}{\gamma }}_{ > 0}}  \\
		<  & {\sum\limits_{i = 1}^N {\left( {\left\| {{\bf {x}}_ * ^{(i)}  - {\bf {x}}_k^{(i)} } \right\|_2^2 } \right)} }  \\
	\end{array} \\ 
\end{array}	
\end{equation}
It is noteworthy that the learned function must be strictly monotone at $\bf x^*$, fortunately, it is possible in point registration. {\bf Theorem.1} does not guarantee that the error of each sample reduced in each iteration, but it guarantees the reduction of average error. The algorithm are summerized in Alg.1 and Alg.2. Here, $\epsilon$ means the terminated condition when the updating parameters' norm is less than it, and $maxIter$ for terminating the infinite infering.
\begin{algorithm}[t]
	\caption{Training phase}
	\label{alg:LearnSUM}
	\begin{algorithmic}[1]
		\REQUIRE $\{(\mathbf{x}^{(i)}_0,\mathbf{x}_*^{(i)},\mathbf{h}^{(i)})\}_{i=1}^N,K,\lambda$
		\ENSURE $\{\mathbf{D}_k\}_{k=1}^{K}$
		\FOR{$k= 0$ \TO $K-1$}
		\STATE Compute $\mathbf{D}_{k+1}$ with~\eqref{eq:Dt_1argmin}.
		\FOR{$i = 1$ \TO $N$}
		\STATE Update  $\mathbf{x}_{k+1}^{(i)}:=\mathbf{x}_{k}^{(i)}-\mathbf{D}_{k+1}\mathbf{h}^{(i)}(\mathbf{x}_{k}^{(i)})$.
		\ENDFOR
		\ENDFOR
	\end{algorithmic}
\end{algorithm}

\begin{algorithm}[t]
	\caption{Testing phase}
	\label{alg:updateParam}
	\begin{algorithmic}[1]
		\REQUIRE $\mathbf{x}_0,\mathbf{h},\{\mathbf{D}_k\}_{k=1}^{K},maxIter,\epsilon$
		\ENSURE $\mathbf{x}$
		\STATE Set $\mathbf{x}:=\mathbf{x}_0$
		\FOR{$k=1$ \TO ${K}$}
		\STATE Update $\mathbf{x} := \mathbf{x}-\mathbf{D}_k\mathbf{h}(\mathbf{x})$
		\ENDFOR
		\STATE Set $iter := {K}+1$.
		\WHILE{$\Vert\mathbf{D}_{K} \mathbf{h}(\mathbf{x})\Vert\geq\epsilon$ \AND $iter \leq maxIter$}
		\STATE Update $\mathbf{x} := \mathbf{x}-{\mathbf{D}}_{K}\mathbf{h}(\mathbf{x})$
		\STATE Update $iter := iter + 1$
		\ENDWHILE
	\end{algorithmic}
\end{algorithm}
              
\section{Experiments and Discussion}
In this section, we evaluate our improved DO by comparing it with other six State-Of-The-Art (SOTA) algortihms. They are Coherent Point Drift (CPD) \cite{2010CPD}, Gaussian Mixture Model Registration (GMMReg) \cite{GMMReg}, Iterarive Closest Point (ICP) \cite{1992ICP}, Kernel Correlation with Correspondence Estimation (KCCE) \cite{KCCE2011}, Bayesian Coherent Point Drift (BCPD) \cite{HiroseBCPD} which is the latest method, and the Original Discriminative Optimization (DO) \cite{vongkulbhisal2017discriminative}. To provide a quantitative comparison, we listed six algorthms' experimental results on two evalution criteria. All of the compared algorithms are implemented in the available source codes from their website. We tested all algorithms on the public available dataset. The experiments are executed on a desktop equipped with Intel Xeon 2.5GHz, Nividia P5000 GPU, and 16G RAM.

\subsection{Experimental Dataset Preparation}
\begin{figure}[t]
	\centering
	\subfloat[Synthetic Dataset]{\includegraphics[height=0.4\linewidth]{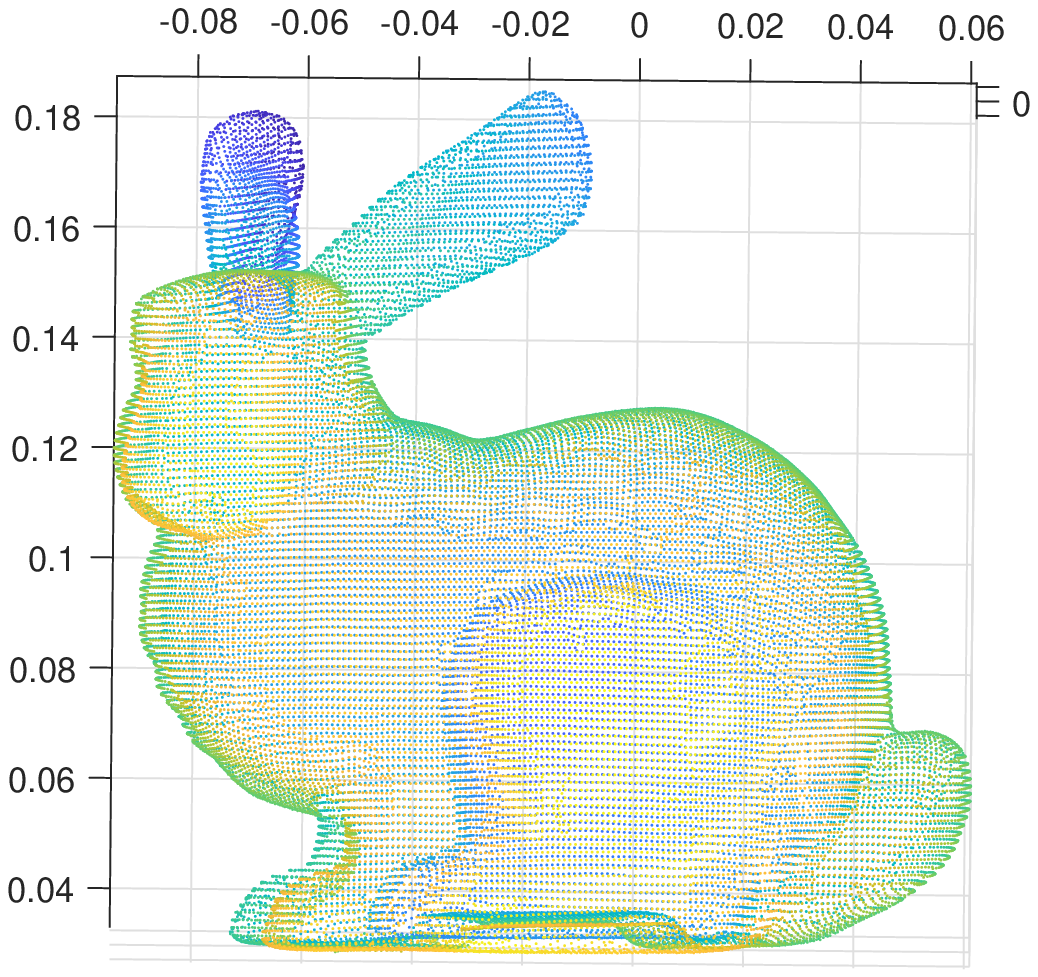}}
	\label{subfig:dat1-bunny}
	\hspace{1em}
	\subfloat[Real Dataset]{\includegraphics[height=0.4\linewidth]{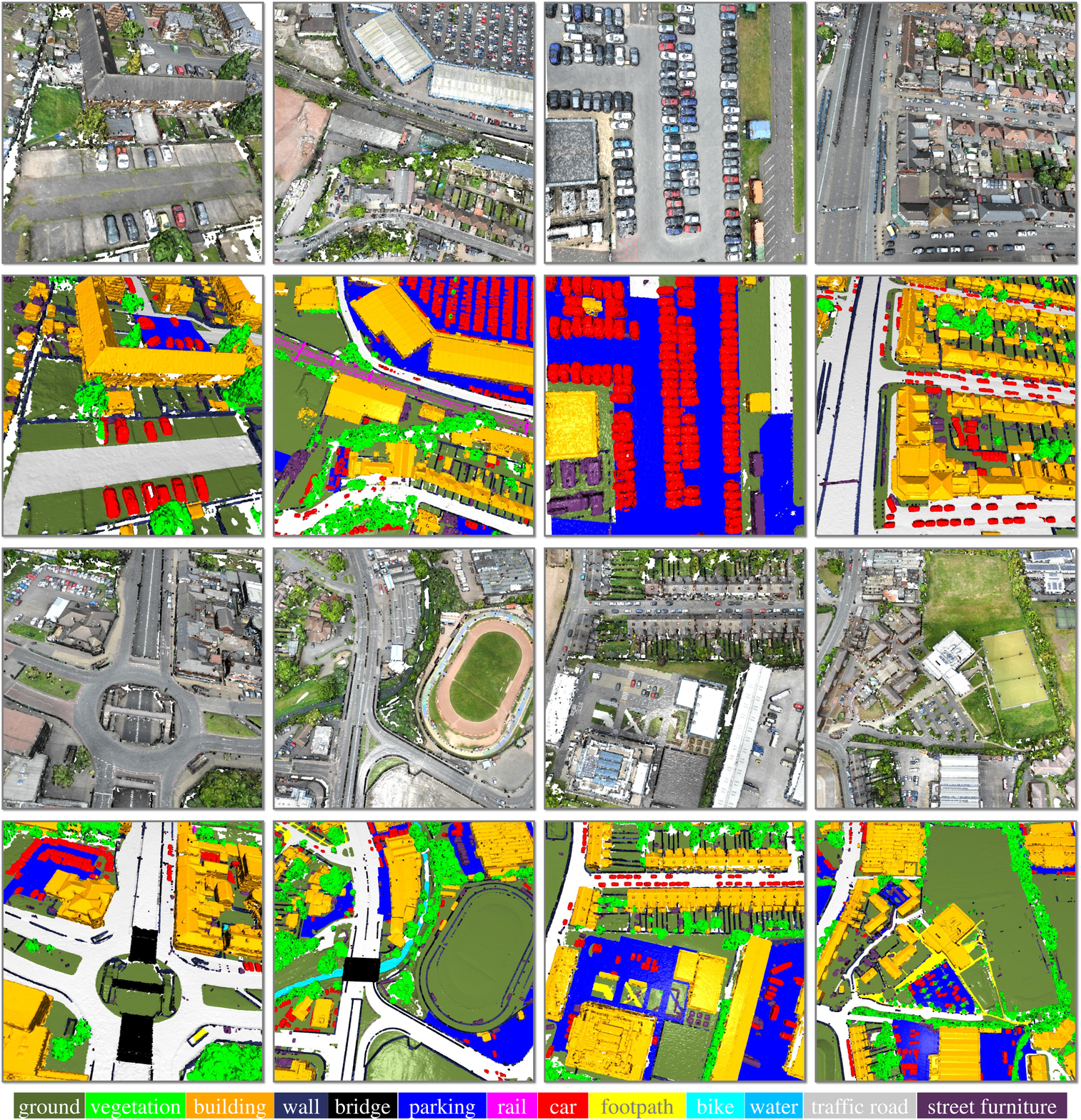}}
	\label{subfig:dat2-city}
	\caption{Experimental Point Cloud Datasets}
	\label{fig:experimentsdataset}
\end{figure}

We tested our improved DO on two dataset (shown in Fig.\ref{fig:experimentsdataset}): i) 3D synthetic dataset \cite{stanfordbunny}; ii) 3D real urban LIDAR dataset \cite{Qingyong:2021dataset}.
\subsubsection{Synthetic dataset}
Similar to the original DO paper, the synthetic experiments we used is the Stanford Bunny dataset, which is consist of 35947 points. We utilized the off-the-shelf function $pcdownsample$ in MATLAB to averagely sample 514 points as the model point cloud $\bf M$. For the sake of highlight the competitiveness of our improved DO, we increased the difficulty level on all testing samples. We designed six types of perturbations on the dataset, and they are noise, point number, outlier, incompleteness, rotation, and translation. Especially, the translation perturbation is a basic but indispensable case, which is absent in the DO paper. For each perturbation, one of the following setting is applied to the model set to create a scene point cloud. When we perturbed one type, the rest types are set to our default values in the curly brace rather than the DO's in the square bracket. Each perturbation level generated 100 testing samples.
\begin{enumerate}
	\item Noise Standard Deviation (NoiseSTD): the noise standard deviation of all scene points ranges between 0 to 0.1; {our default = 0.05}, [DO's default = 0];
	\item Scene Point Number (SceneNumber): the number of all scene points ranges from 100 to 4000; {our default = 400}, [DO's default = 200 to 600];
	\item Outlier Number (Outlier): the outlier number ranges from 0 to 600; {our default = 300},[DO's default = 0]; 
	\item Incomplete Ratio (Incomplete): the incomplete ratio ranges from 0 to 0.7; {our default = 0.3}, [DO's default = 0].
	\item Rotation Angle (Rotation): the initial rotation angle ranges from 0 to 180 degrees (rotated along a random 3D vector); {our default = 60}, [DO's default = 0]; 
	\item Translation (Translation): the initial translation ranges from 0 to 1.0; {our default = 0.3}, [DO's default = 0].  
\end{enumerate}
\subsubsection{Real dataset}
\textcolor{black}{We tested our algorithm on the latest LIDAR dataset \cite{Qingyong:2021dataset}, which is an urban-scale photogrammetric 3D point clouds dataset scanned from Birmingham, Cambridge, and York. This dataset includes three billion annotated 3D points that covers 7.6 $km^2$ of the city landscape. We use the Birmingham 3D LIDAR point cloud as our experimental dataset. In contrast to the synthetic dataset experiment which added a sparse outliers in testing samples, here, we added structured outliers on the real dataset instead of the sparse outliers and the rest of the. We cropped part of the LIDAR point cloud and averagely sample a point cloud as model point. We also individually generated 100 testing samples for each perturbation level.}
\subsection{Training and Evaluation Criteria}
In training phase, we generated 30000 training samples, each training sample is generated as the following perturbation settings: 1) Random Noise Standard Deviation: 0 to 0.05, 2) Random Scene Point Number: 400 to 800, 3) Random Outliers Number: 0 to 300, 4) Random Incomplete Ratio: 0 to 0.3, 5) Random Rotation Angle: 0 to 90 (degree), and 6) Random Translation: 0 to 0.3. Comparing with the perturbation of testing samples, the perturbation level of training sample is more slight. It is the advantage of the DO-based algorithm than the deep neural network based algorithm. We trained 30 maps: ${\bf D}_1, {\bf D}_2, ..., {\bf D}_{30}$. The trade-off parameter $\lambda$ is set to 0.0002, and we set the Gaussian Deviation $\sigma ^2$ to 0.03.

\begin{figure}[htpb]
	\centering
	\subfloat[Training Error (OURS)]{\includegraphics[width=0.45\linewidth]{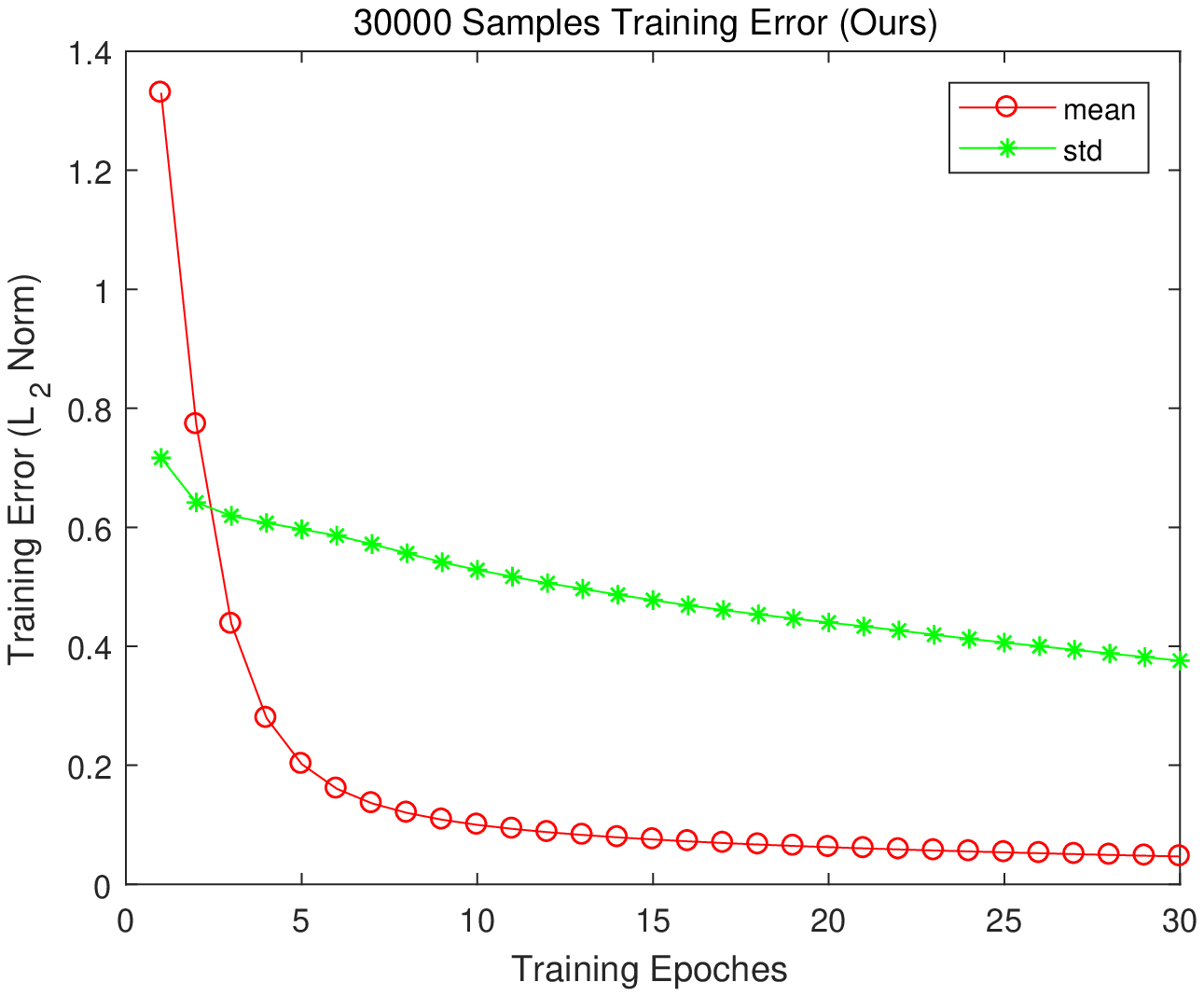}}
	\label{subfig:ours}
	\hspace{1em}
	\subfloat[Training Error (DO)]{\includegraphics[width=0.45\linewidth]{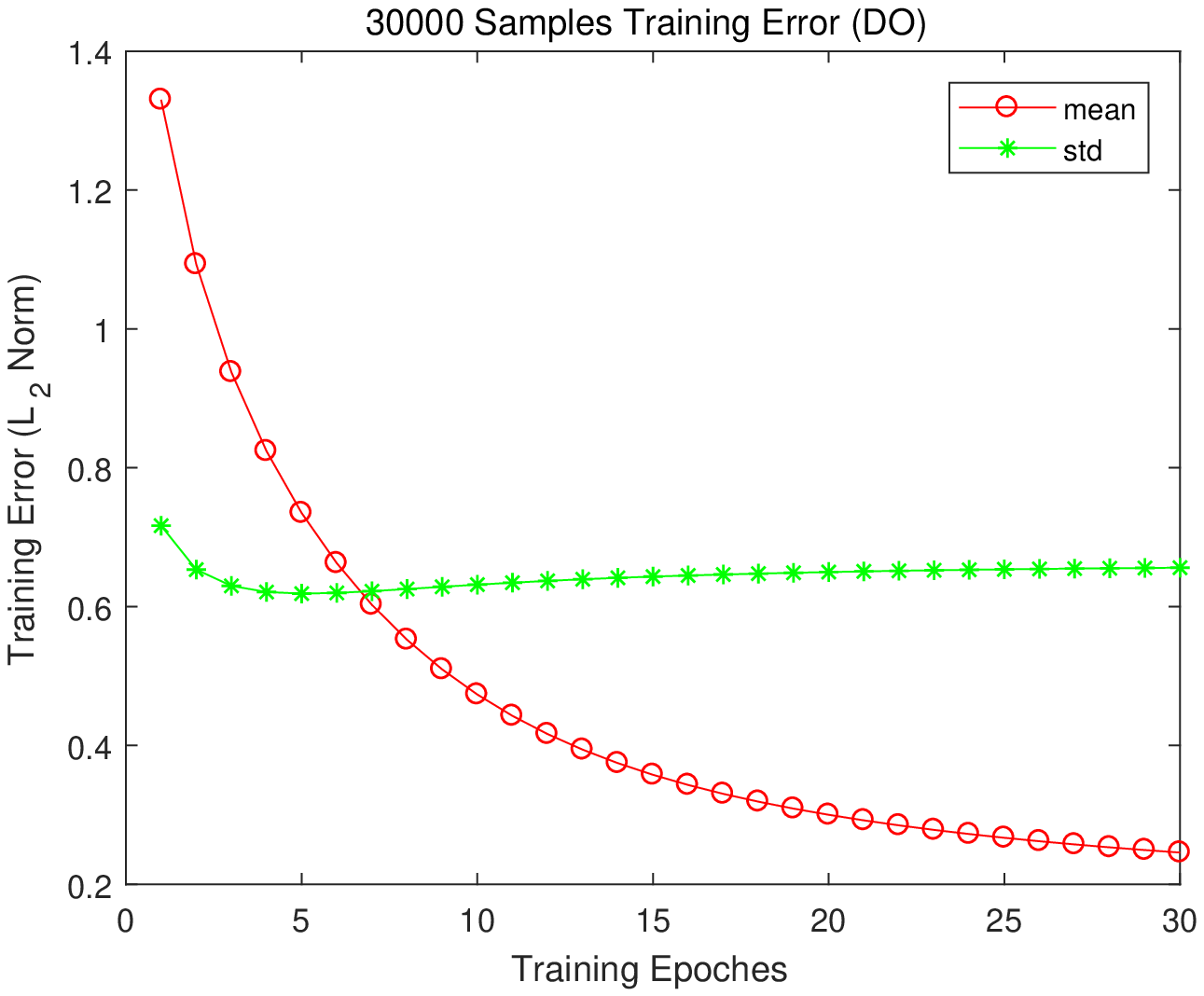}}
	\label{subfig:DObase}
	
	\subfloat[Rotation Error (OURS)]{\includegraphics[width=0.45\linewidth]{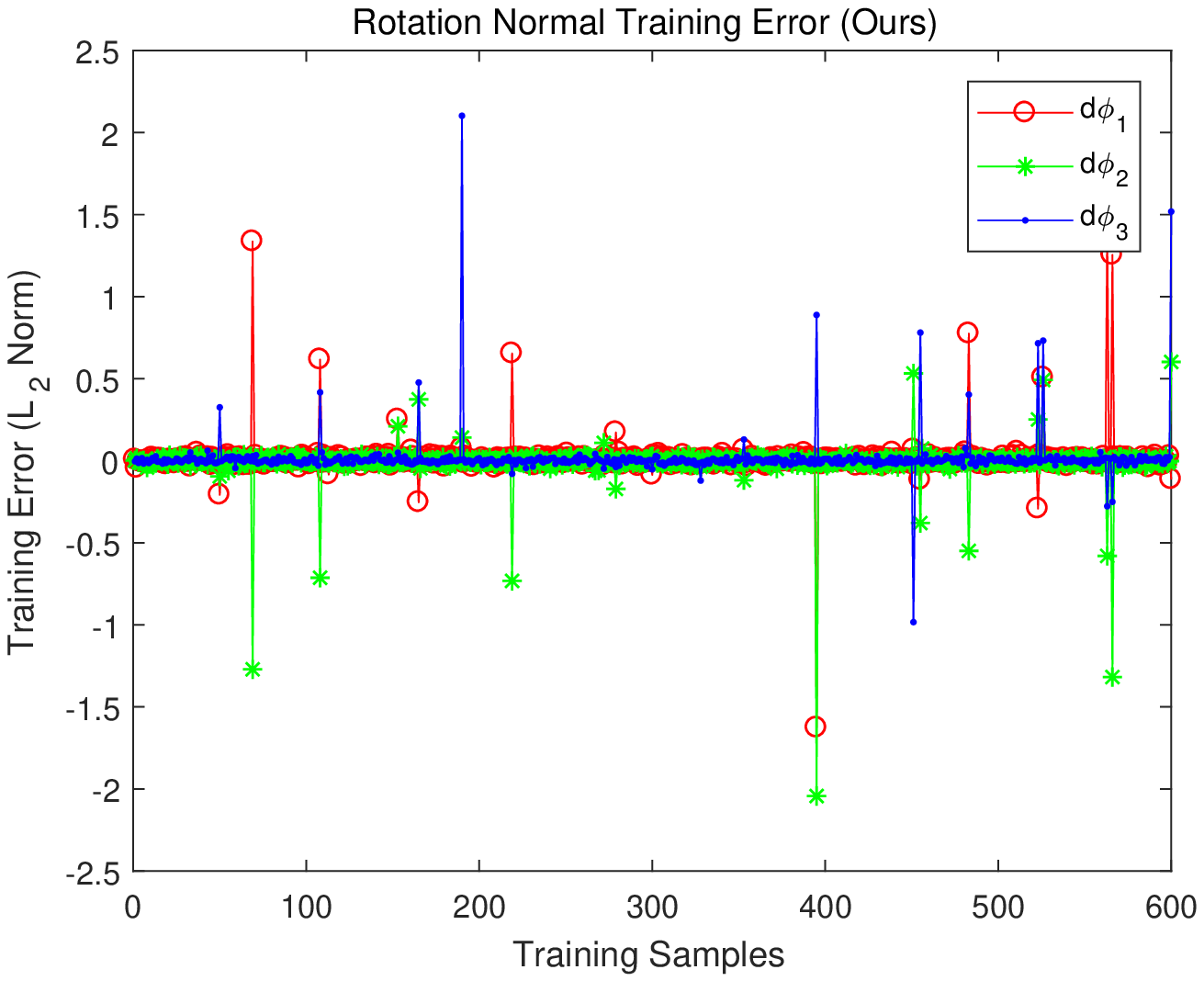}}
	\label{subfig:ours_rotationserrors}
	\hspace{1em}
	\subfloat[Rotation Error (DO)]{\includegraphics[width=0.45\linewidth]{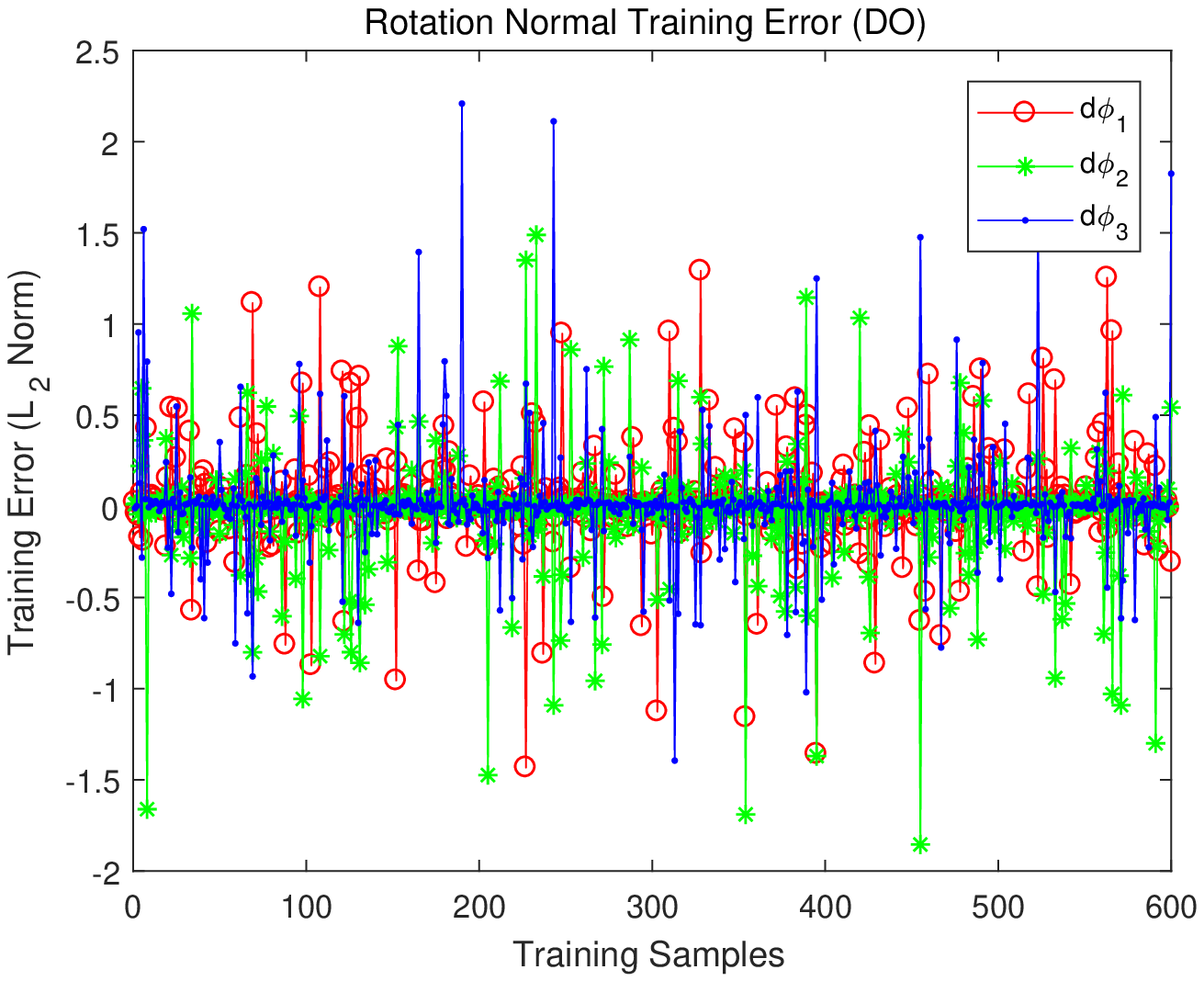}}
	\label{subfig:DObase_rotation}
	
	\subfloat[Translation Error (OURS)]{\includegraphics[width=0.45\linewidth]{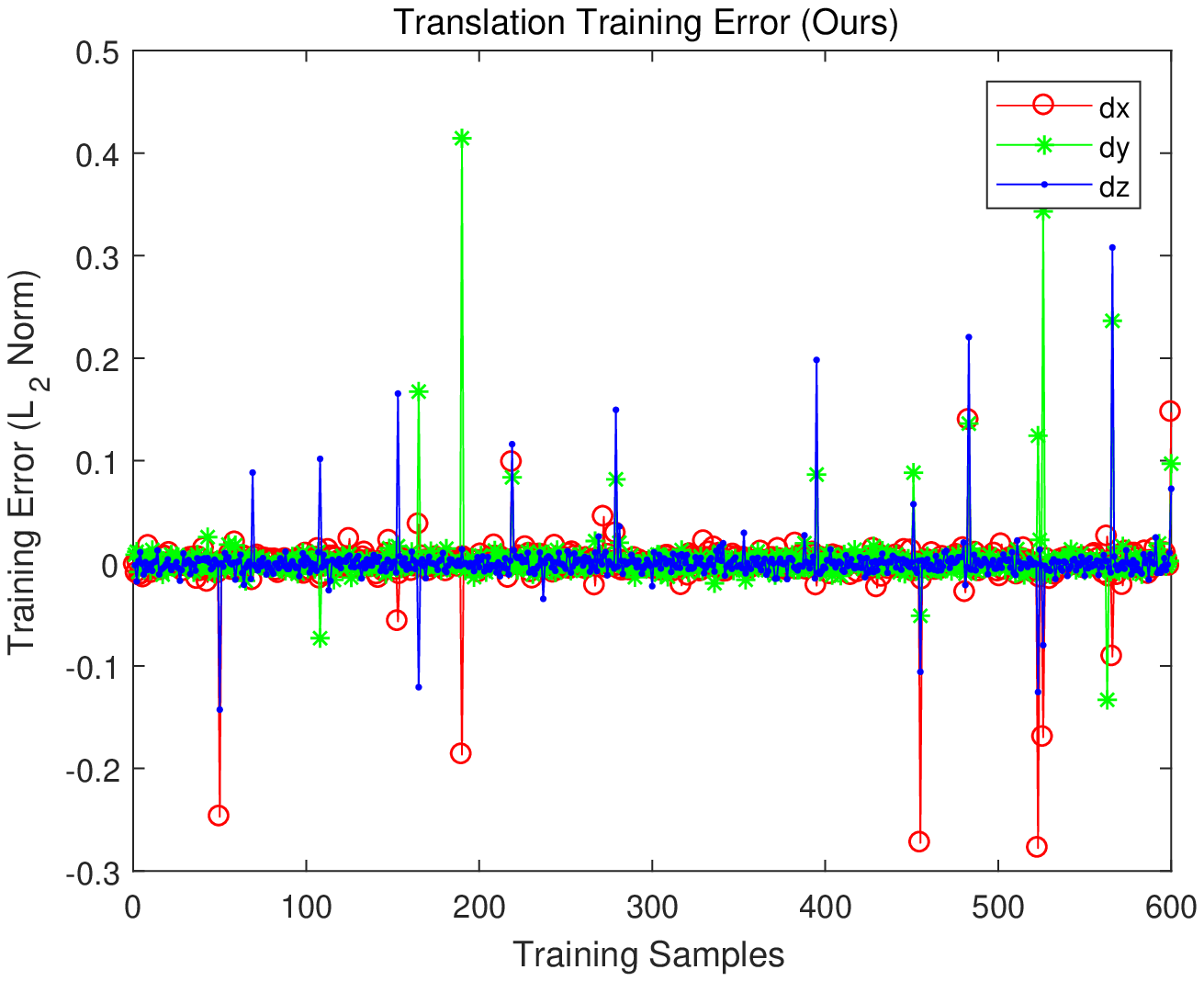}}
	\label{subfig:ours_translationerrors}
	\hspace{1em}
	\subfloat[Translation Error (DO)]{\includegraphics[width=0.45\linewidth]{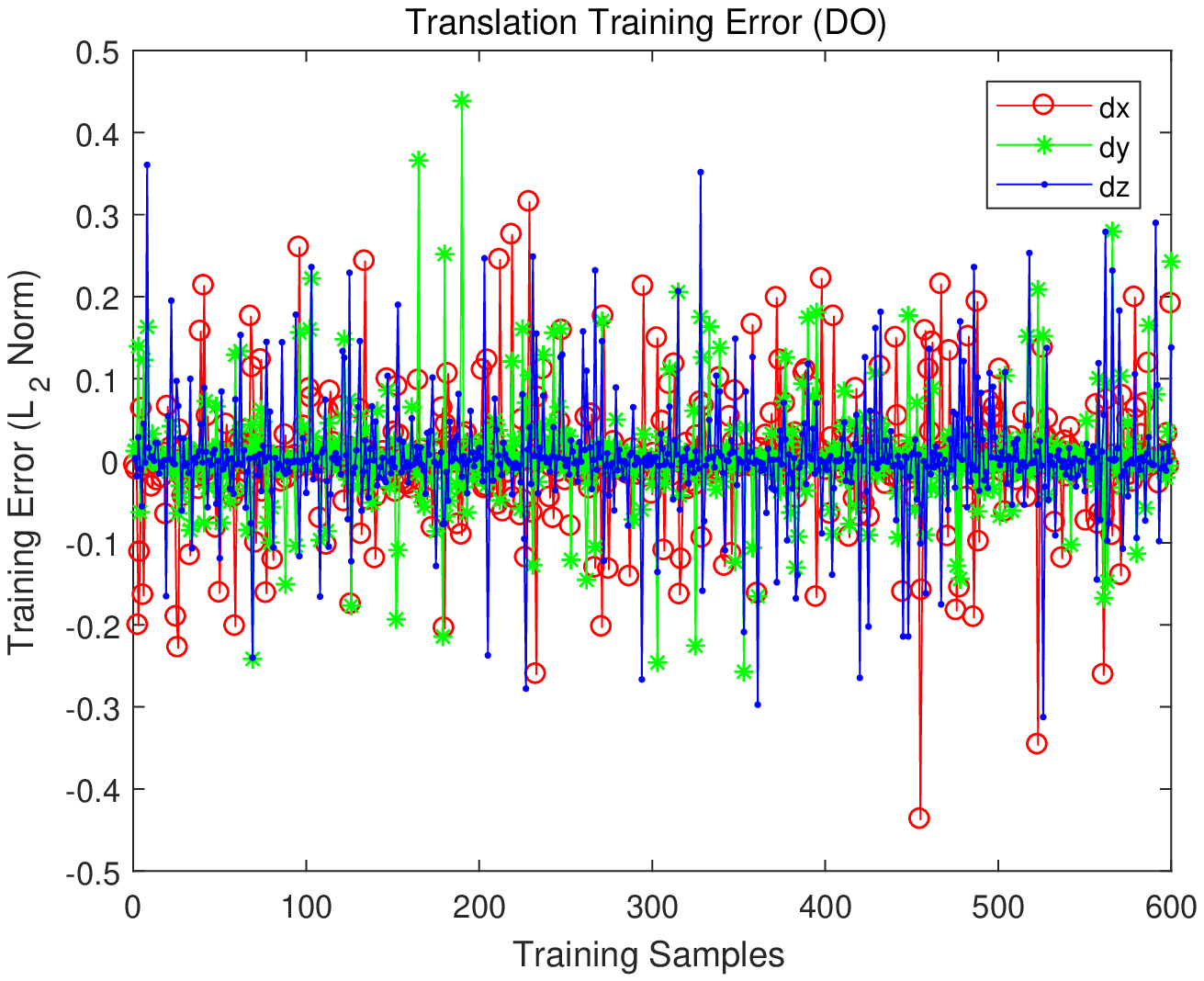}}
	\label{subfig:DObase_translation}
	\caption{Training Curves.}
	\label{fig:training curves}
\end{figure}

We compared the training error curves of our improved DO with that of the original DO. Both of the mean error and standard deviation on 30000 training samples are depicted in Fig.\ref{fig:training curves}.(a) and Fig.\ref{fig:training curves}.(b). We can see that the training loss of our algorithm decreased faster than the original DO, and achieved a high training precision. The training error's standard deviation decrease in all training epoches indicates that our improved DO is superior to the DO in learning. We randomly select 600 training samples and display their parameters regression error (depicted in Fig.\ref{fig:training curves}.(c) and Fig.\ref{fig:training curves}.(e)), while the training error on the same samples regressed by the DO are depicted in Fig.\ref{fig:training curves}.(d) and Fig.\ref{fig:training curves}.(f). The figures demonstrate that most of the parameters are predicted precisely in our improved DO but roughly in the original DO algorithm.

\begin{figure}[htpb]
	\centering
	\subfloat[Map 1 (DO)]{\includegraphics[width=0.45\linewidth]{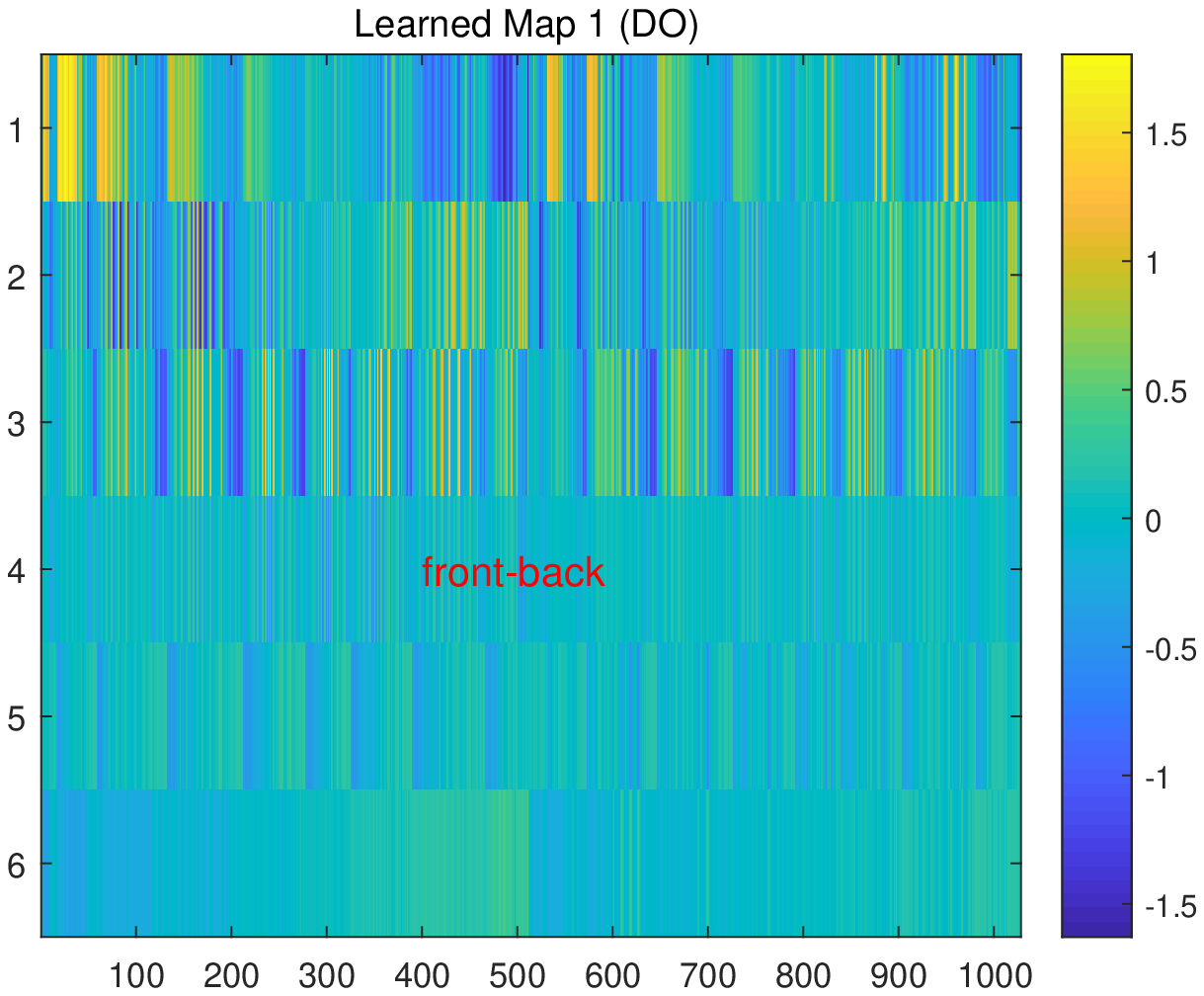}}
	\label{subfig:fig_dmap1}
	\hspace{1em}
	\subfloat[Map 1 (OURS)]{\includegraphics[width=0.45\linewidth]{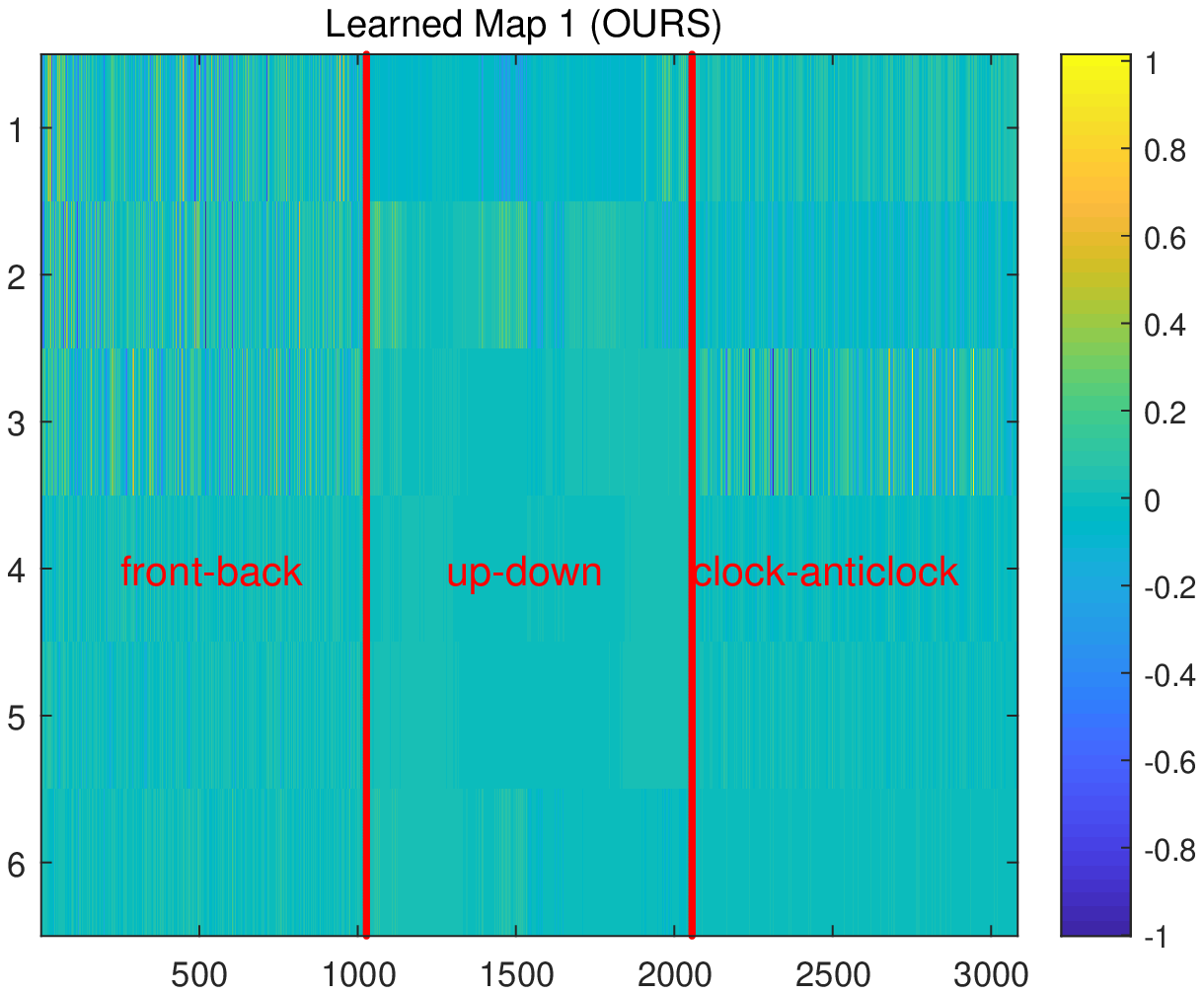}}
	\label{subfig:fig_first_itera}
	
	\subfloat[Map 15 (DO)]{\includegraphics[width=0.45\linewidth]{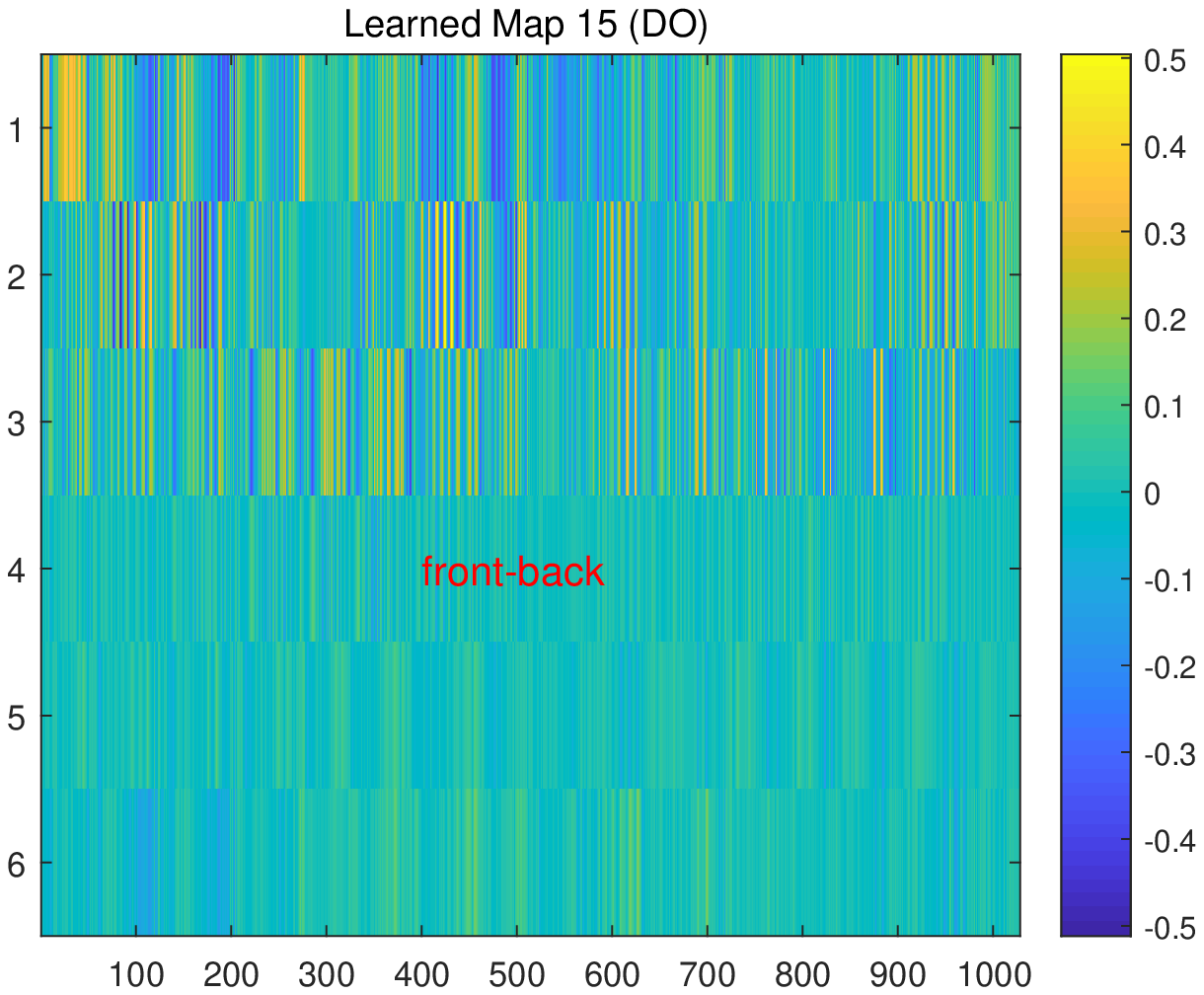}}
	\label{subfig:fig_dmap15}
	\hspace{1em}
	\subfloat[Map 15 (OURS)]{\includegraphics[width=0.45\linewidth]{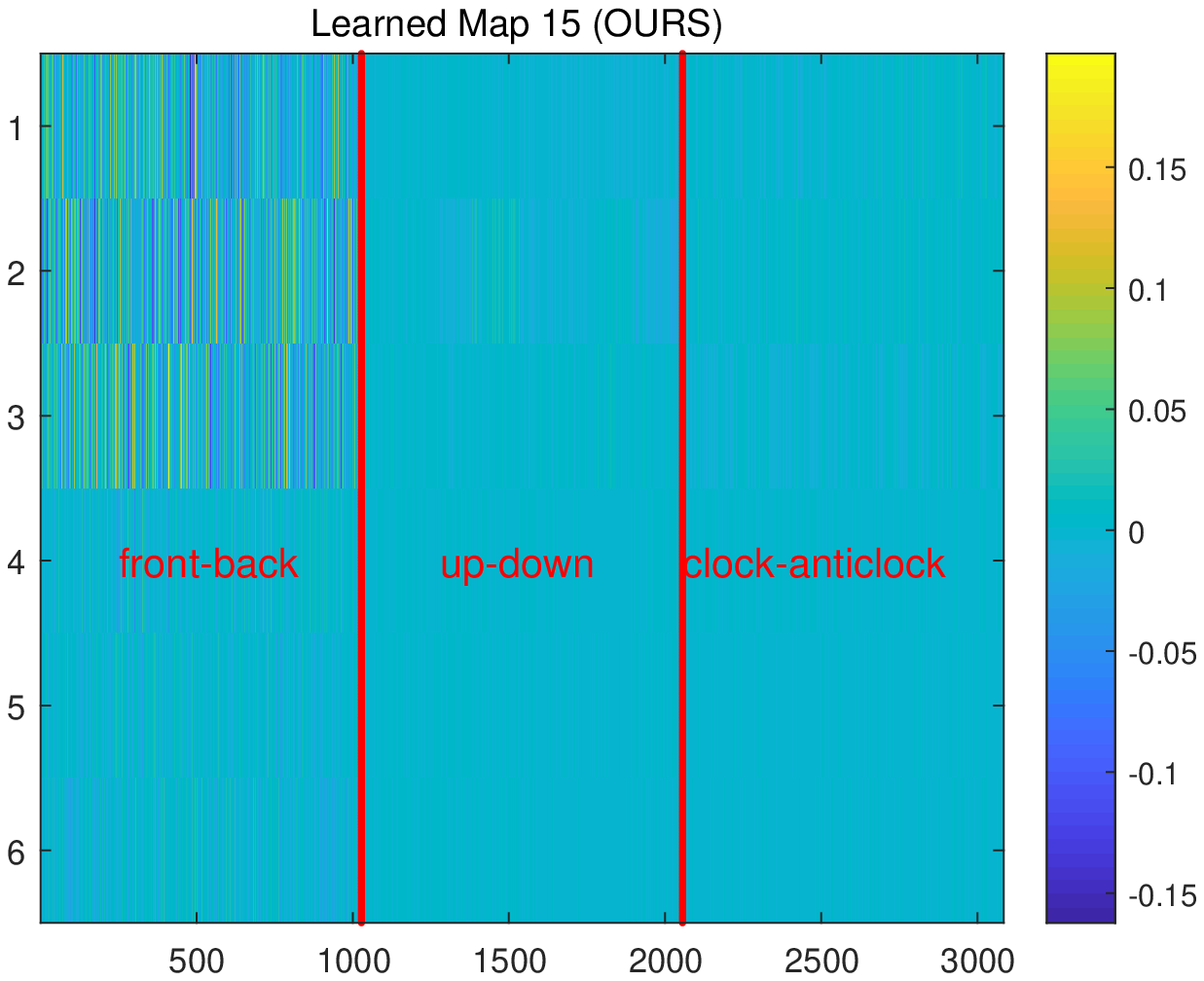}}
	\label{subfig:fig_mid_itera}
	
	\subfloat[Map 30 (DO)]{\includegraphics[width=0.45\linewidth]{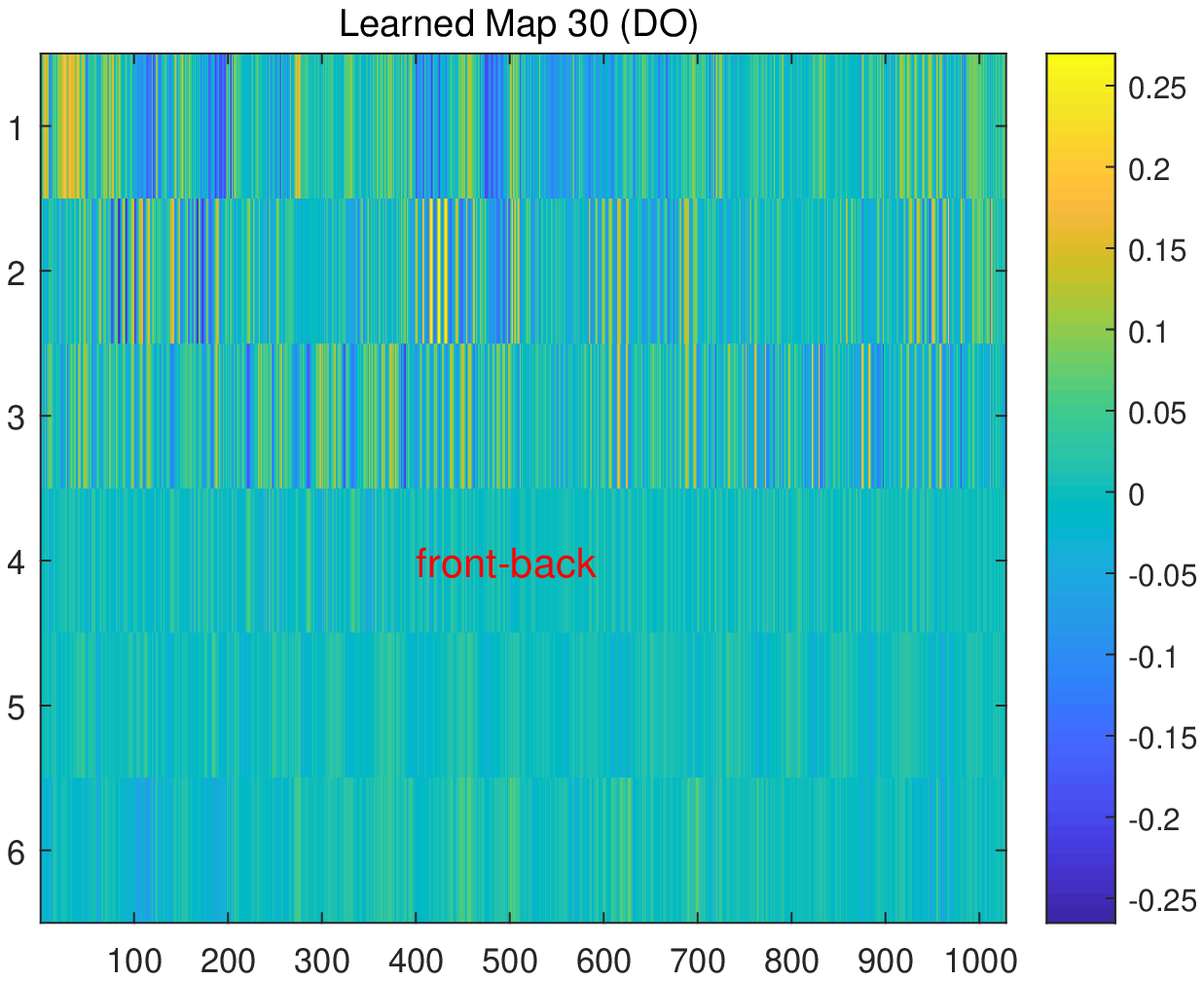}}
	\label{subfig:fig_dmap30}
	\hspace{1em}
	\subfloat[Map 30 (OURS)]{\includegraphics[width=0.45\linewidth]{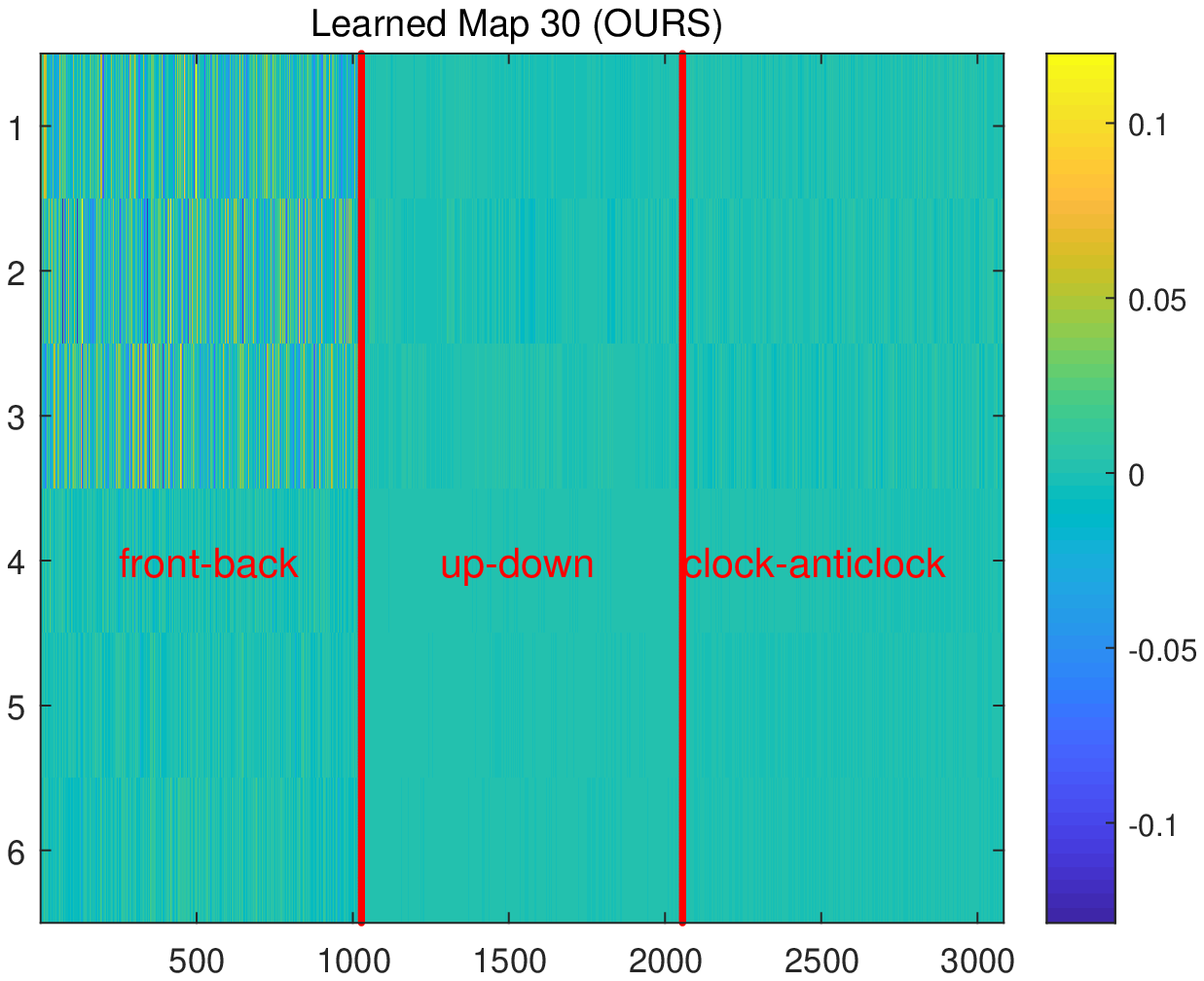}}
	\label{subfig:fig_final_itera}
	\caption{Comparison of the Learned Maps.}
	\label{fig:bothmaps}
\end{figure}

\begin{figure}[!t]
	\centering
	\subfloat[Iteration 1 (DO)]{\includegraphics[width=0.45\linewidth,height=0.45\linewidth]{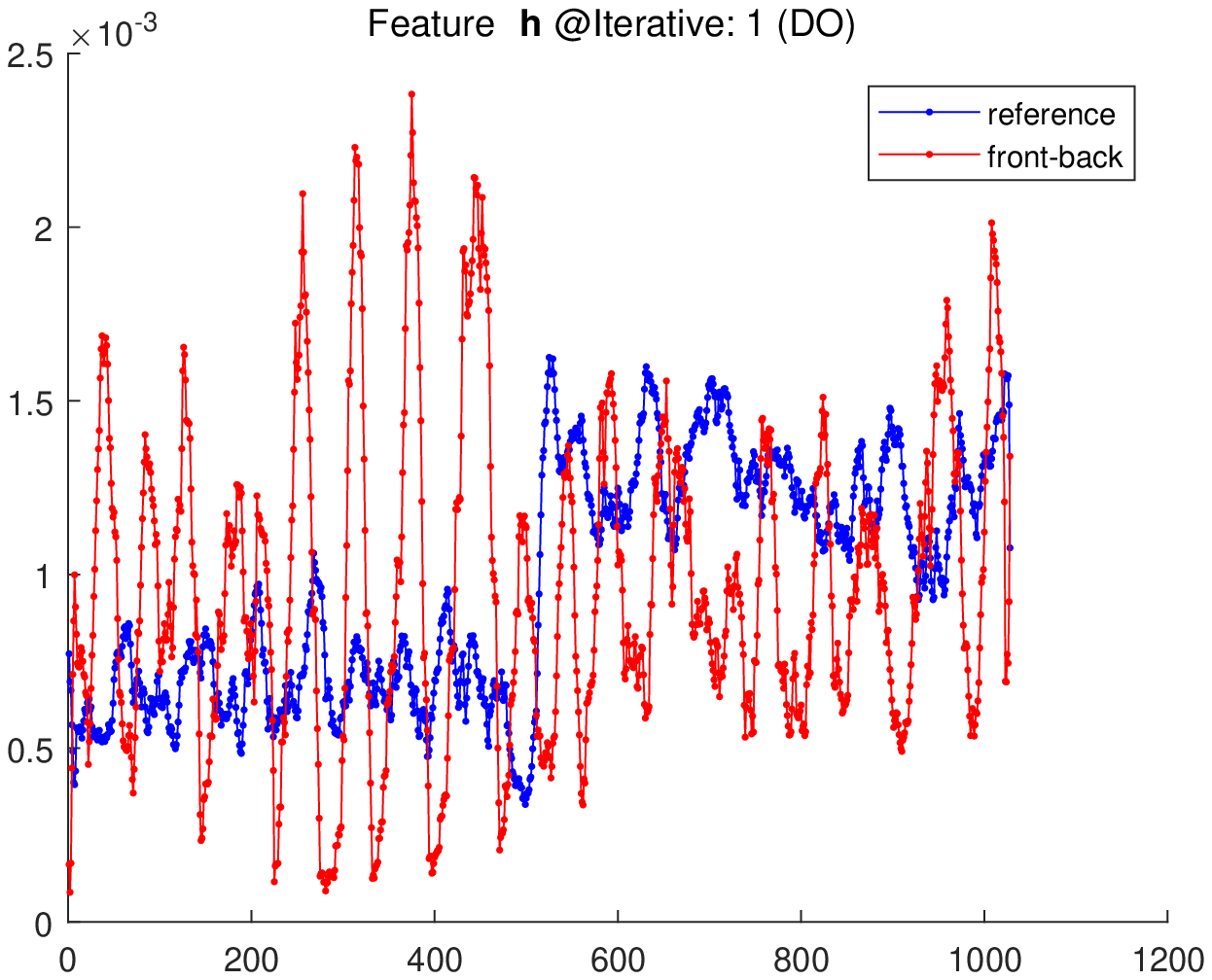}}
	\label{subfig:doh1}
	\hspace{1em}
	\subfloat[Iteration 1 (OURS)]{\includegraphics[width=0.45\linewidth,height=0.45\linewidth]{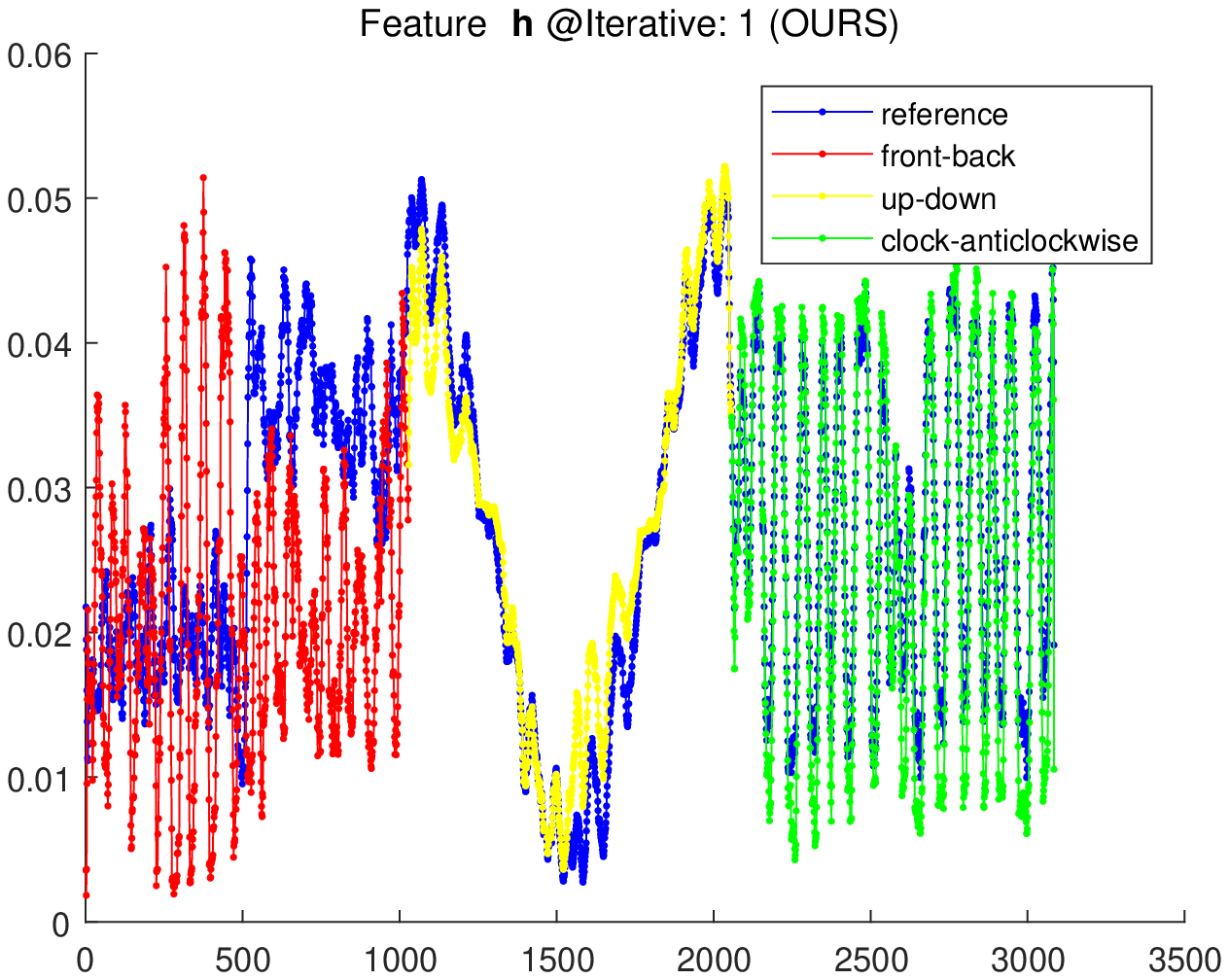}}
	\label{subfig:h1}
	
	\subfloat[Iteration 20 (DO)]{\includegraphics[width=0.45\linewidth,height=0.45\linewidth]{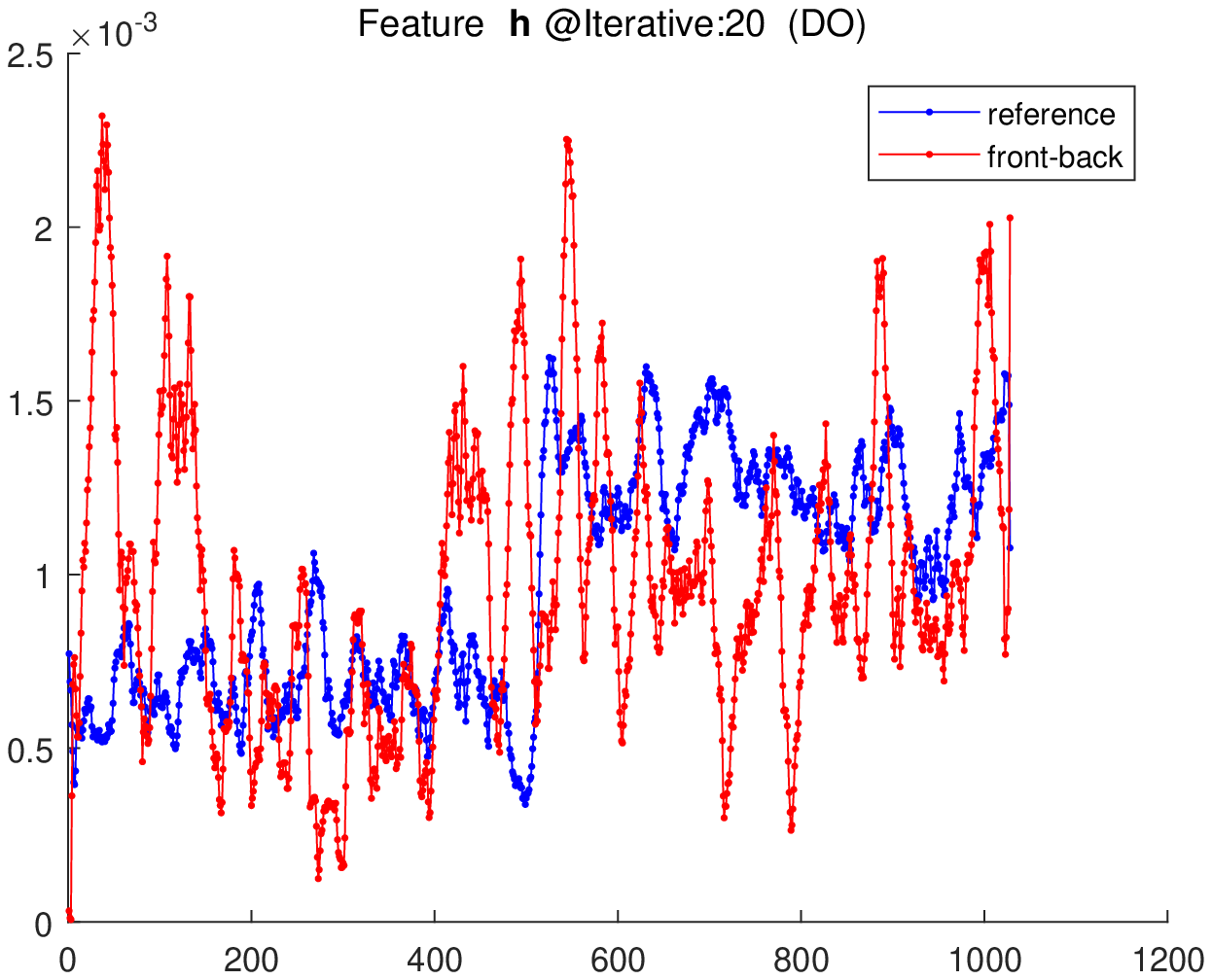}}
	\label{subfig:doh2}
	\hspace{1em}
	\subfloat[Iteration 10 (OURS)]{\includegraphics[width=0.45\linewidth,height=0.45\linewidth]{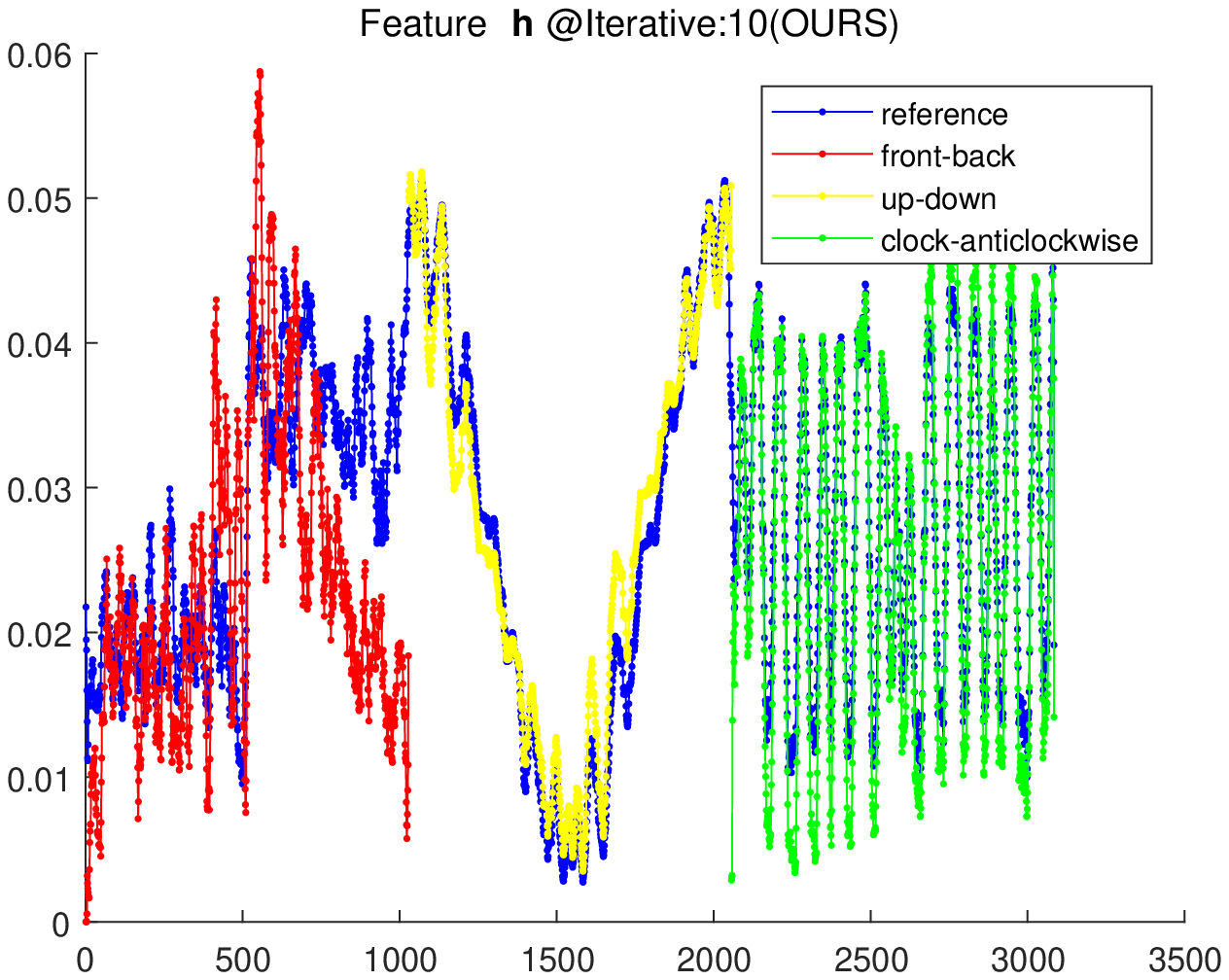}}
	\label{subfig:h2}
	
	\subfloat[Iteration 40 (DO)]{\includegraphics[width=0.45\linewidth,height=0.45\linewidth]{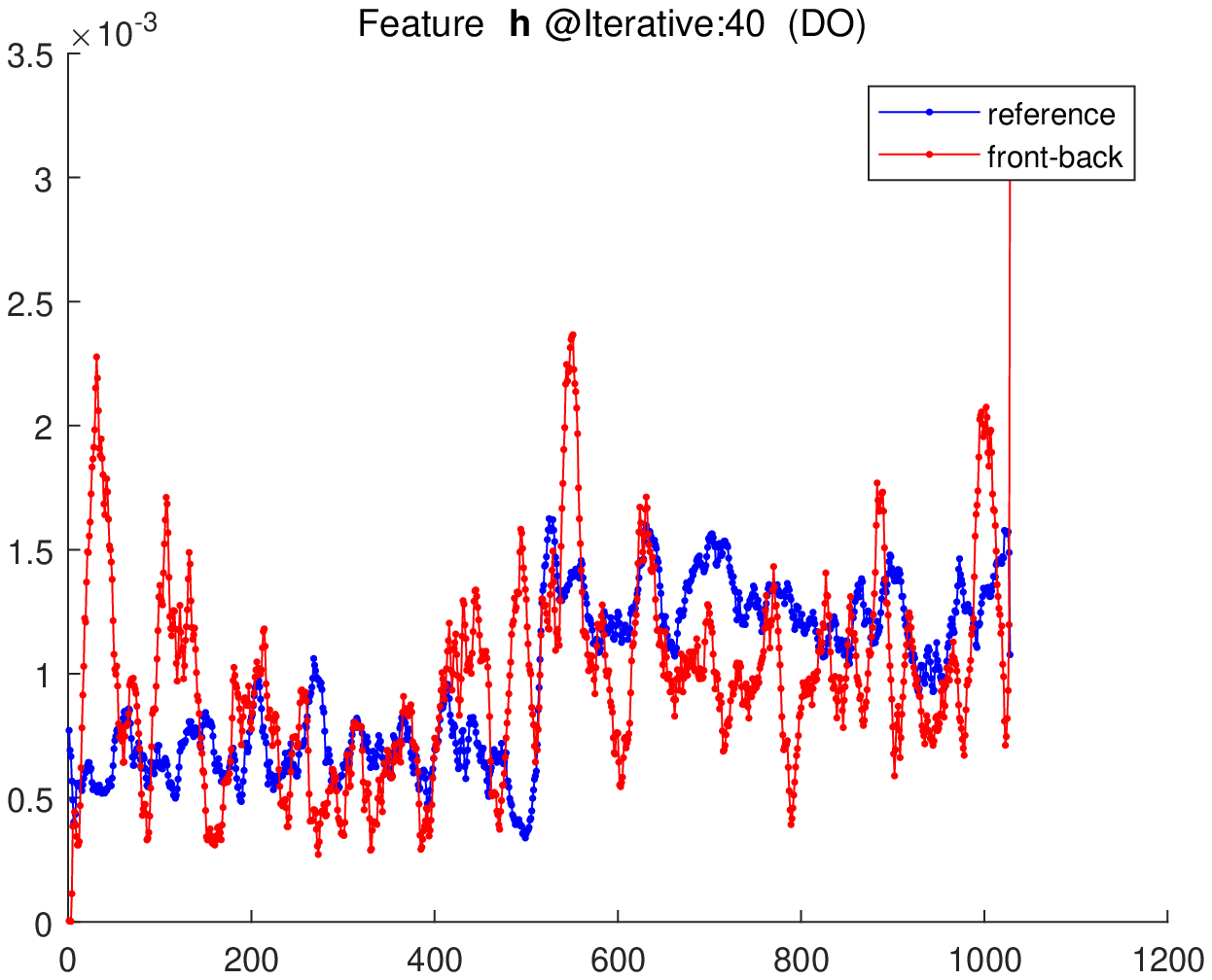}}
	\label{subfig:doh3}
	\hspace{1em}
	\subfloat[Iteration 30 (OURS)]{\includegraphics[width=0.45\linewidth,height=0.45\linewidth]{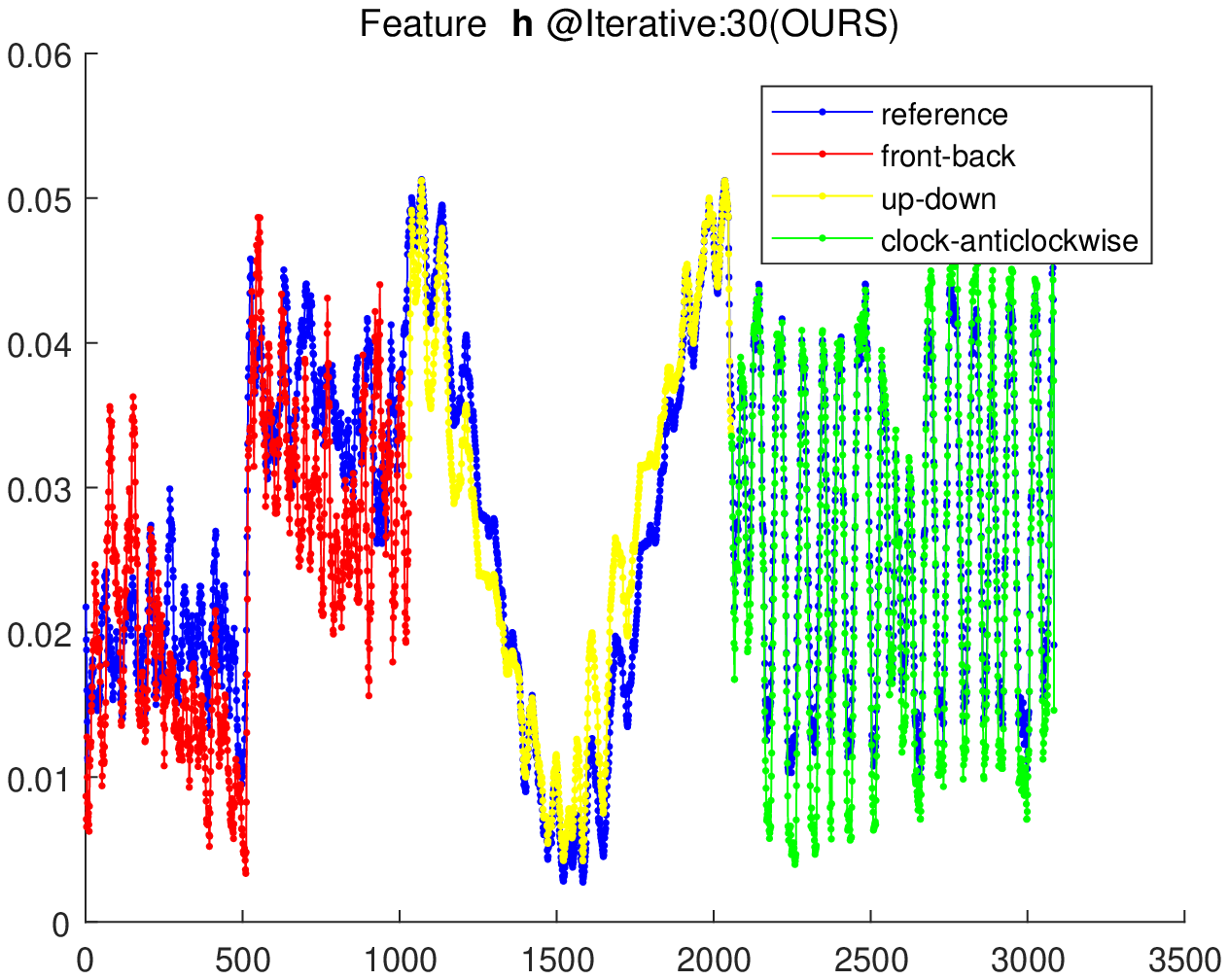}}
	\label{subfig:h3}
	
	\subfloat[Iteration 109(Final) (DO)]{\includegraphics[width=0.45\linewidth,height=0.45\linewidth]{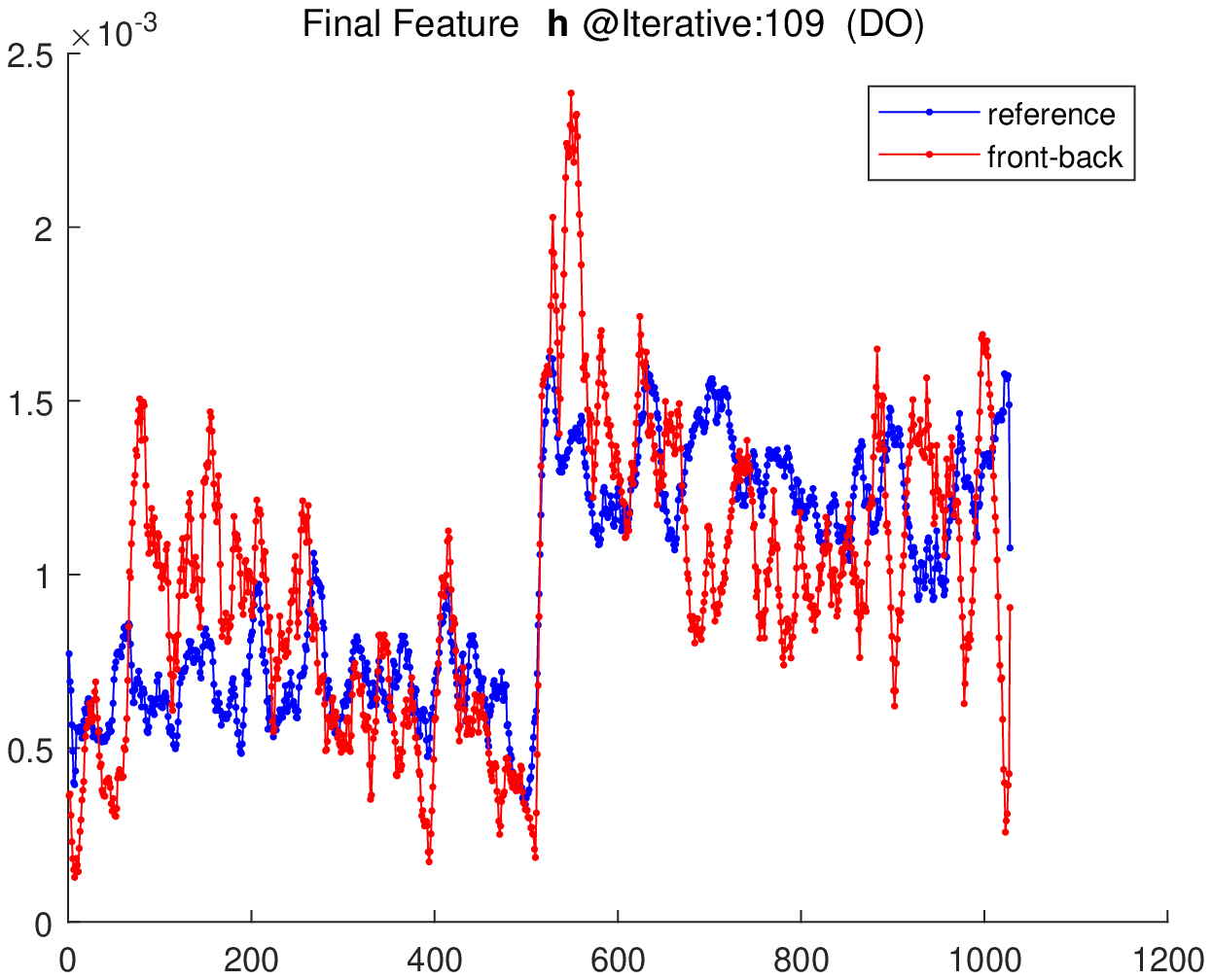}}
	\label{subfig:doh6}
	\hspace{1em}
	\subfloat[Iteration 71(Final) (OURS)]{\includegraphics[width=0.45\linewidth,height=0.45\linewidth]{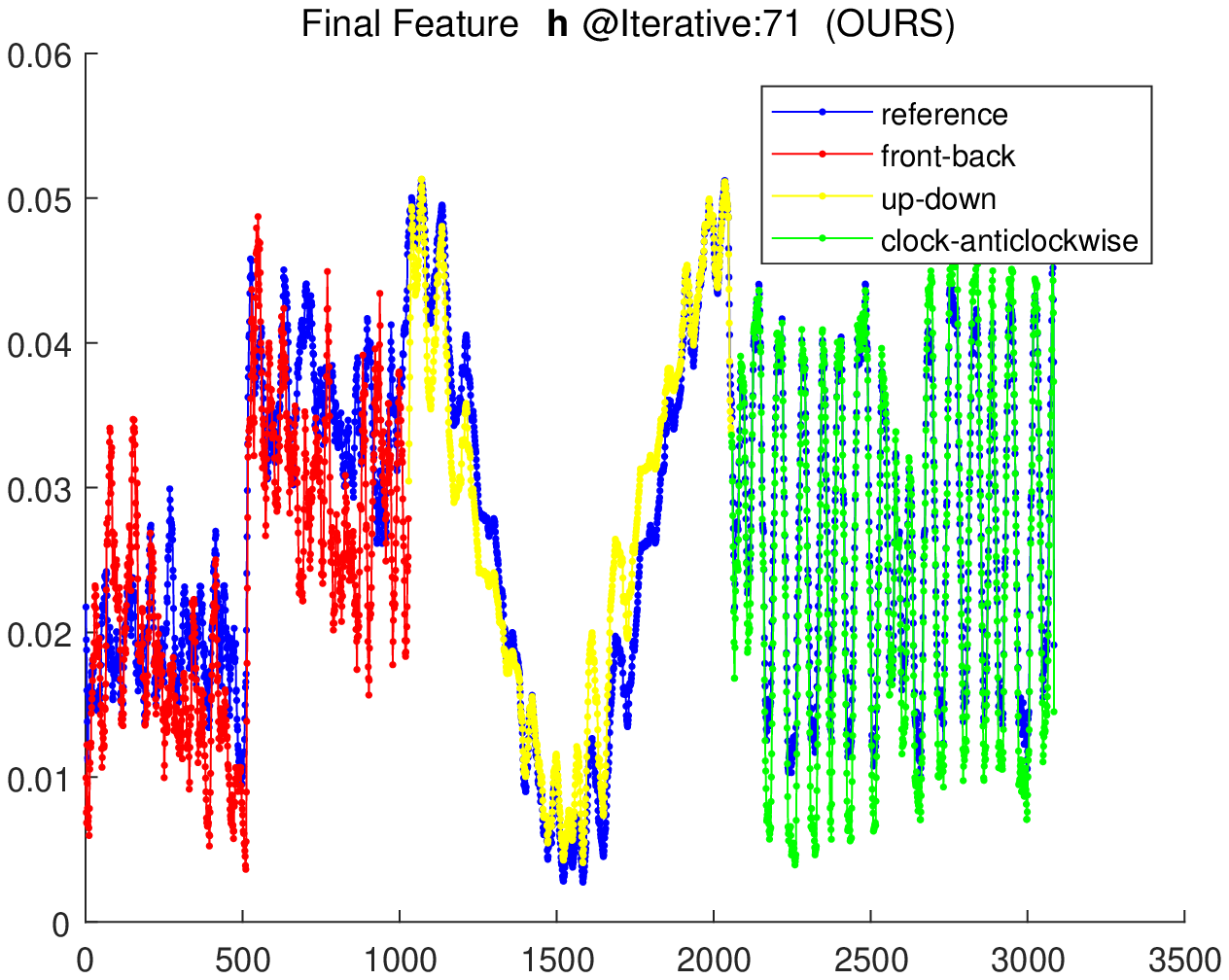}}
	\label{subfig:h6}

	\caption{Plots of Features.}
	\label{fig:bothfeatures}
\end{figure}

We compared the learned maps and histogram from the DO and our improved DO, they are illustrated in Fig.\ref{fig:bothmaps} and Fig.\ref{fig:bothfeatures}. The Fig.\ref{fig:bothmaps}(a,c,e) plot the DO's learned map $D_1$, $D_{15}$, and $D_{30}$ by MATLAB $imagesc$, the figures revealed that the original DO's map is not a sparse matrix. Similarly, our trained maps ploted in Fig.\ref{fig:bothmaps}(b,d,f) are not sparse matrix, which means the extended "up-down" and "clockwise-anticlockwise" sides are counted in the linear regression. Thus, by multiplicating the extended histogram, our map will collect more information to predict the registration parameters more precise and efficient than the original DO.

\begin{figure}[!t]
	\centering
	\subfloat[Initization (DO)]{\includegraphics[width=0.45\linewidth,height=0.45\linewidth]{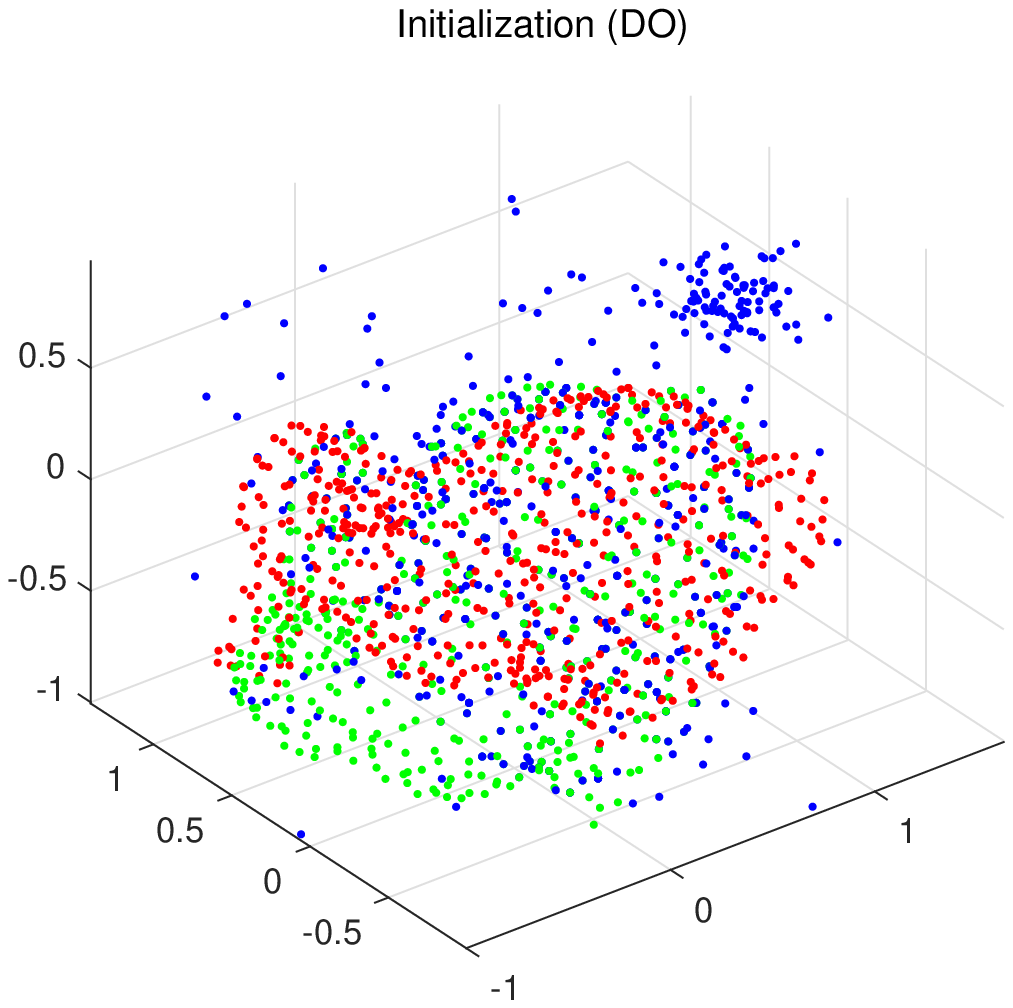}}
	\label{subfig:do1}
	\hspace{1em}
	\subfloat[Initization (OURS)]{\includegraphics[width=0.45\linewidth,height=0.45\linewidth]{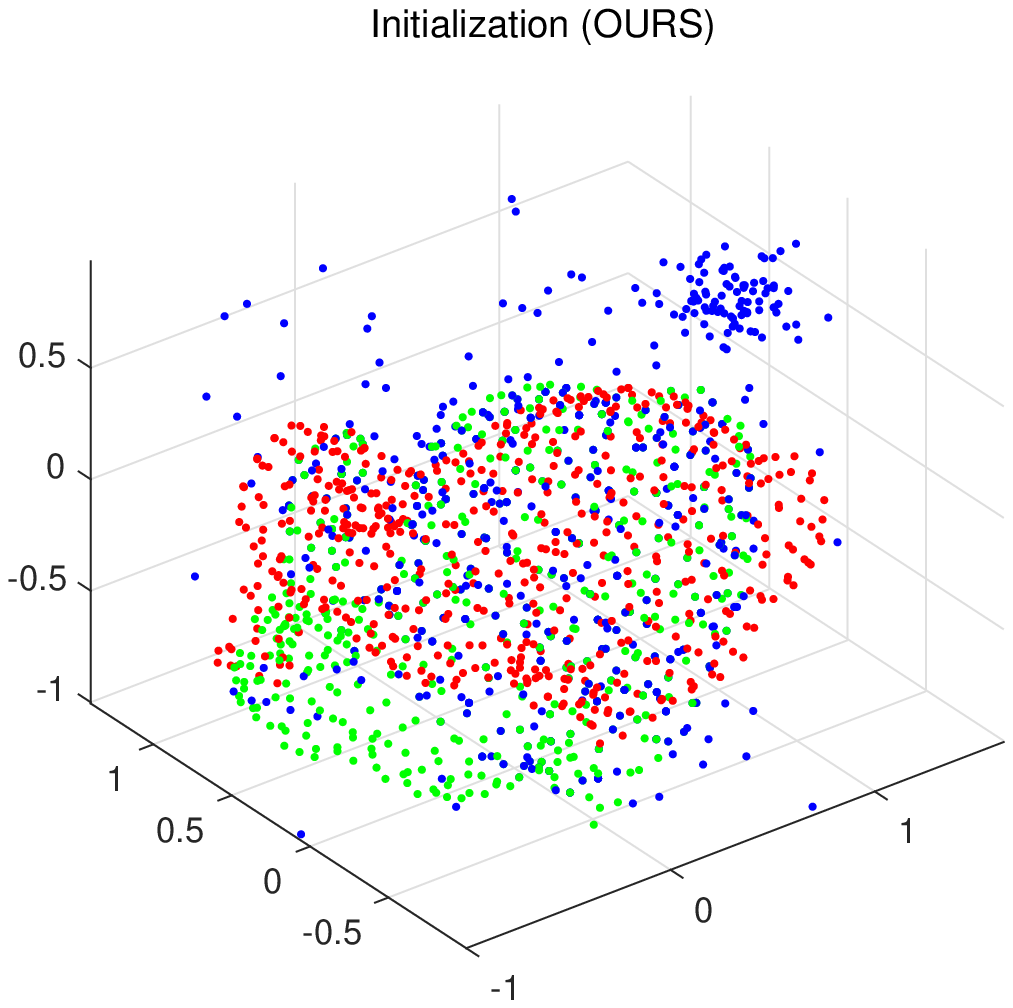}}
	\label{subfig:1}
	
	\subfloat[Iteration 30 (DO)]{\includegraphics[width=0.45\linewidth,height=0.45\linewidth]{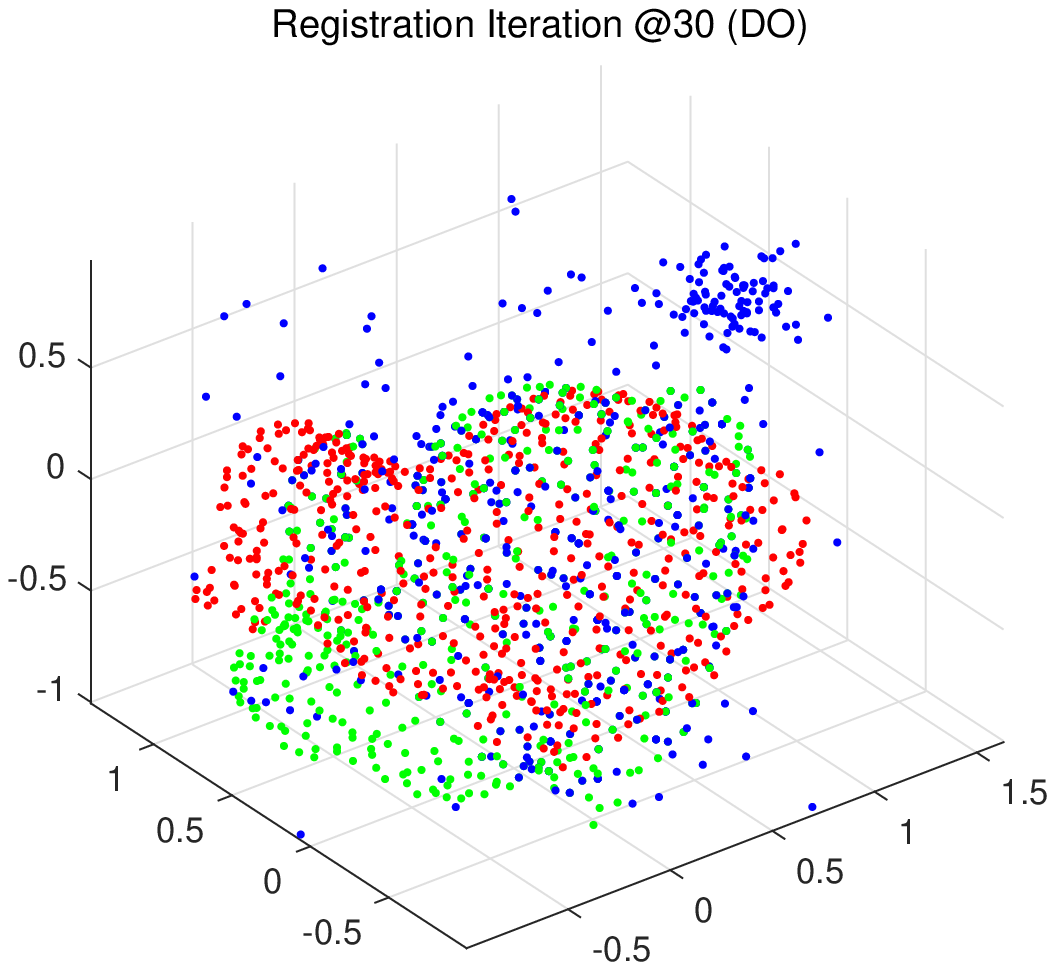}}
	\label{subfig:do3}
	\hspace{1em}
	\subfloat[Iteration 10 (OURS)]{\includegraphics[width=0.45\linewidth,height=0.45\linewidth]{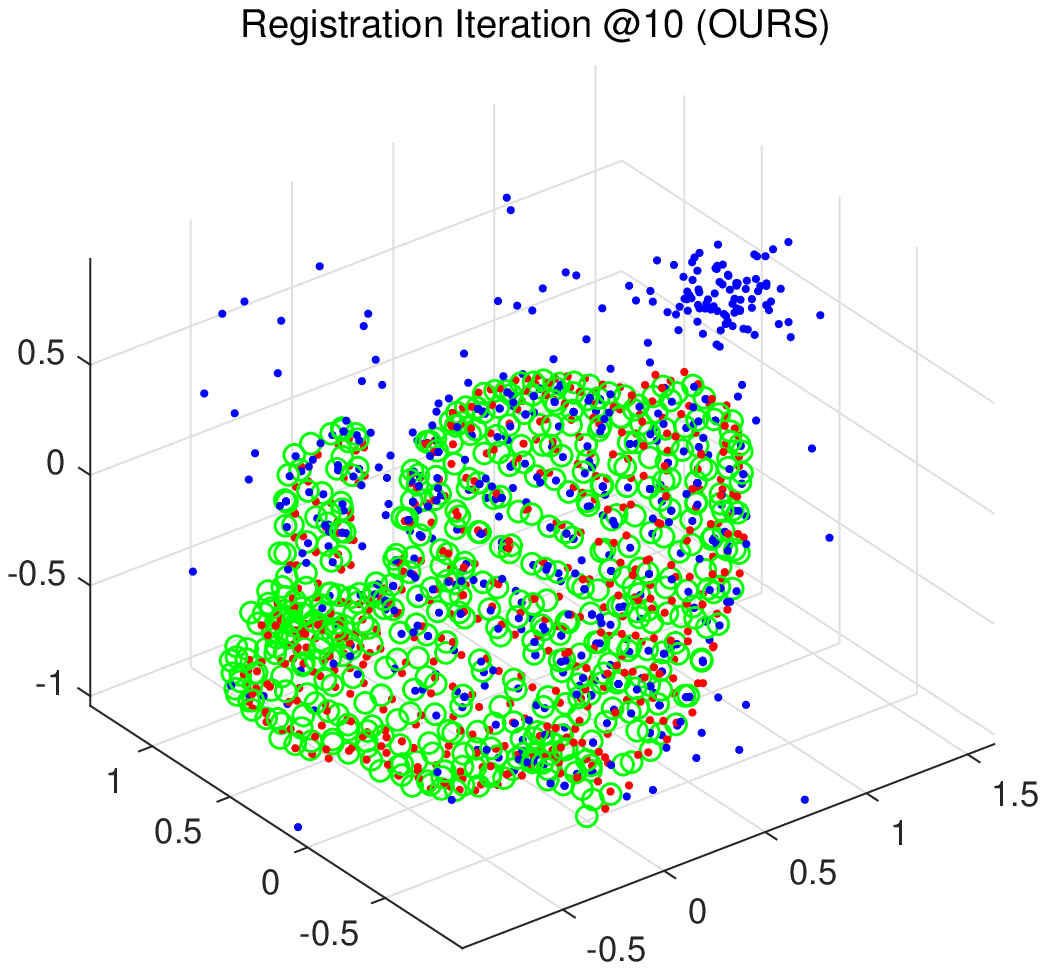}}
	\label{subfig:2}

	\subfloat[Iteration 121(Final) (DO)]{\includegraphics[width=0.45\linewidth,height=0.45\linewidth]{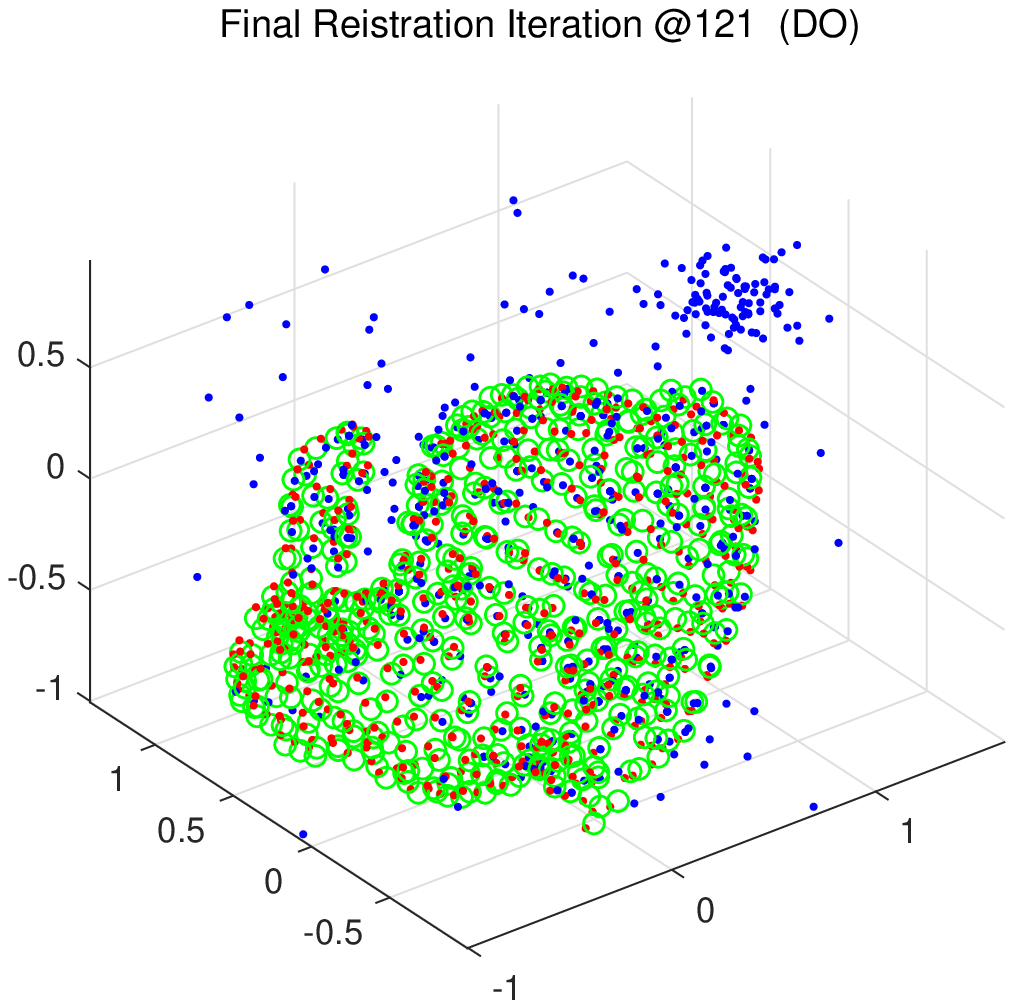}}
	\label{subfig:do5}
	\hspace{1em}
	\subfloat[Iteration 71(Final) (OURS)]{\includegraphics[width=0.45\linewidth,height=0.45\linewidth]{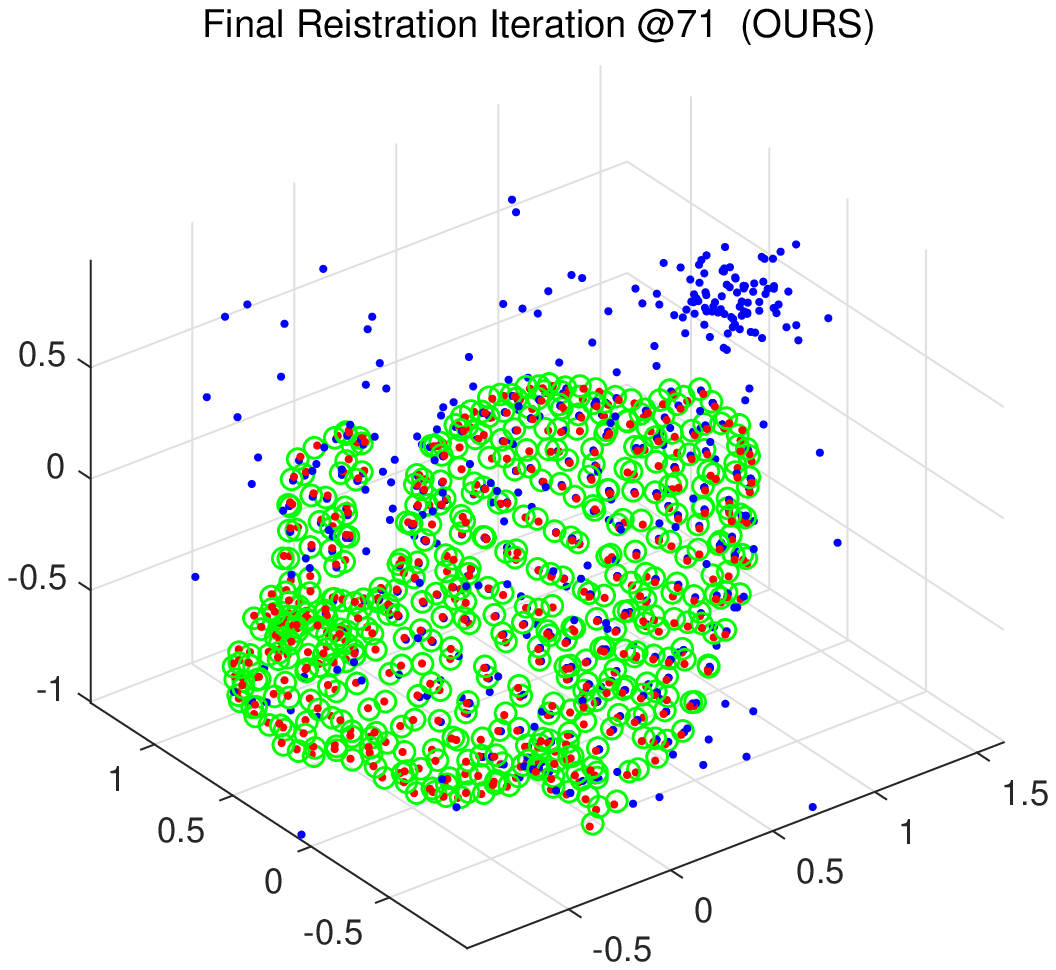}}
	\label{subfig:5}
	\caption{Experimental Result on Synthetic Dataset.}
	\label{fig:dat1-Iter-result}
\end{figure}

To illustrate the both of the histogram's evolution trend during the inferring steps, the histogram of DO and improved DO are ploted in two column in Fig.\ref{fig:dat1-Iter-result}. The first column is DO's histogram and the second column is ours. In order to compare them with a baseline, here, we generate a reference line which is the histogram that extracted from two identical model point cloud. The reference line is painted in blue dot line in Fig.\ref{fig:bothfeatures}. All of the testing scene point cloud's histogram will approach to the reference line, and none of them will be identical with the reference line because of the perturbation. But we can utilize it as a auxiliary criteria in registration. In the first column of Fig.\ref{fig:bothfeatures}, we can see that the histograms painted in red color represent the "front-back" sides histogram. When the inferring continues, they are approaching the reference-line. In the second column of Fig.\ref{fig:bothfeatures}, each figure contains three parts of histograms that represent the "front-back", "up-down", and "clockwise-anticlockwise" sides respectively. Because the "up-down" and "clockwise-anticlockwise" sides histogram are global feature descriptor (i.e. a statistical feature), so that the changes are not apparent than the local feature descriptor "front-back" historgram. The point cloud pair we tested is coarsely registered at iteration 10 by our improved DO. Therefore the histogram from Fig.\ref{fig:dat1-Iter-result}(b) to Fig.\ref{fig:dat1-Iter-result}(d) clearly show the trends of "up-down" histogram approaching the reference-line. Fig.\ref{fig:dat1-Iter-result}(d), Fig.\ref{fig:dat1-Iter-result}(f), and Fig.\ref{fig:dat1-Iter-result}(h) shown the fine tuning progress. 
The evaluation criterias we utilized are: 1) Point Registration Accuracy (PointAcc), 2) Point Registration Root-Mean-Sqart-Error (PointRMSE). The PointAcc can be calculated as follows:
\begin{equation}
{\rm{Acc}}_{pt}  = {{{\rm{num}}(|{\bf{T}}({\bf{S}},{\bf{\hat x}}) - {\bf{S}}^ *  | < t_{pt} )} \mathord{\left/
		{\vphantom {{{\rm{num}}(|{\bf{T}}({\bf{S}},{\bf{\hat x}}) - {\bf{S}}^ *  | < t_{pt} )} {{\rm{num}}({\bf{S}}^ *  )}}} \right.
		\kern-\nulldelimiterspace} {{\rm{num}}({\bf{S}}^ *  )}}	
\end{equation}
Here, ${\rm{num}}( * )$ means the numbers when expression $*$ is true.  $t_{pt}=0.1$ is the threshold.
The PointRMSE can be calculated as follows:
\begin{equation}
{\rm{RMSE}}_{pt}  = {\rm{(}}\frac{1}{{N_S }}\sum\nolimits_{n = 1}^{N_S } {\left\| {{\bf{T}}(s_n ,{\bf{\hat x}}) - s_n^ *  } \right\|_2^2 } {\rm{)}}^{{1 \mathord{\left/
			{\vphantom {1 2}} \right.
			\kern-\nulldelimiterspace} 2}} 	
\end{equation}
Here, $s_n^ *$ is the ground truth of scene point.

\subsection{Testing Results and Analysis}
\begin{figure}[!t]
	\centering
	\subfloat[NoiseStd Acc]{\includegraphics[width=0.45\linewidth]{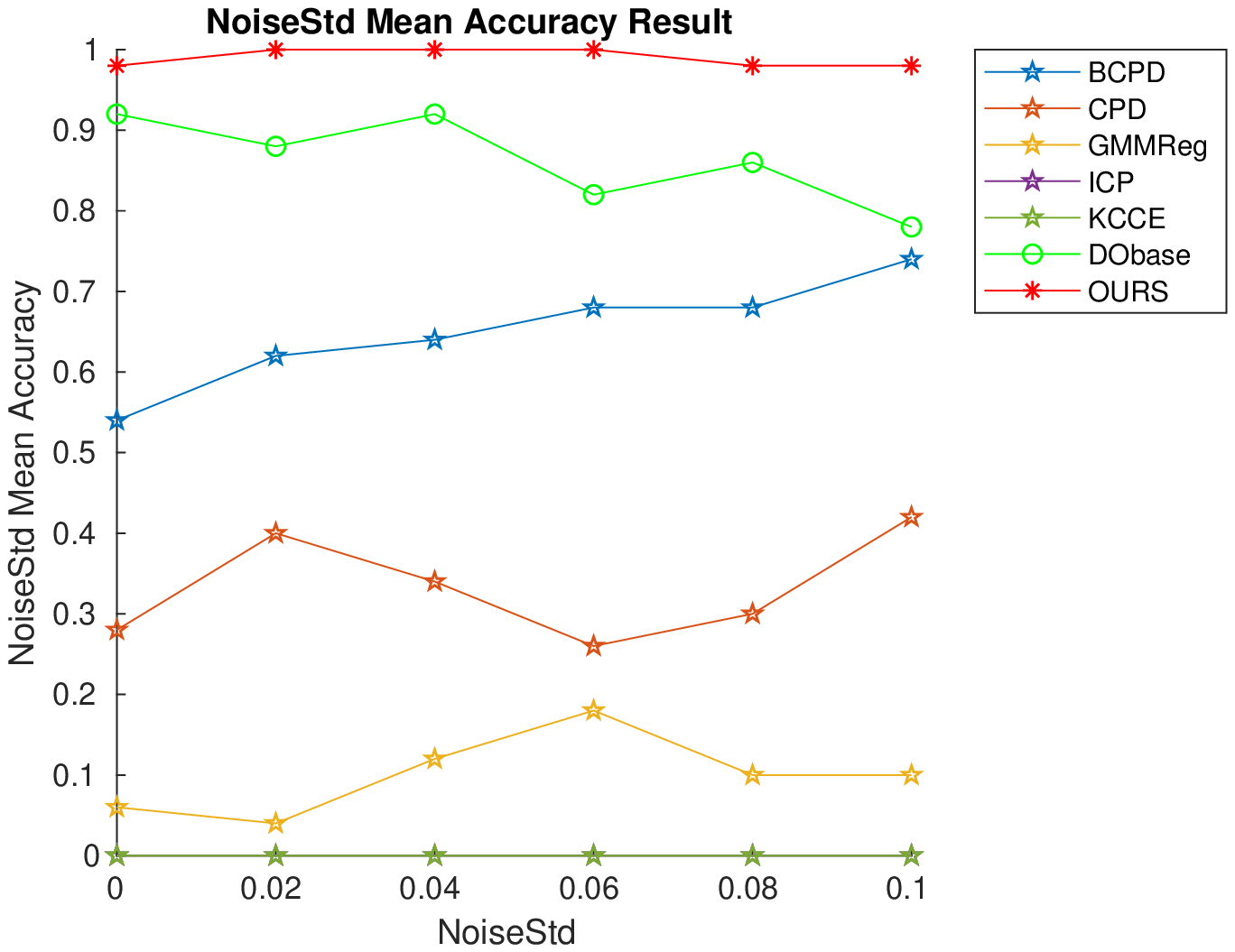}}
	\label{subfig:d1_t_NoiseStd_Acc}
	\hspace{1em}
	\subfloat[Outliers]{\includegraphics[width=0.45\linewidth]{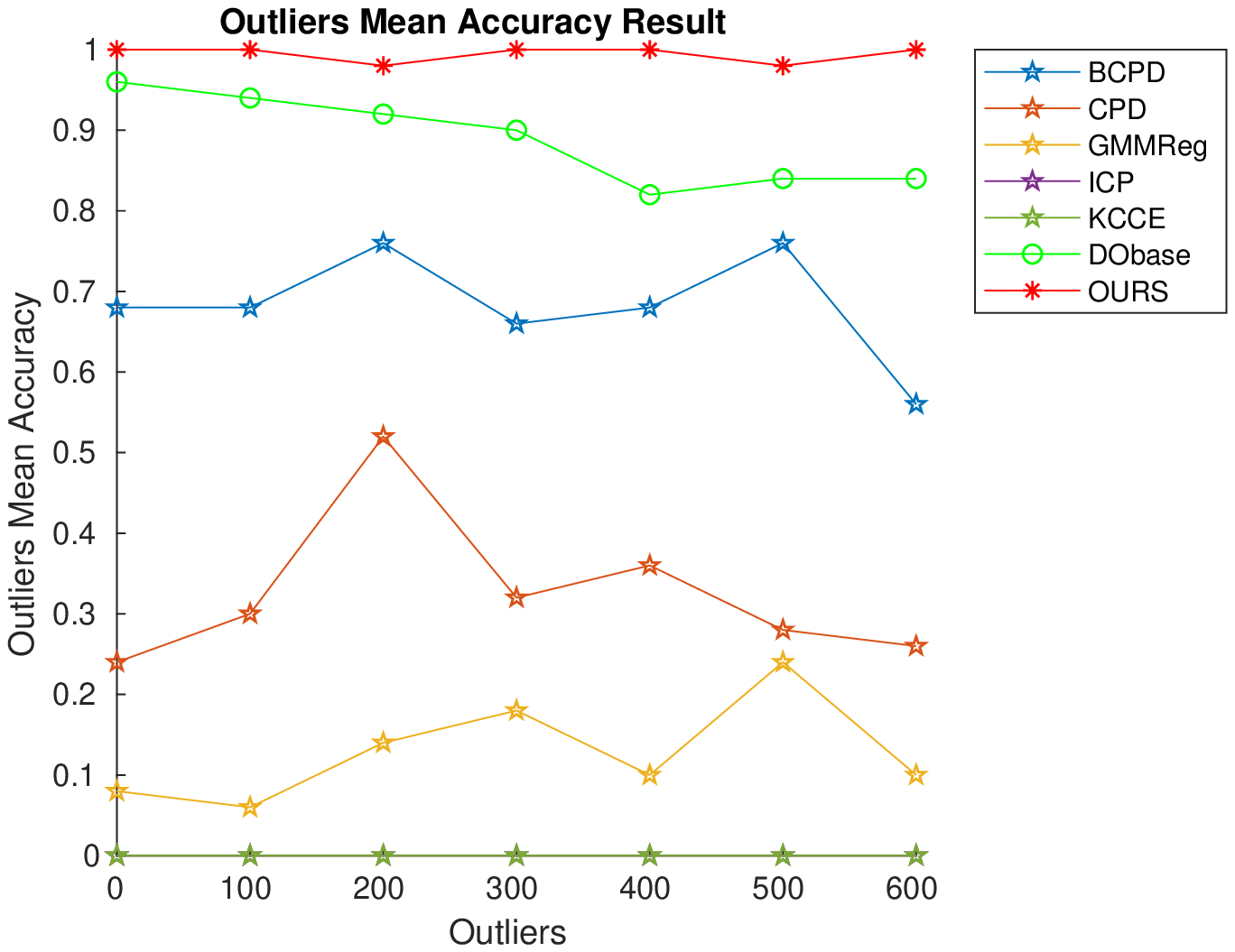}}
	\label{subfig:d1_t_Outliers_Acc}
	
	\subfloat[PointNum]{\includegraphics[width=0.45\linewidth]{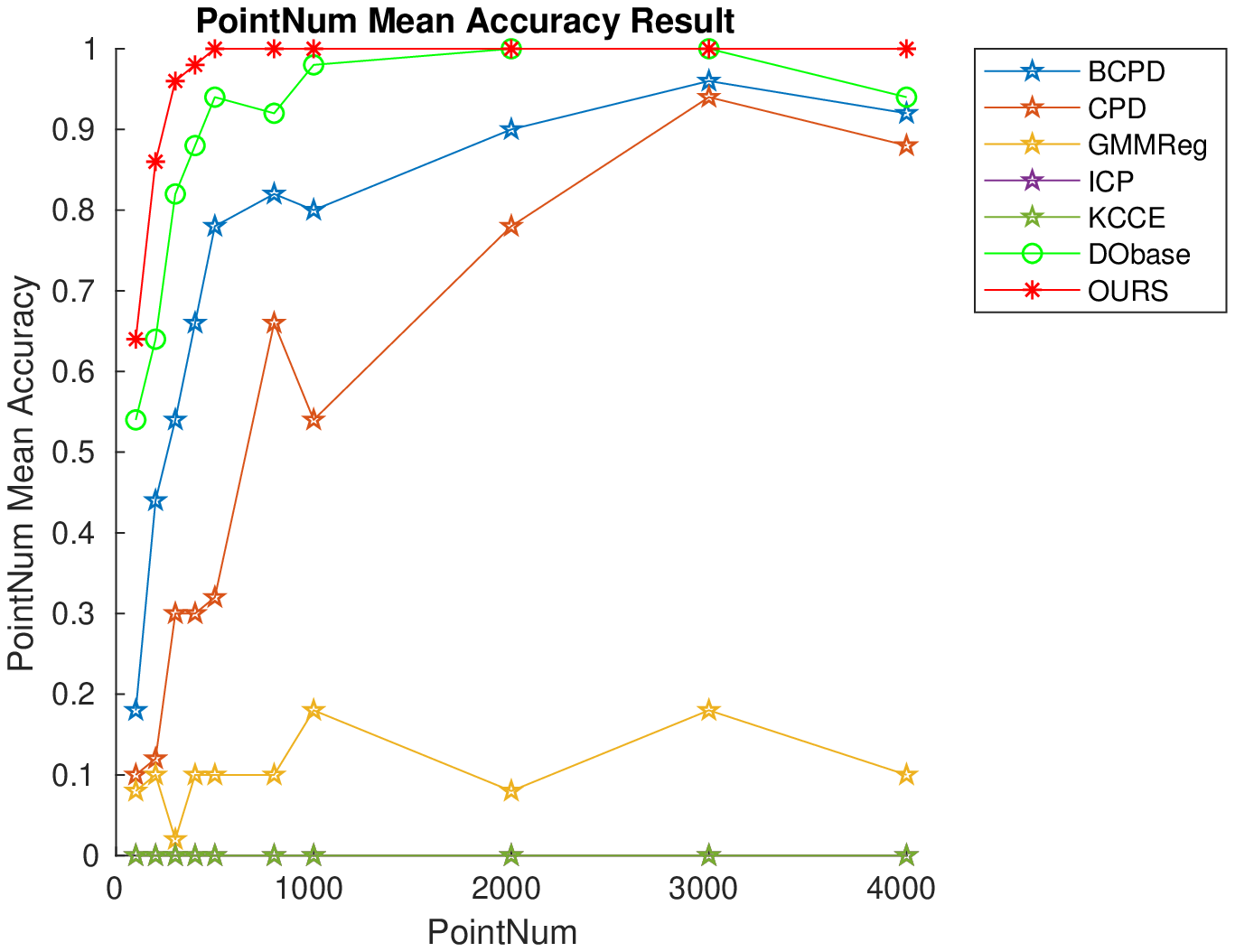}}
	\label{subfig:d1_t_PointNum_Acc}
	\hspace{1em}
	\subfloat[Incomplete]{\includegraphics[width=0.45\linewidth]{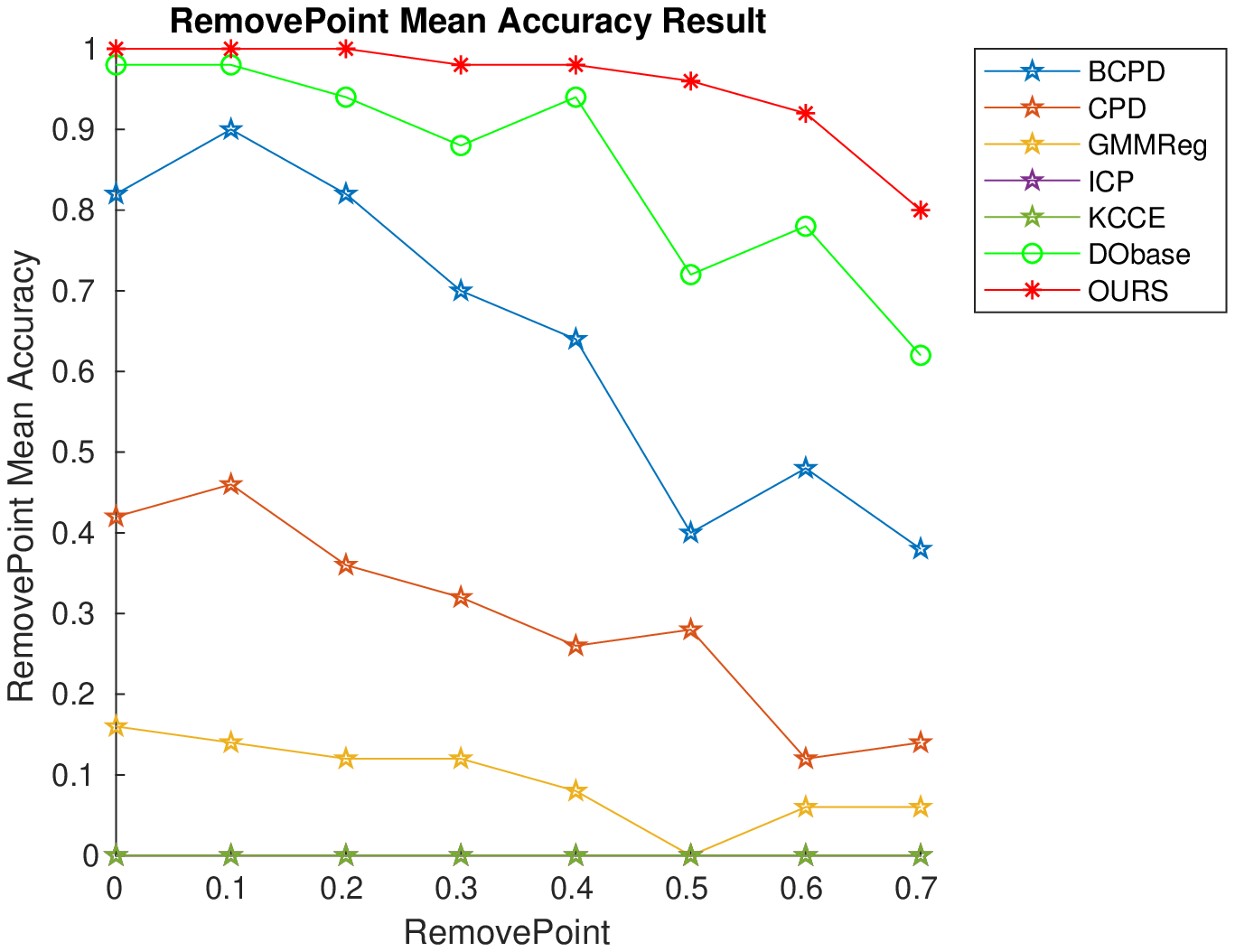}}
	\label{subfig:d1_t_RemovePoint_Acc}
	
	\subfloat[Rotation]{\includegraphics[width=0.45\linewidth]{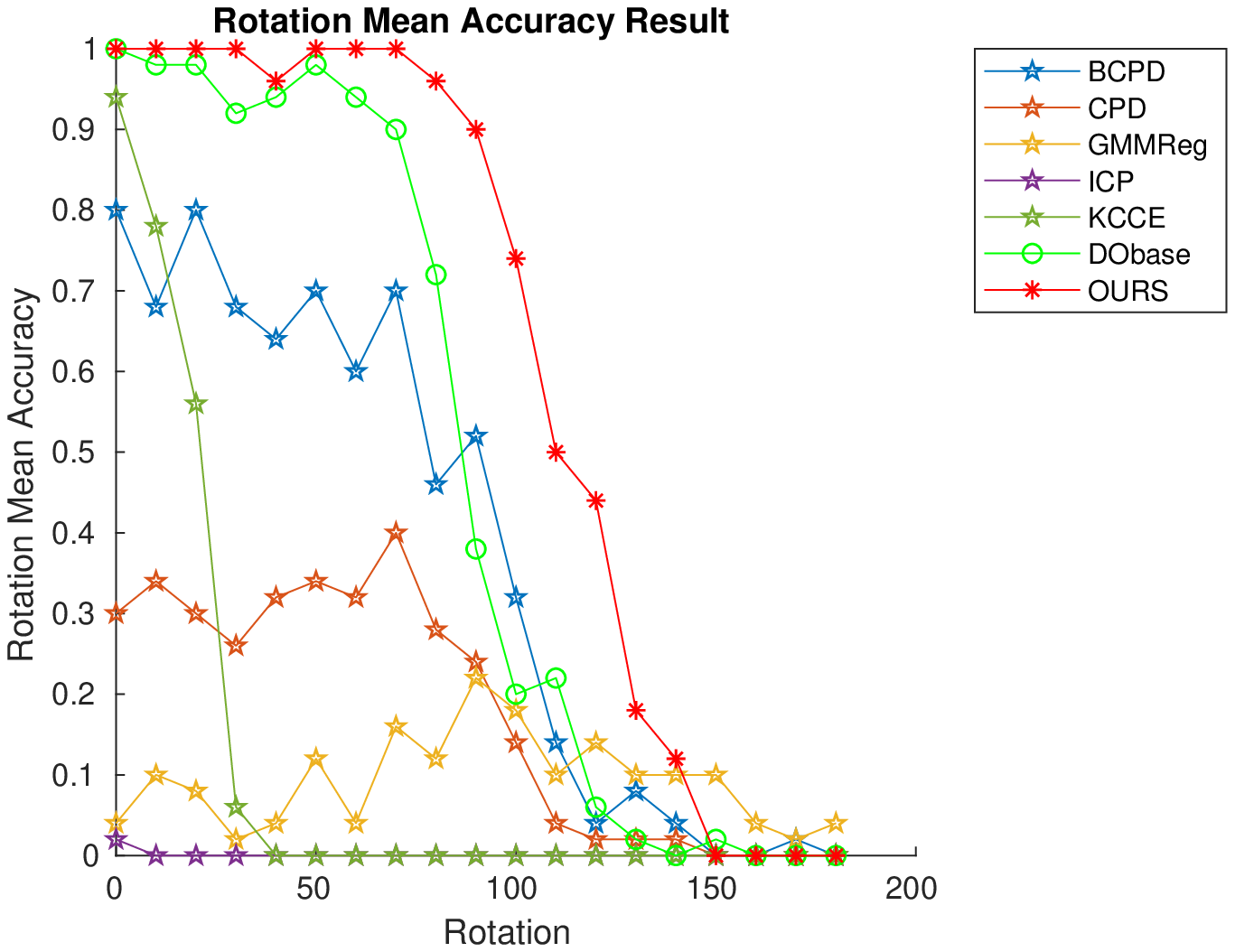}}
	\label{subfig:d1_t_Rotation_Acc}
	\hspace{1em}
	\subfloat[Translation]{\includegraphics[width=0.45\linewidth]{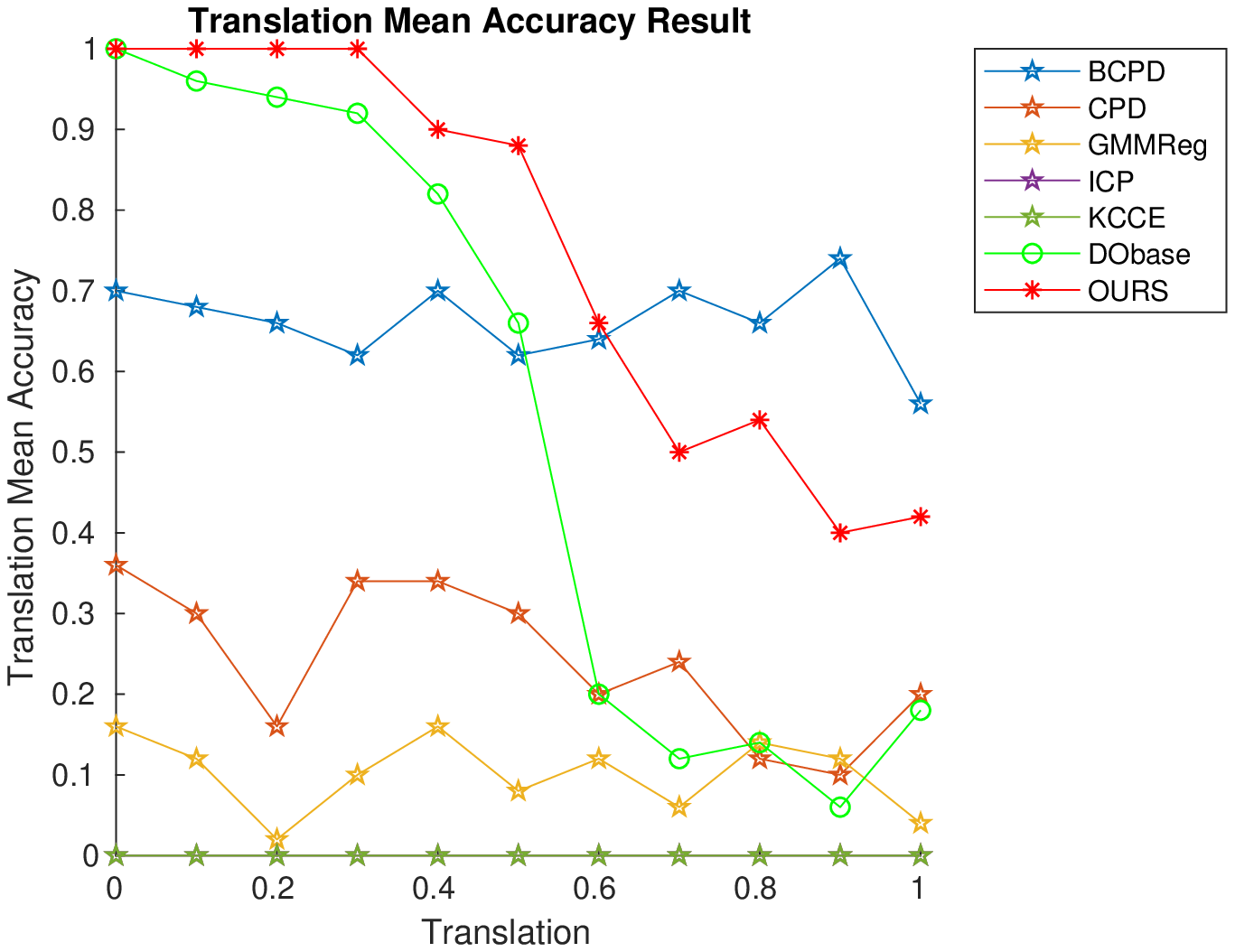}}
	\label{subfig:d1_t_Translation_Acc}
	
	\caption{ PointAcc on Dataset-1.}
	\label{fig:dat1-Acc-result}
\end{figure}

\begin{figure}[htpb]
	\centering	
	\subfloat[NoiseStd]{\includegraphics[width=0.45\linewidth]{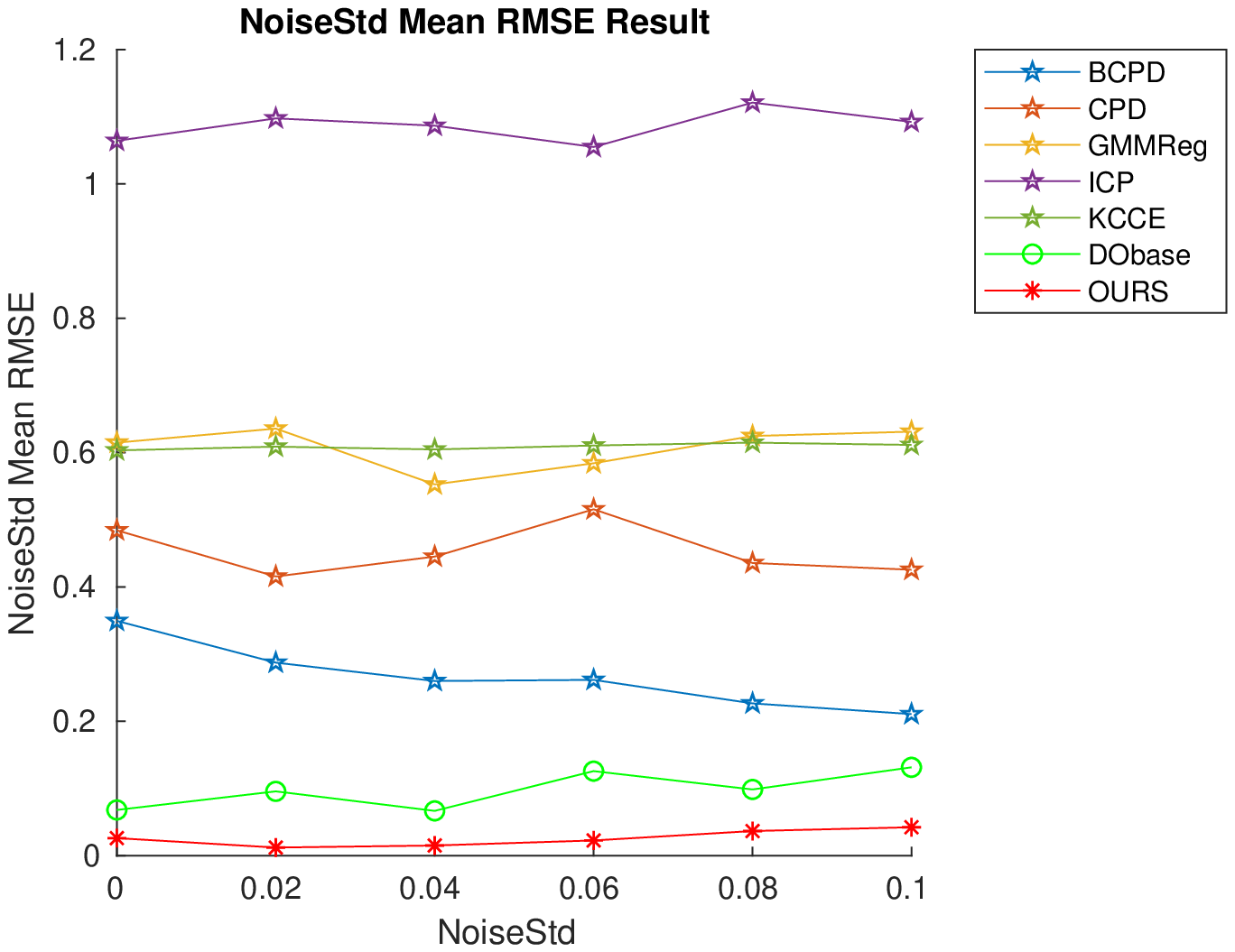}}
	\label{subfig:d1_t_NoiseStd_RMSE}	
	\hspace{1em}
	\subfloat[Outliers]{\includegraphics[width=0.45\linewidth]{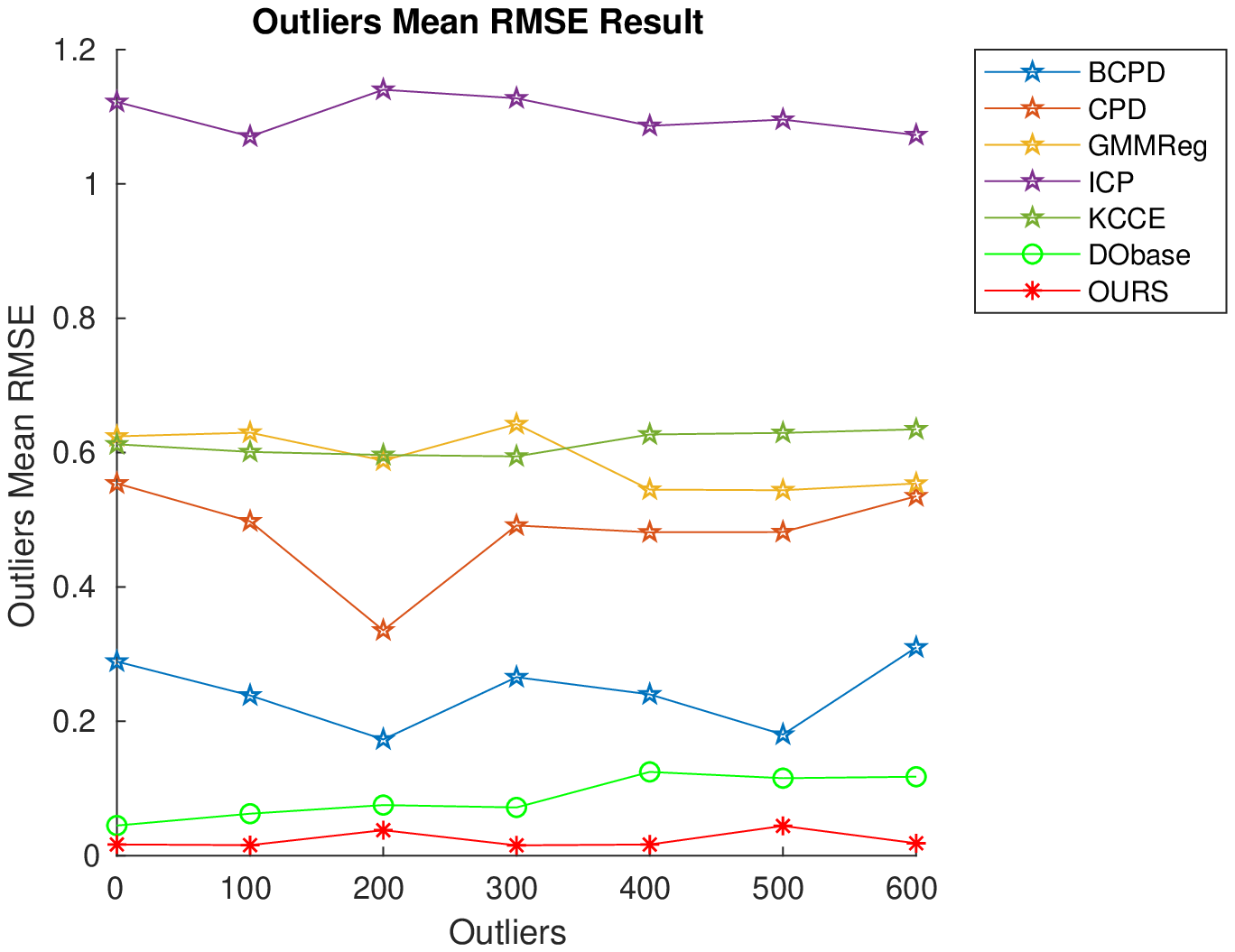}}
	\label{subfig:d1_t_Outliers_RMSE}
	
	\subfloat[PointNum]{\includegraphics[width=0.45\linewidth]{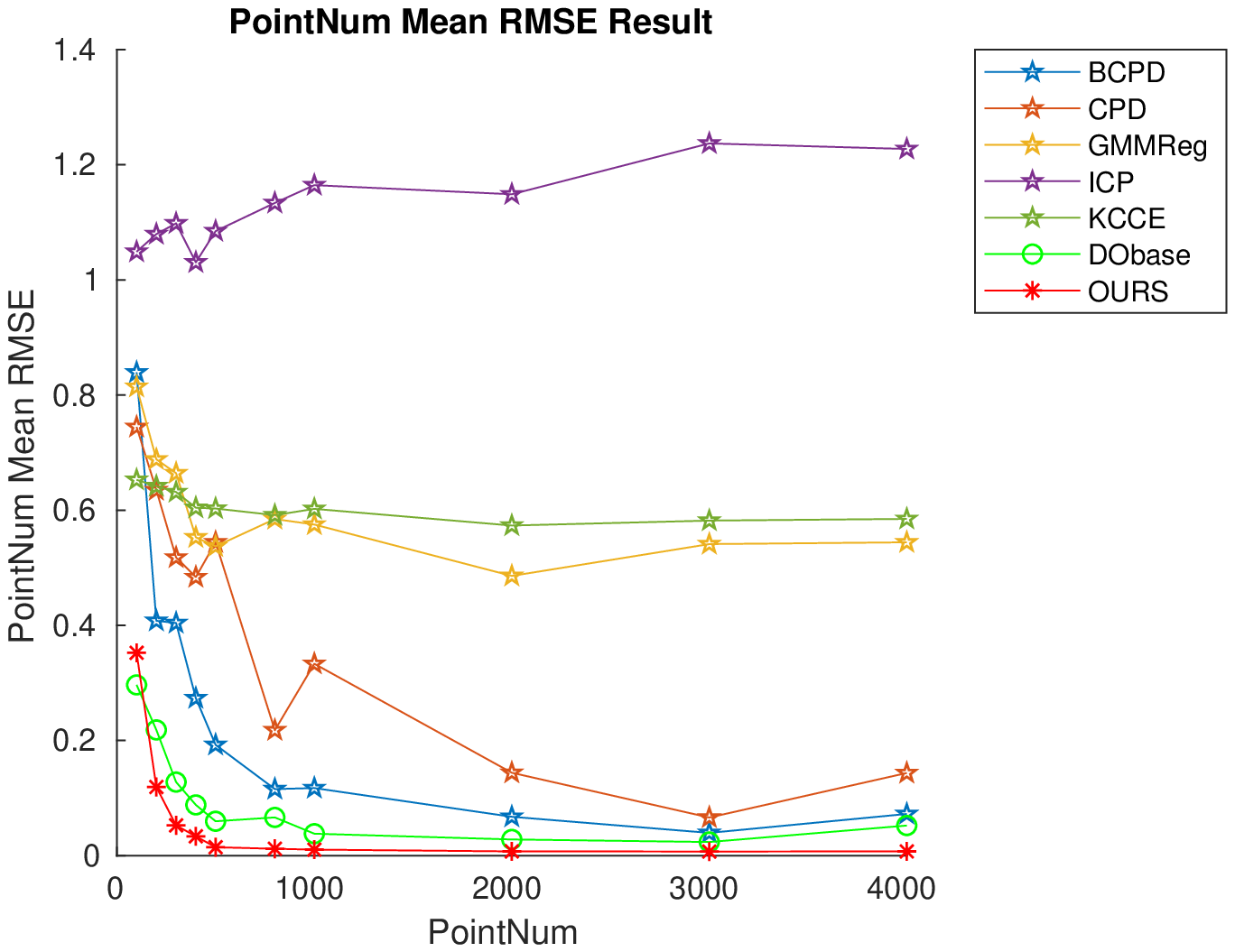}}
	\label{subfig:d1_t_PointNum_RMSE}
	\hspace{1em}	
	\subfloat[Incomplete]{\includegraphics[width=0.45\linewidth]{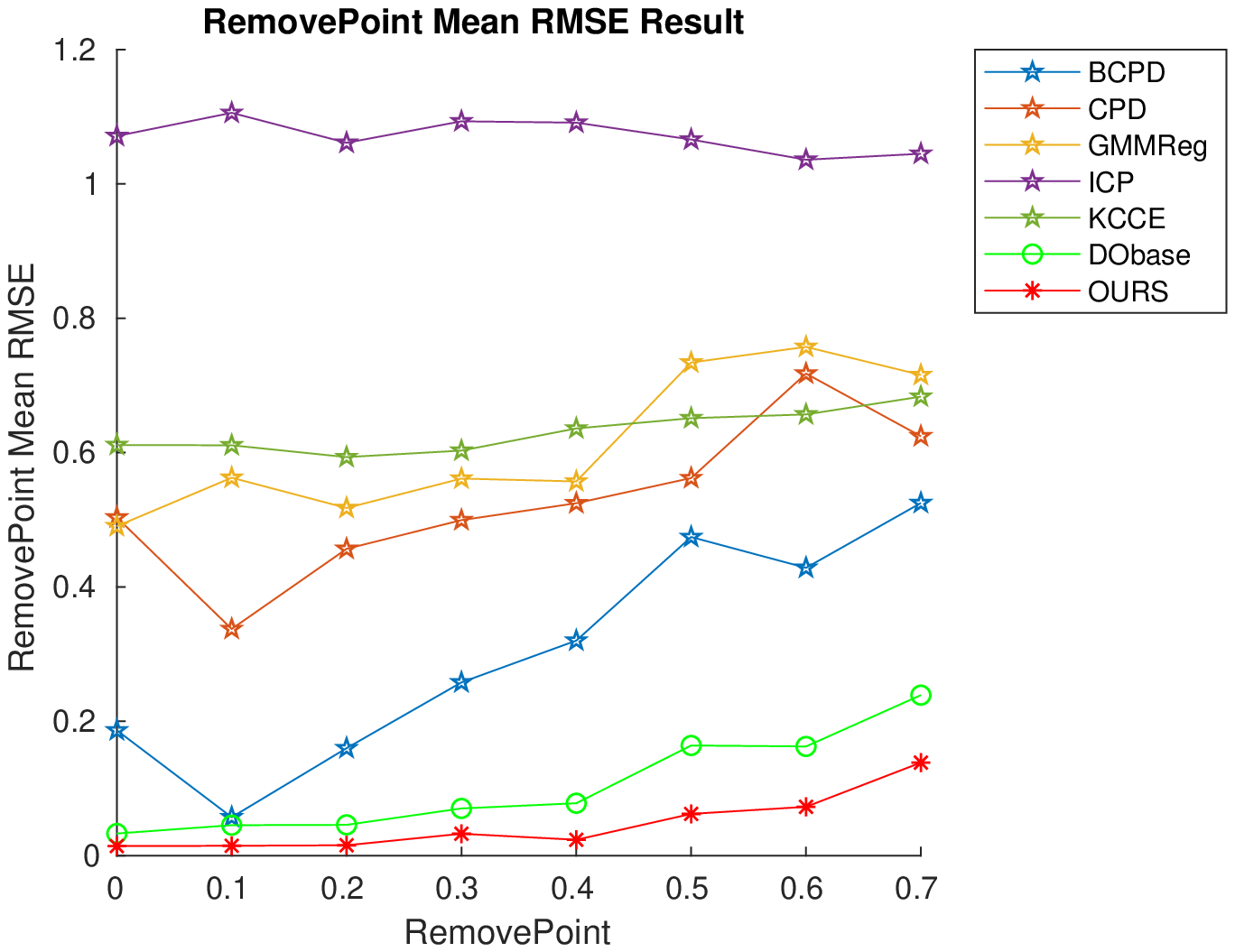}}
	\label{subfig:d1_t_RemovePoint_RMSE}
	
	\subfloat[Rotation]{\includegraphics[width=0.45\linewidth]{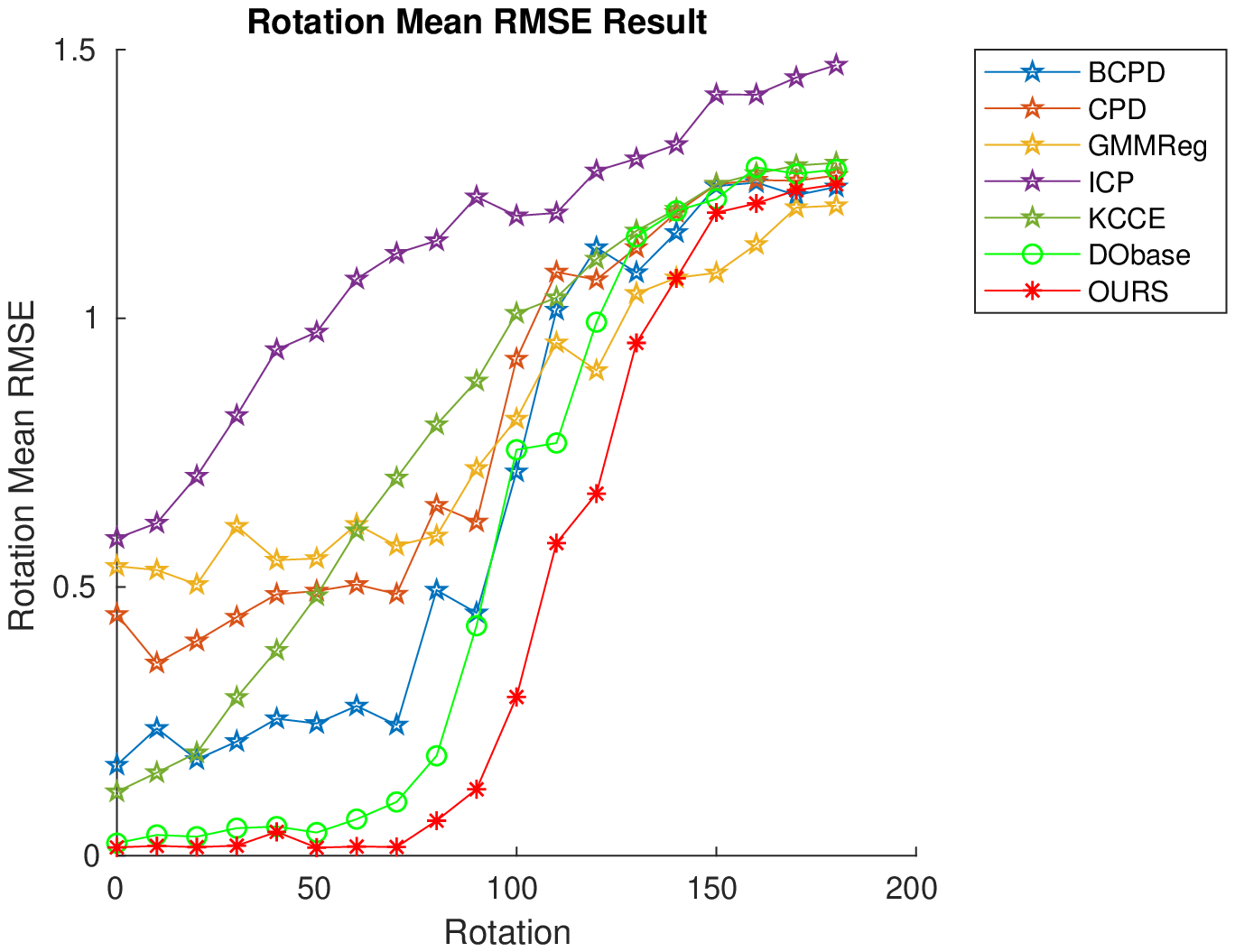}}
	\label{subfig:d1_t_Rotation_RMSE}
	\hspace{1em}
	\subfloat[Translation]{\includegraphics[width=0.45\linewidth]{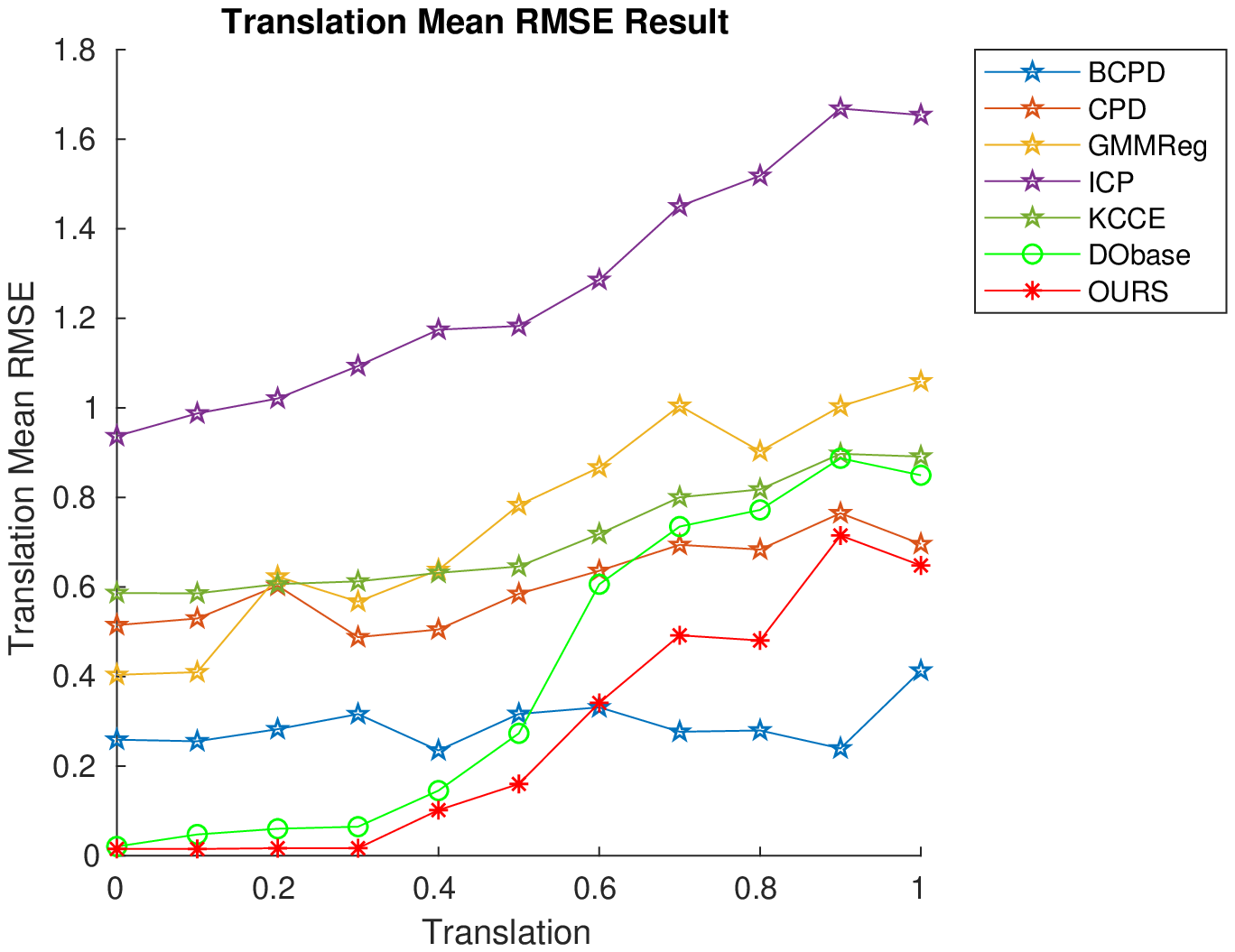}}
	\label{subfig:d1_t_Translation_RMSE}
	
	\caption{PointRMSE on Synthetic Dataset.}
	\label{fig:dat1-RMSE-result}
\end{figure}

In the testing phase, we set the terminated condition $maxIter = 1000$ and $\epsilon = 0.005$, so that the termination occurs when the iterations beyond 1000 or updating's norm below 0.005. We compared our improved DO with other six algorithms (i.e. ICP, CPD, BCPD, KCCE, GMMReg, and DO) in the synthetic dataset. We ploted the iteration process from our improved DO and the original DO in Fig.\ref{fig:dat1-Iter-result}. From the iterative process figures demonstrated that our improved DO algorithm converges faster than the DO algorithm. To providing a quantitiative comparison, we collected the experimental results, and calcualted the mean PointAcc and mean PointRMSE on each perturbation level, then ploted them in Fig.\ref{fig:dat1-Acc-result} and Fig.\ref{fig:dat1-RMSE-result}. Every sub-figure demonstrates all algorithms' performance on different distortion. the distortions include NoiseSTD, Outliers, PointNum, Incomplte, Rotation, and Translation. Fig.\ref{fig:dat1-Acc-result} shows that our improved DO achieved the best point registration mean accuracy in the majority of cases. Our algorithm is robust in the NoiseSTD and Outliers degradation. Eventhough the accuracy of our algorithm decreases in the Incomplete and Rotation distortion, but our method still outperform the other algorithms, and allway better than the original DO in each cases. The statistical results in Fig.\ref{fig:dat1-Acc-result} illustrate that the ICP method had a low accuracy in all perturbation cases, that because the ICP need an ideal initialization, otherwise trapping into a local minimum. The KCCE and GMMReg method also performed a bad accuracy in all cases revealed they are sensitive to degradations. The CPD algorithm utilized an posterior probebility to evaluate the point correspondence and performed not too bad. The latest BCPD method outperform the CPD in all cases, because the variational bayesian inference was adopted in the algorithm's posterior probebility to evalute the correspondence. But both of the CPDs did not outperform the original DO in most of cases. The translation perturbation experiments shown in Fig.\ref{fig:dat1-Acc-result}(f) and Fig.\ref{fig:dat1-RMSE-result}(f) demonstate that our algorithm performed worse than the BCPD algorithm when the initial value increased. This can be interpreted as: when the distance between model and scene pionts getting far away, the Gaussian weight will be a infinitesimal value, thus the histogram will be a sparse vector, thus the regressor will be failed. When the translation initial value above 0.7, the DO's performance declined dramatically. But our improved DO still better than DO because of the extra histograms. Fig.\ref{fig:dat1-RMSE-result} summarized the statistical results of mean PointRMSE. As can be seen from the figure, ICP, GMMREG and KCCE performed poorly. BCPD still performed better than CPD in all cases, and DO better than the previous algorithm. Our improved DO achevied the best performance in almost cases, except the translation degradation case. For each perturbation experiment, we calculated the average PointAcc and average PointRMSE result from each algorithm, and listed them in the Table.\ref{tab:data1-ACC} and Table.\ref{tab:data1-RMSE}, and highlight the best result is in bold. The result demonstrated that our proposed algorithm outperform than the other algorithms.

\begin{table}[]
	\centering
	\caption{ Average PointAcc on Synthetic Dataset}
	\label{tab:data1-ACC}
	\resizebox{0.5\textwidth}{!}{%
		\begin{tabular}{|c|c|c|c|c|c|c|c|}
			\hline
			\multicolumn{8}{|c|}{Average PointAcc}                              \\ \hline
			& BCPD  & CPD   & GMMReg & ICP   & KCCE  & DO    & OURS           \\ \hline
			NoiseStd    & 0.650 & 0.333 & 0.100  & 0.000 & 0.000 & 0.863 & \textbf{0.990} \\ \hline
			Outliers    & 0.683 & 0.326 & 0.129  & 0.000 & 0.000 & 0.889 & \textbf{0.994} \\ \hline
			PointNum    & 0.700 & 0.494 & 0.104  & 0.000 & 0.000 & 0.866 & \textbf{0.944} \\ \hline
			Incomplete   & 0.643 & 0.295 & 0.093  & 0.000 & 0.000 & 0.855 & \textbf{0.955} \\ \hline
			Rotation    & 0.380 & 0.176 & 0.093  & 0.001 & 0.123 & 0.487 & \textbf{0.621} \\ \hline
			Translation & 0.662 & 0.242 & 0.102  & 0.000 & 0.000 & 0.545 & \textbf{0.755} \\ \hline
		\end{tabular}
	}
\end{table}

\begin{table}[]
	\centering
	\caption{ Average PointRMSE on Synthetic Dataset}
	\label{tab:data1-RMSE}
	\resizebox{0.5\textwidth}{!}{%
		\begin{tabular}{|c|c|c|c|c|c|c|c|}
			\hline
			\multicolumn{8}{|c|}{Average PointRMSE}                                  \\ \hline
			& BCPD  & CPD   & GMMReg & ICP   & KCCE  & DO    & OURS           \\ \hline
			NoiseStd    & 0.266 & 0.454 & 0.607  & 1.086 & 0.609 & 0.098 & \textbf{0.026} \\ \hline
			Outliers    & 0.242 & 0.482 & 0.590  & 1.102 & 0.614 & 0.087 & \textbf{0.024} \\ \hline
			PointNum    & 0.253 & 0.383 & 0.599  & 1.125 & 0.607 & 0.100 & \textbf{0.062} \\ \hline
			Incomplete   & 0.301 & 0.528 & 0.612  & 1.071 & 0.631 & 0.105 & \textbf{0.047} \\ \hline
			Rotation    & 0.676 & 0.807 & 0.802  & 1.118 & 0.802 & 0.576 & \textbf{0.464} \\ \hline
			Translation & 0.291 & 0.609 & 0.751  & 1.270 & 0.708 & 0.405 & \textbf{0.273} \\ \hline
		\end{tabular}
	}
\end{table}

\begin{table}[]
	\centering
	\caption{ Average PointAcc on Real Dataset}
	\label{tab:data2-ACC}
	\resizebox{0.5\textwidth}{!}{%
		\begin{tabular}{|c|c|c|c|c|c|c|c|}
			\hline
			\multicolumn{8}{|c|}{Average PointAcc}                               \\ \hline
			& BCPD  & CPD   & GMMReg & ICP   & KCCE  & DO    & OURS           \\ \hline
			NoiseStd    & 0.743 & 0.383 & 0.020  & 0.000 & 0.003 & 0.777 & \textbf{0.993} \\ \hline
			Outliers    & 0.789 & 0.369 & 0.049  & 0.000 & 0.011 & 0.783 & \textbf{0.991} \\ \hline
			PointNum    & 0.728 & 0.516 & 0.052  & 0.000 & 0.004 & 0.808 & \textbf{0.952} \\ \hline
			Incomplete   & 0.718 & 0.328 & 0.055  & 0.000 & 0.005 & 0.743 & \textbf{0.953} \\ \hline
			Rotation    & 0.417 & 0.176 & 0.084  & 0.000 & 0.175 & 0.404 & \textbf{0.648} \\ \hline
			Translation & 0.662 & 0.242 & 0.102  & 0.000 & 0.000 & 0.545 & \textbf{0.755} \\ \hline
		\end{tabular}
	}
\end{table}

\begin{table}[]
	\centering
	\caption{ Average PointRMSE on Real Dataset}
	\label{tab:data2-RMSE}
	\resizebox{0.5\textwidth}{!}{%
		\begin{tabular}{|c|c|c|c|c|c|c|c|}
			\hline
			\multicolumn{8}{|c|}{Average PointRMSE}                               \\ \hline
			& BCPD  & CPD   & GMMReg & ICP   & KCCE  & DO    & OURS           \\ \hline
			NoiseStd    & 0.217 & 0.400 & 0.641  & 1.122 & 0.494 & 0.132 & \textbf{0.024} \\ \hline
			Outliers    & 0.177 & 0.414 & 0.627  & 1.128 & 0.495 & 0.132 & \textbf{0.020} \\ \hline
			PointNum    & 0.237 & 0.343 & 0.619  & 1.144 & 0.497 & 0.130 & \textbf{0.045} \\ \hline
			Incomplete   & 0.242 & 0.420 & 0.637  & 1.128 & 0.515 & 0.169 & \textbf{0.049} \\ \hline
			Rotation    & 0.635 & 0.754 & 0.793  & 1.154 & 0.757 & 0.626 & \textbf{0.430} \\ \hline
			Translation & 0.291 & 0.609 & 0.751  & 1.270 & 0.708 & 0.405 & \textbf{0.273} \\ \hline
		\end{tabular}
	}
\end{table}

\begin{figure}[htpb]
	\centering
	\subfloat[NoiseStd]{\includegraphics[width=0.45\linewidth]{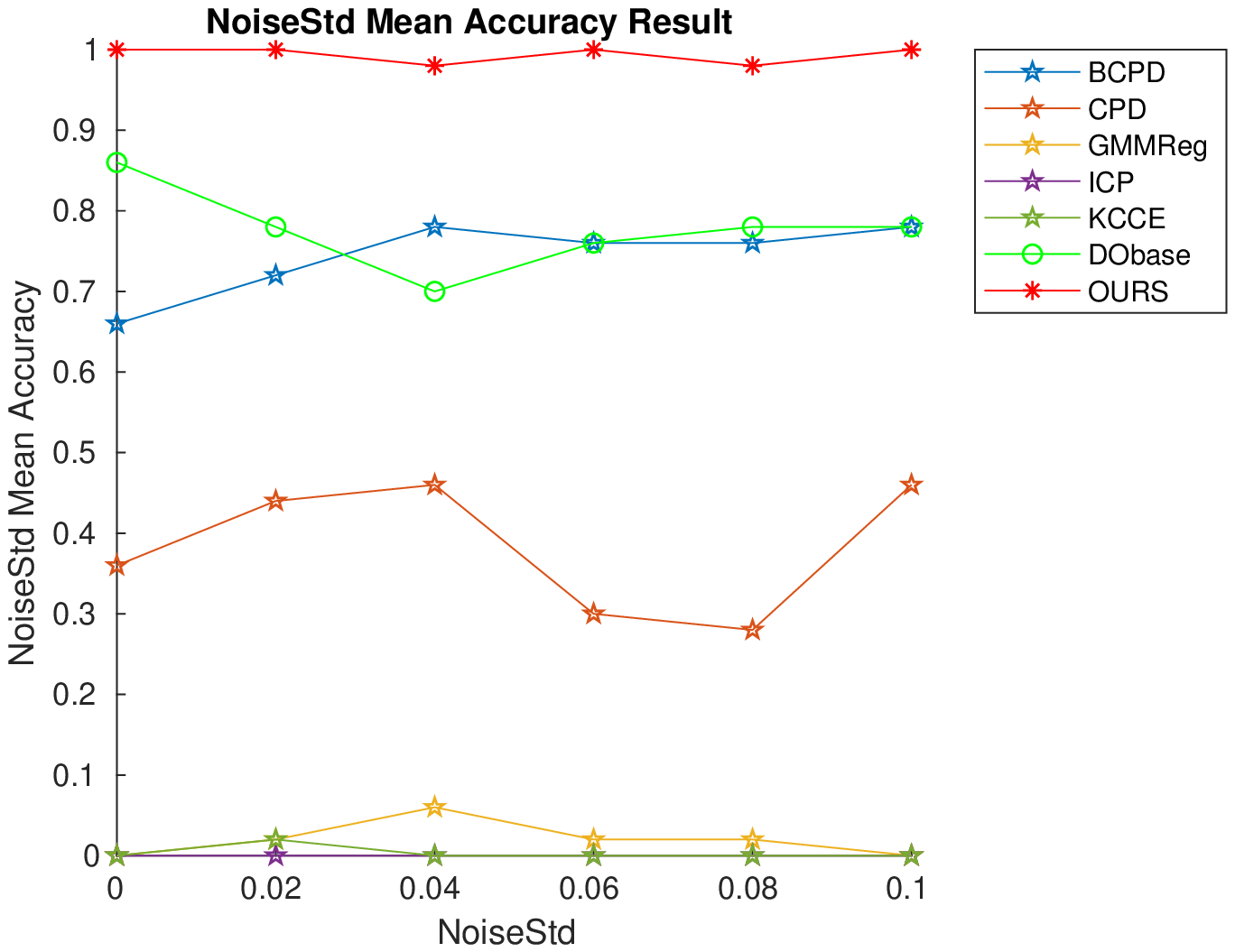}}
	\label{subfig:d2_t_NoiseStd_Acc}
	\hspace{1em}
	\subfloat[Outliers]{\includegraphics[width=0.45\linewidth]{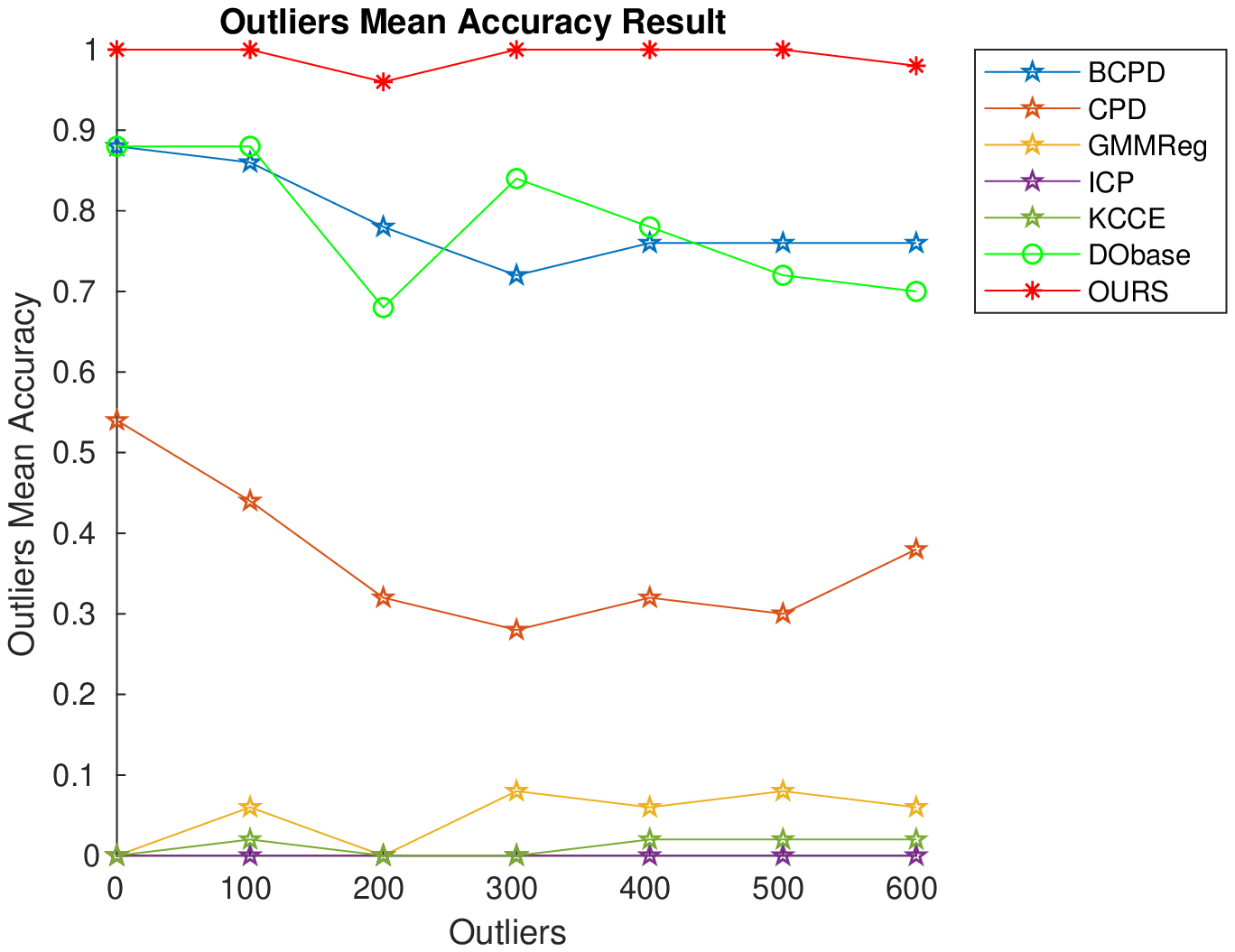}}
	\label{subfig:d2_t_Outliers_Acc}

	\subfloat[PointNum]{\includegraphics[width=0.45\linewidth]{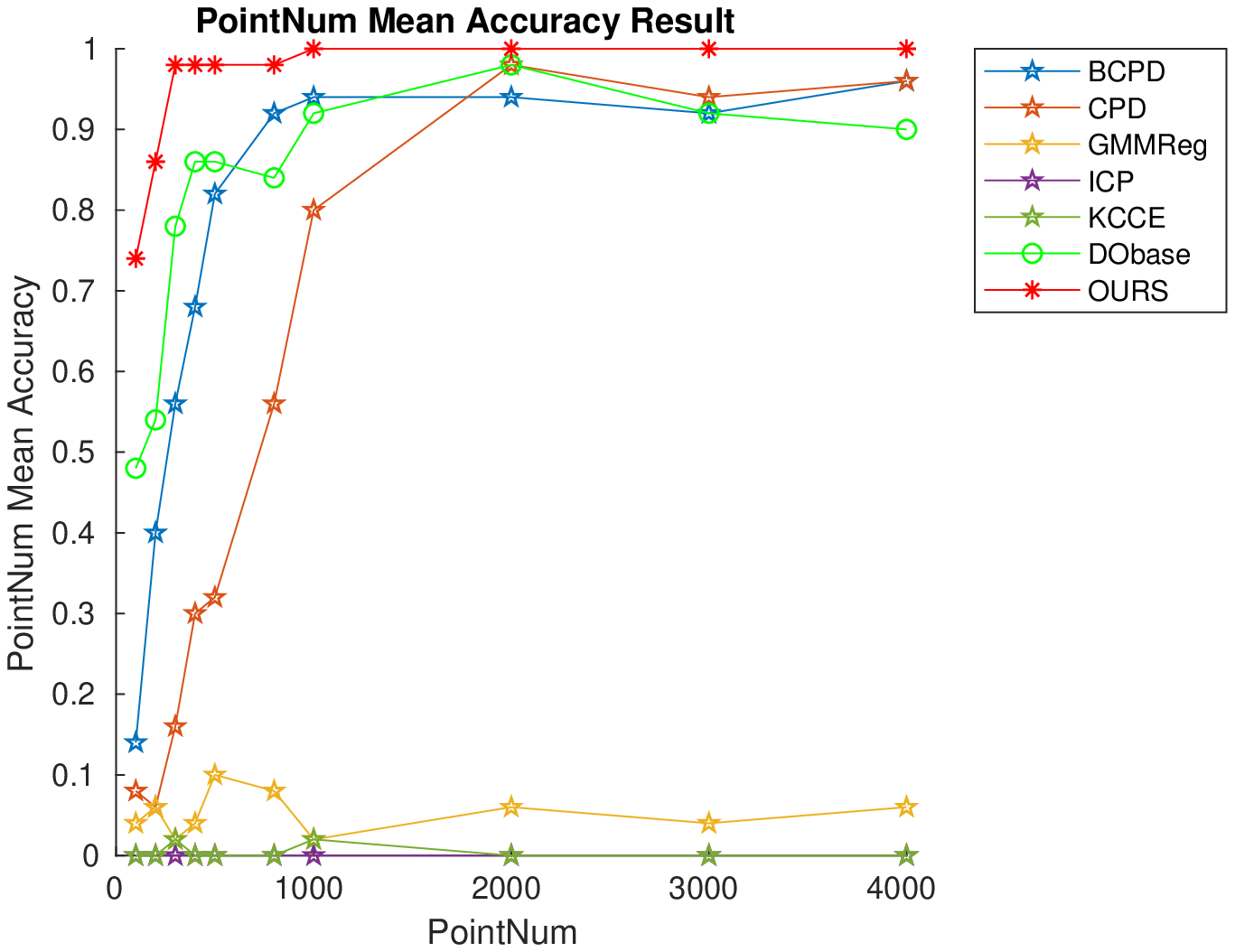}}
	\label{subfig:d2_t_PointNum_Acc}
	\hspace{1em}	
	\subfloat[Occlusion]{\includegraphics[width=0.45\linewidth]{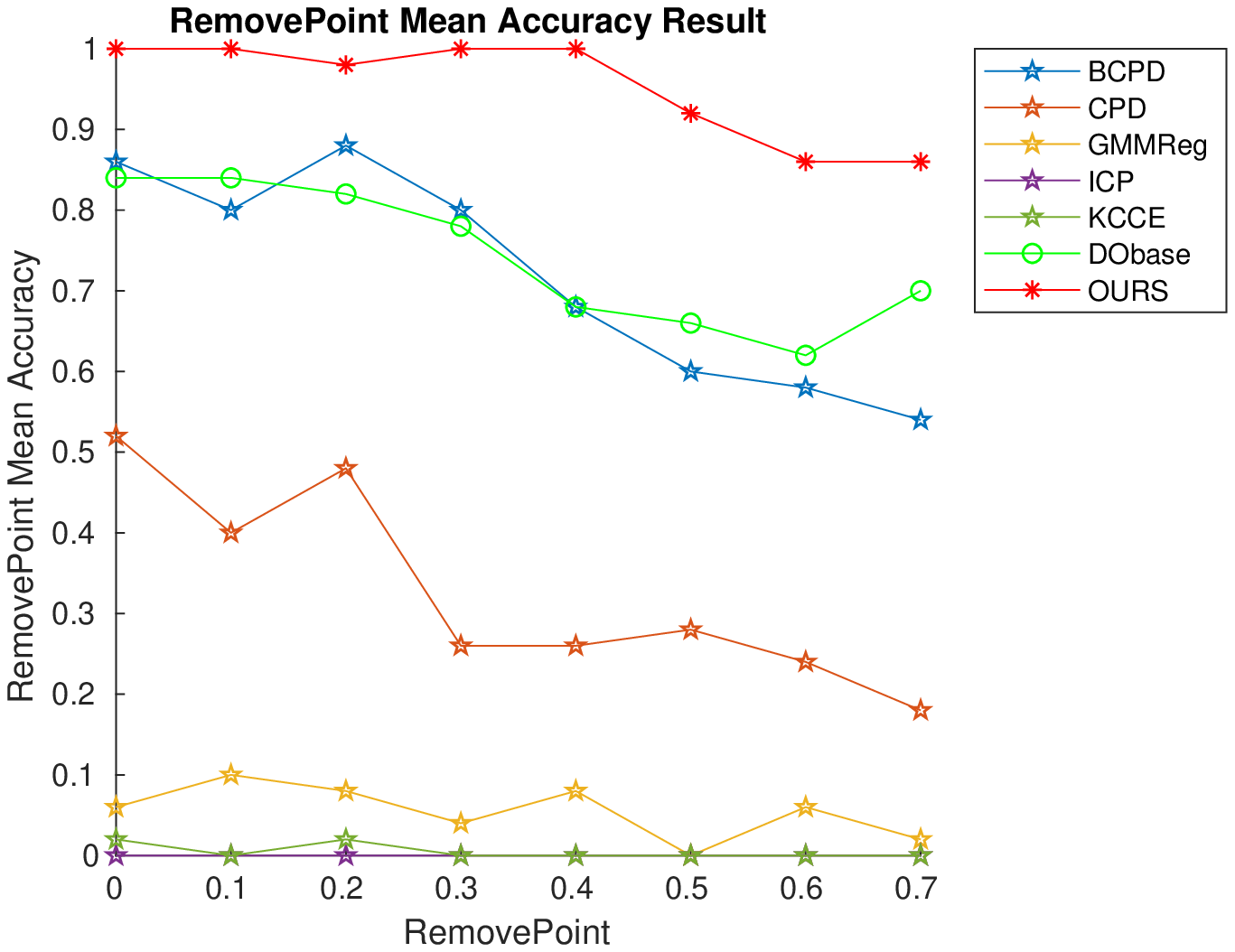}}
	\label{subfig:d2_t_RemovePoint_Acc}

	\subfloat[Rotation]{\includegraphics[width=0.45\linewidth]{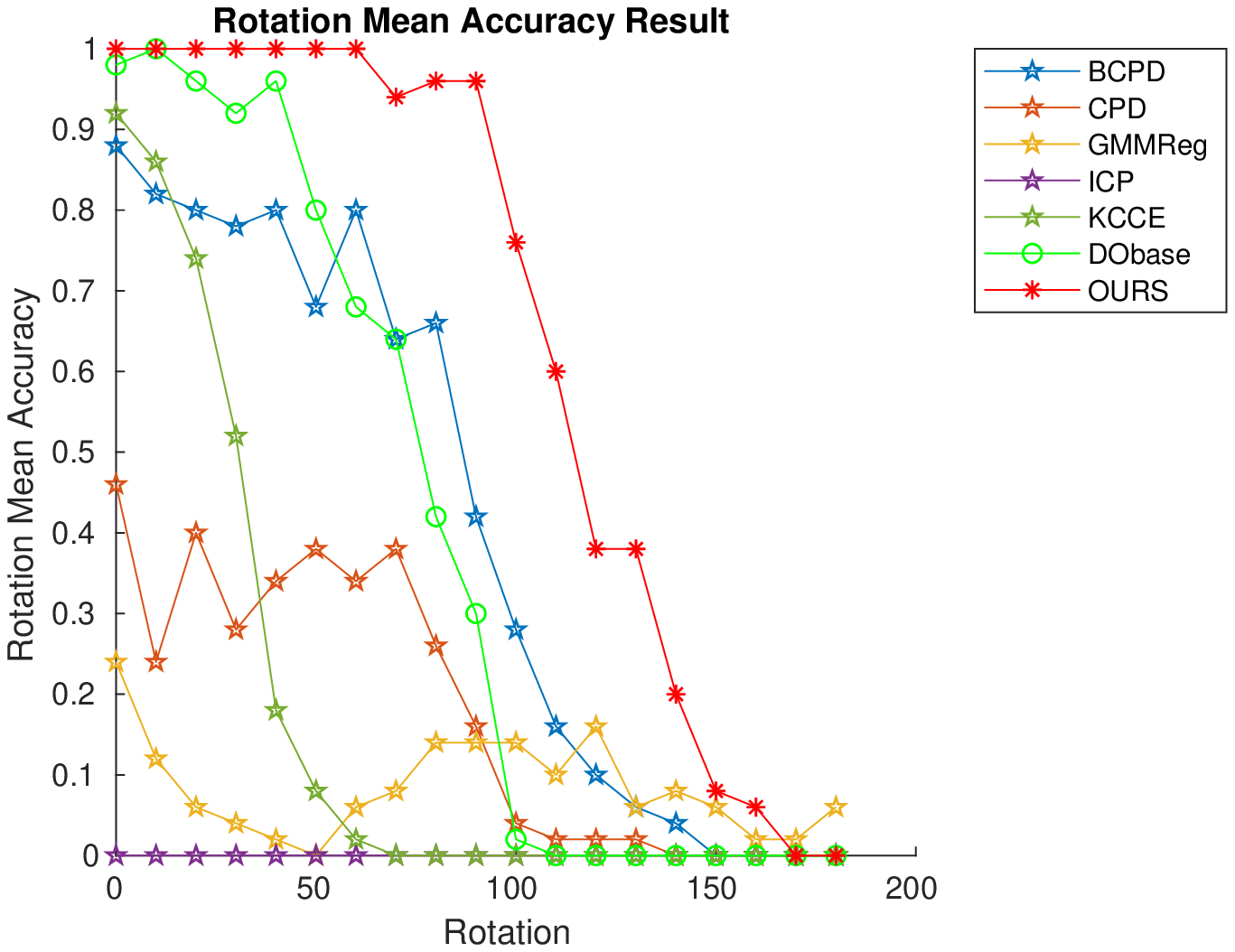}}
	\label{subfig:d2_t_Rotation_Acc}
	\hspace{1em}
	\subfloat[Translation]{\includegraphics[width=0.45\linewidth]{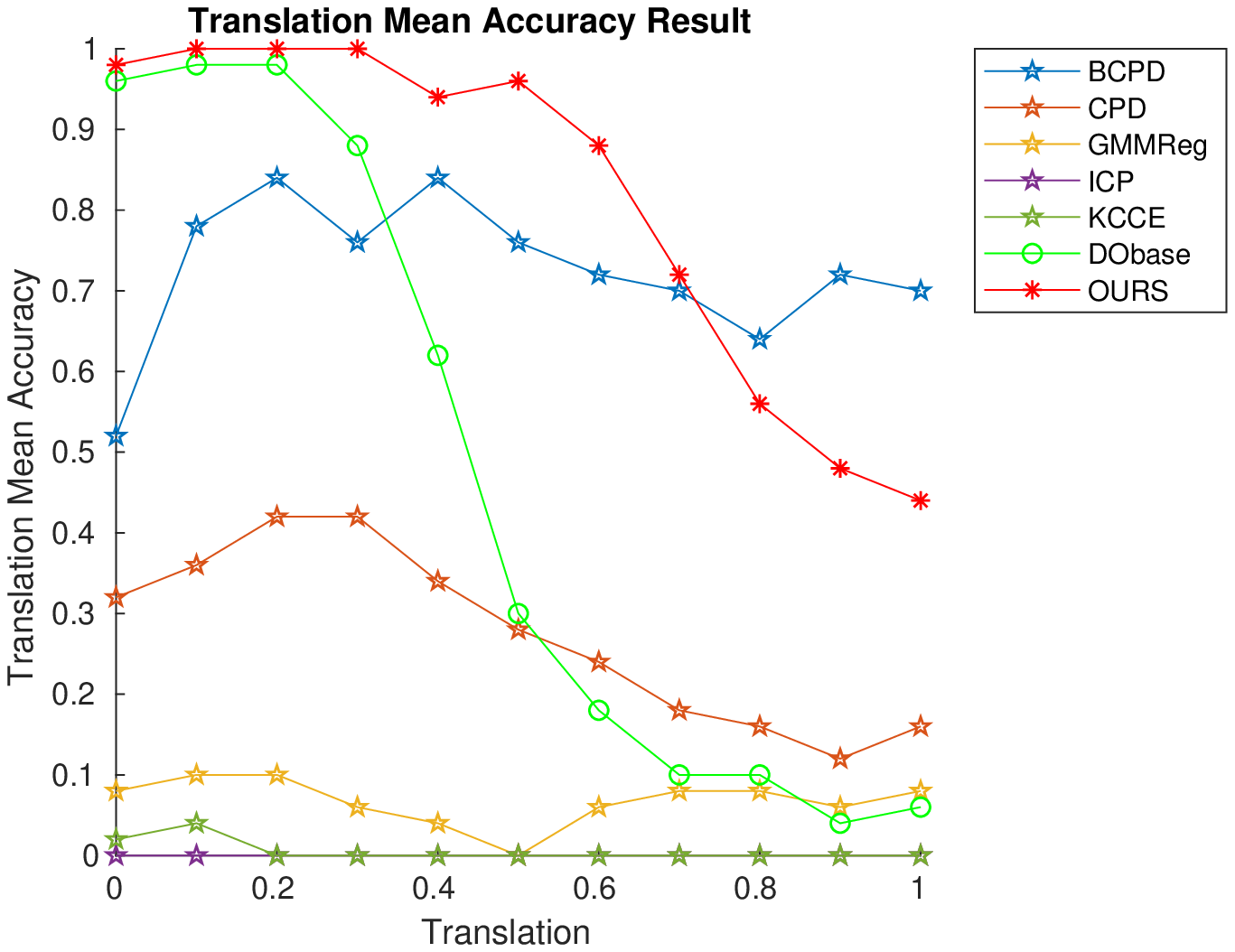}}
	\label{subfig:d2_t_Translation_Acc}
	
	\caption{Mean Accuracy Experimental Result on Dataset-2.}
	\label{fig:dat2-t-Acc-result}
\end{figure}

\begin{figure}[htpb]
	\centering	
	\subfloat[NoiseStd RMSE]{\includegraphics[width=0.45\linewidth]{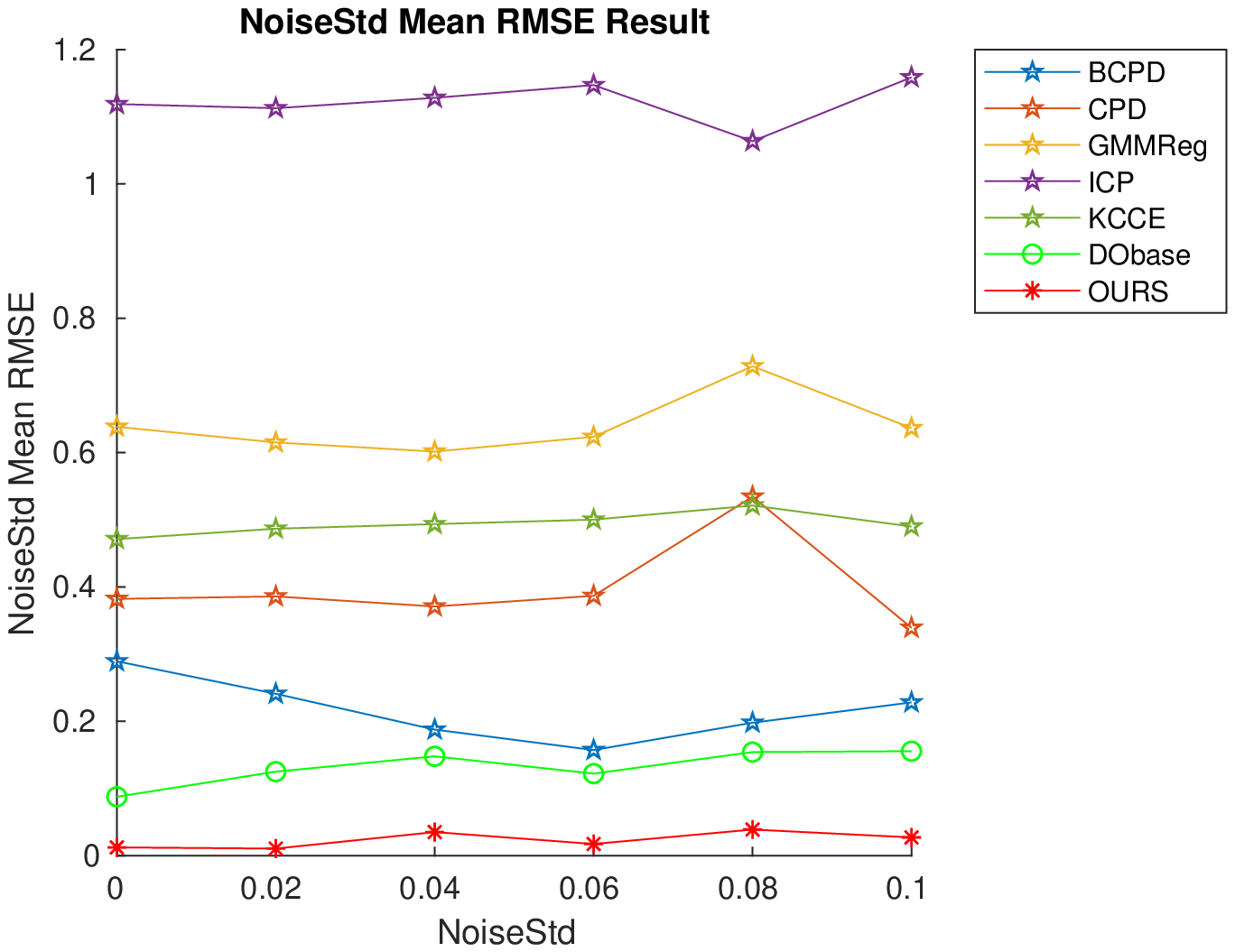}}
	\label{subfig:d2_t_NoiseStd_RMSE}
	\hspace{1em}	
	\subfloat[Outliers RMSE]{\includegraphics[width=0.45\linewidth]{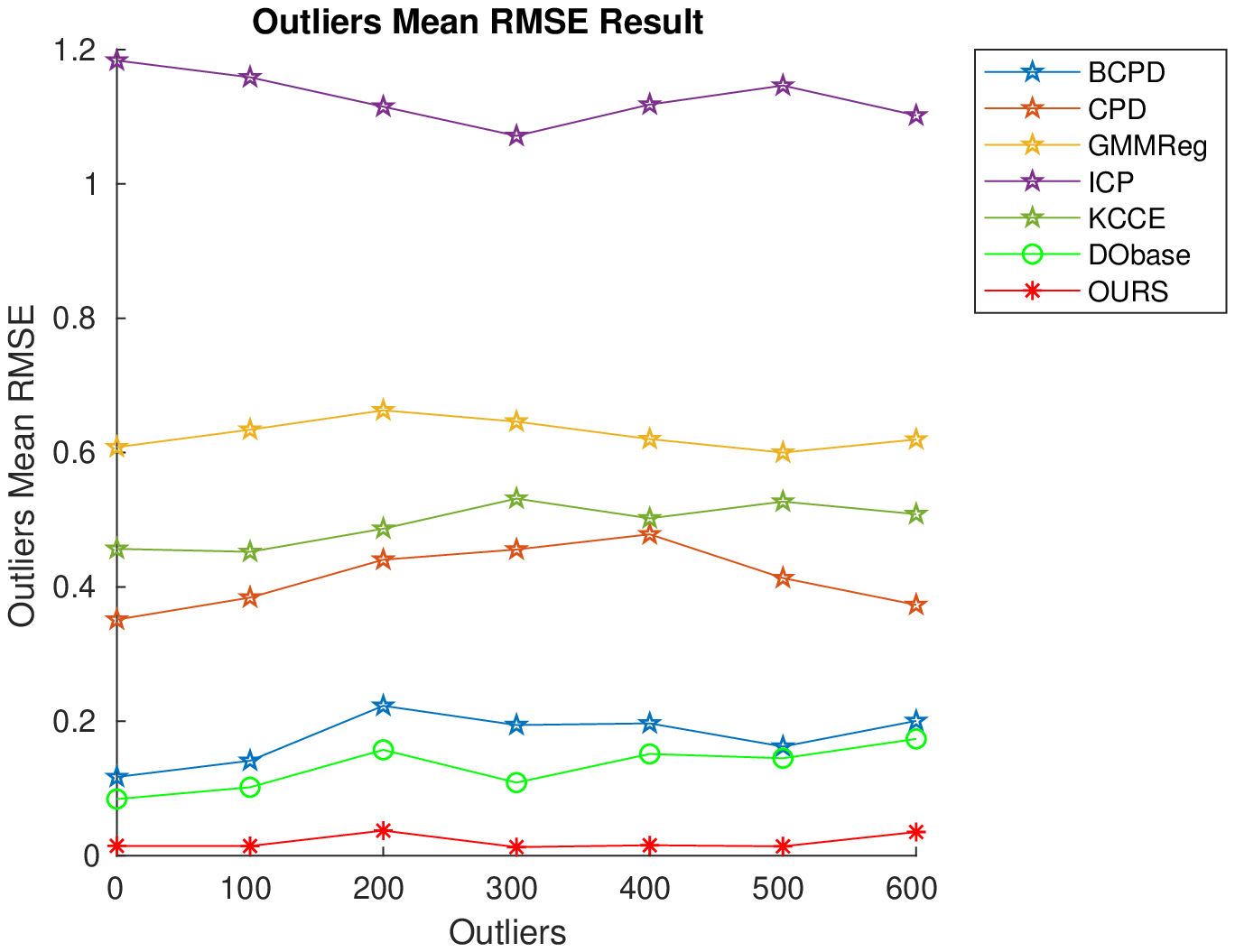}}
	\label{subfig:d2_t_Outliers_RMSE}

	\subfloat[PointNum RMSE]{\includegraphics[width=0.45\linewidth]{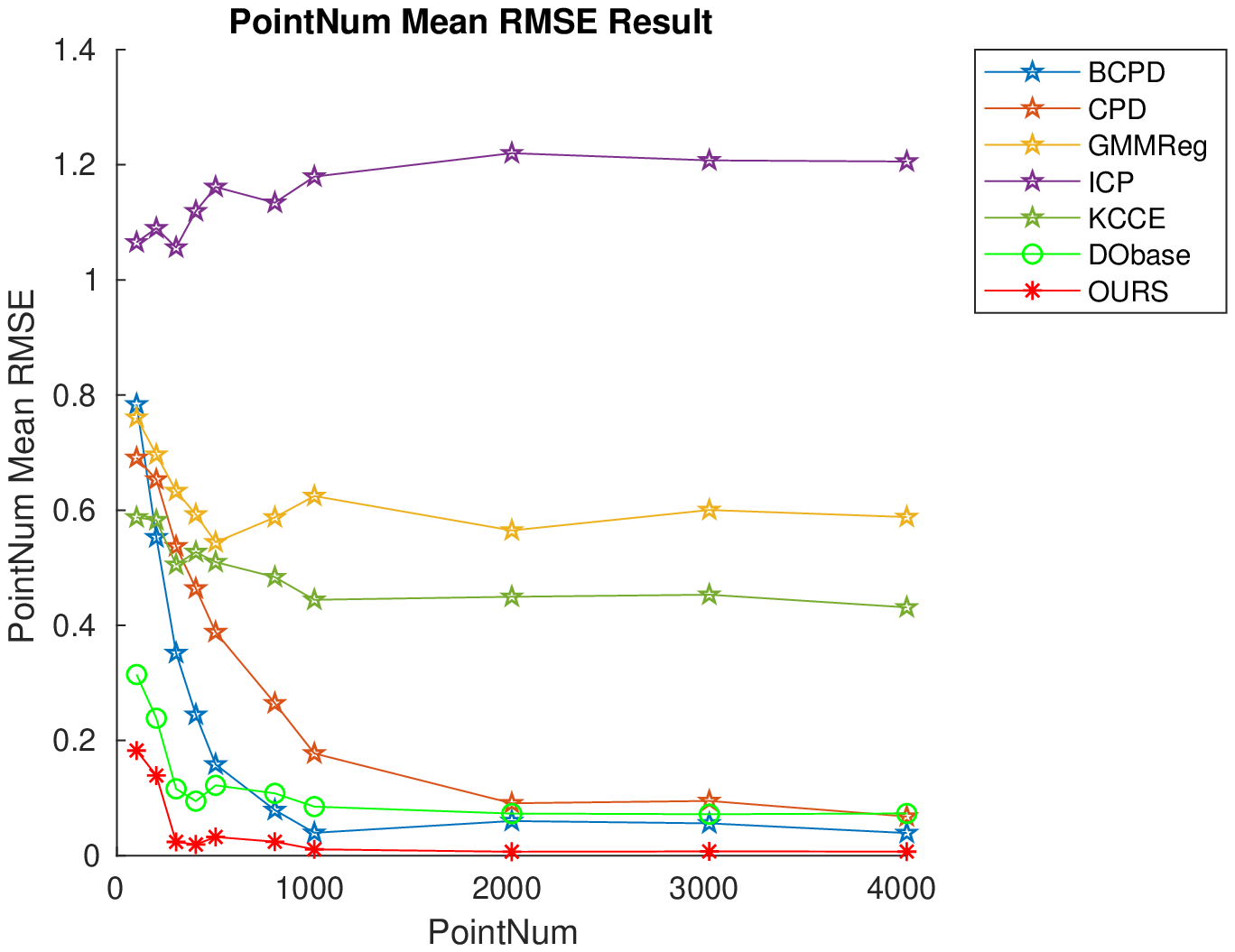}}
	\label{subfig:d2_t_PointNum_RMSE}
	\hspace{1em}
	\subfloat[Occlusion RMSE]{\includegraphics[width=0.45\linewidth]{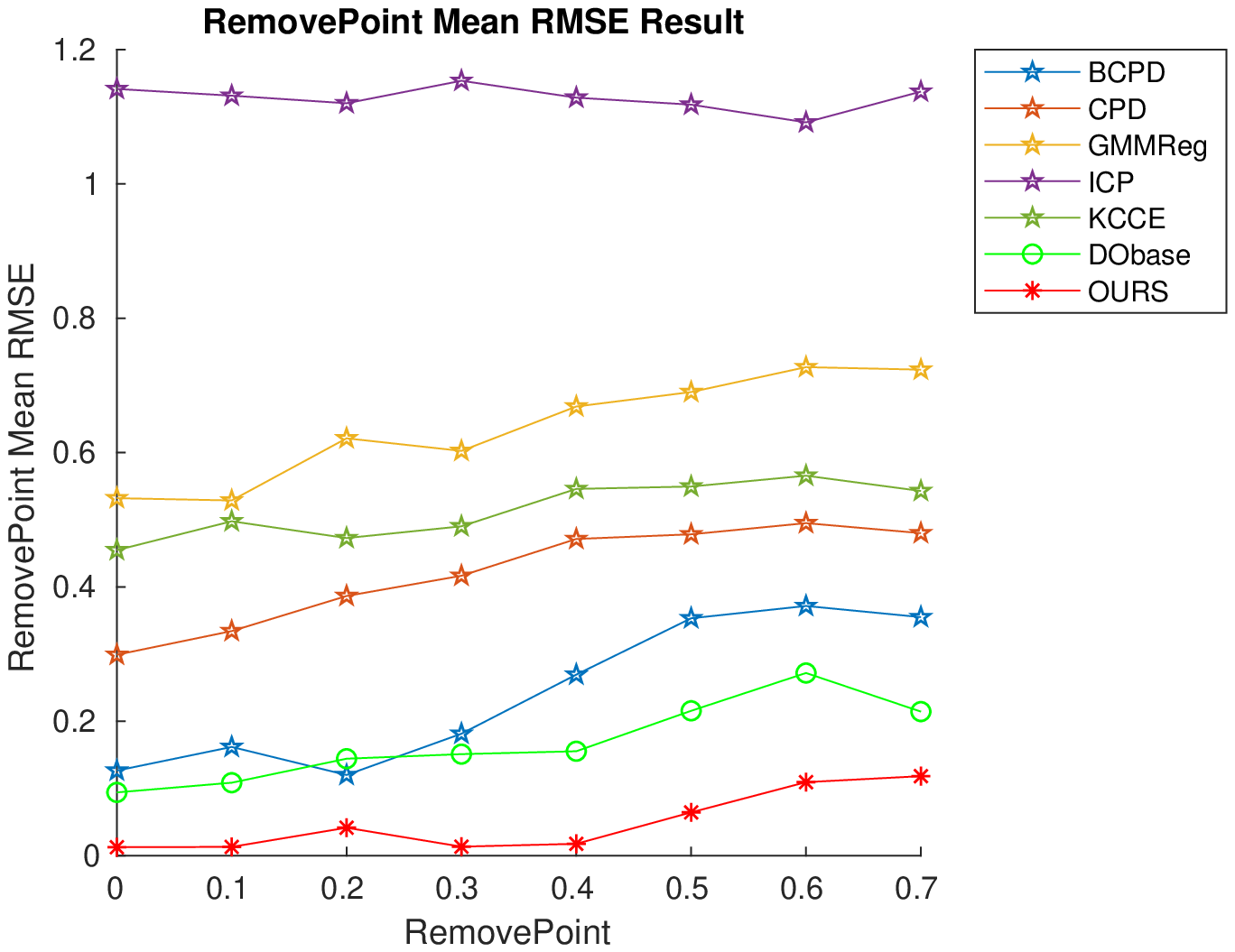}}
	\label{subfig:d2_t_RemovePoint_RMSE}

	\subfloat[Rotation RMSE]{\includegraphics[width=0.45\linewidth]{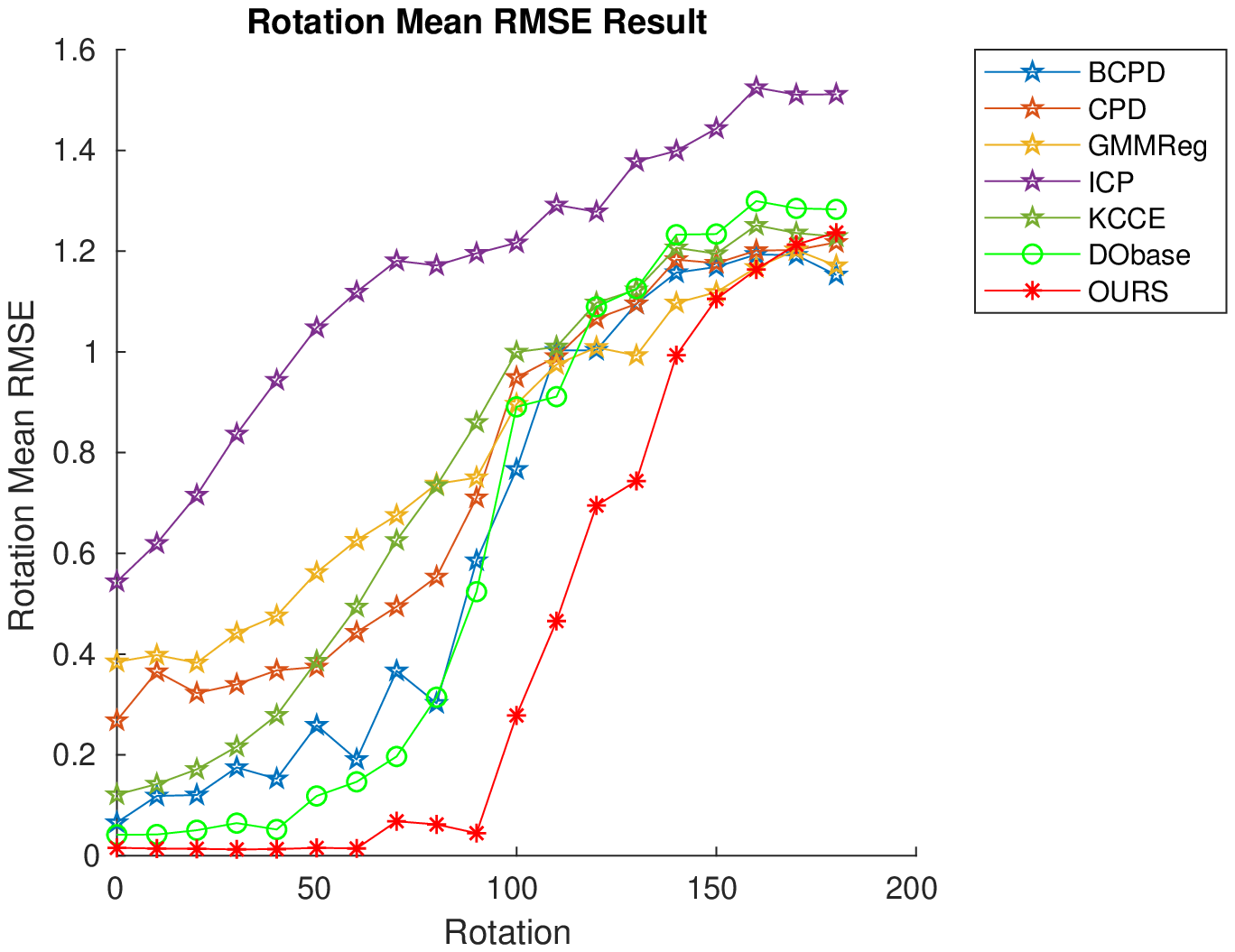}}
	\label{subfig:d2_t_Rotation_RMSE}
	\hspace{1em}
	\subfloat[Translation RMSE]{\includegraphics[width=0.45\linewidth]{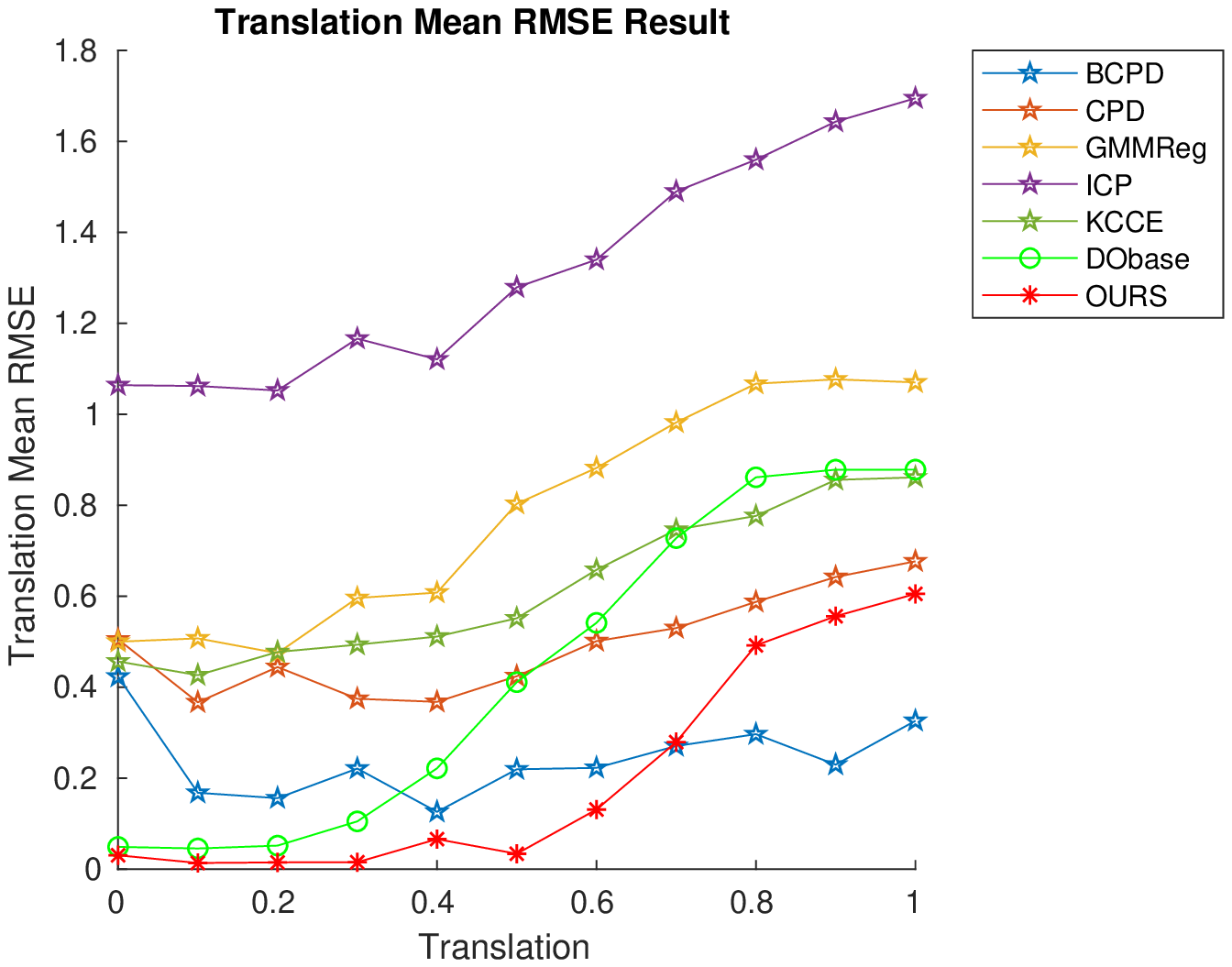}}
	\label{subfig:d2_t_Translation_RMSE}
	
	\caption{Mean RMSE Experimental Result on Dataset-2.}
	\label{fig:dat2-t-RMSE-result}
\end{figure}

We tested our improved Disciminative Optimization algorithm on the 3D LIDAR Oxford SensatUrban dataset, and the experimental results of all algorithm are ploted in Fig.\ref{fig:dat2-t-Acc-result} and Fig.\ref{fig:dat2-t-RMSE-result}, respectively. Differ from the Stanford Bunny dataset which forms a closed surface, the Oxford SensatUrban Dataset is an urban-scale LIDAR dataset which is an un-closed surface. The original DO performs well in the point cloud of closed surface. But for the unclosed surface dataset, DO's performance becomes close to that of the BCPD in the NoiseSTD, Outliers, PointNum, and Occalusion perturbation experiments. In the Rotation and Translation perturbation experiments, the BCPD's performance better than that of DO. That means the original DO is sensetive to the shape of points. The ICP, GMMReg, KCCE, and CPD still has a low point registration accuracy and high point registration RMSE in most of cases. The structured outliers generated a density mixture Gaussain which influenced the Gaussian-based approaches serverely. So that the performance of CPD, GMMReg, and KCCE are declined fastly. Our improved DO is robust to the shape of points, and we still achieved the best performance in almost all testing cases. The Table.\ref{tab:data2-ACC} and Table.\ref{tab:data2-RMSE} listed the experimental results of average PointAcc and average PointRMSE on real dataset, which demonstrated that our algorithm outperformed the other algorithms.

\section{Conclusion}
\textcolor{black}{In this paper, we proposed an improved Discriminative Optimization against the orginal DO. By extending the relationship between point cloud from "front-back" to "front-back", "up-down", and "clockwise-anticlockwise", the information of histogram was enrich, so that it is of benefit to the linear regression in the improved discriminative optimization algorithm. We compared our improved DO with other classical point cloud registration algorithms on two LIDAR dataset. The experimental results demonstrate that our improved DO performs well against The-State-Of-Art methods. Especially, it outperform the original DO in all cases.}

\ifCLASSOPTIONcaptionsoff
  \newpage
\fi

\bibliographystyle{IEEEtran}
\bibliography{refbib}

\begin{thebibliography}{10}
\providecommand{\url}[1]{#1}
\csname url@samestyle\endcsname
\providecommand{\newblock}{\relax}
\providecommand{\bibinfo}[2]{#2}
\providecommand{\BIBentrySTDinterwordspacing}{\spaceskip=0pt\relax}
\providecommand{\BIBentryALTinterwordstretchfactor}{4}
\providecommand{\BIBentryALTinterwordspacing}{\spaceskip=\fontdimen2\font plus
\BIBentryALTinterwordstretchfactor\fontdimen3\font minus
  \fontdimen4\font\relax}
\providecommand{\BIBforeignlanguage}[2]{{%
\expandafter\ifx\csname l@#1\endcsname\relax
\typeout{** WARNING: IEEEtran.bst: No hyphenation pattern has been}%
\typeout{** loaded for the language `#1'. Using the pattern for}%
\typeout{** the default language instead.}%
\else
\language=\csname l@#1\endcsname
\fi
#2}}
\providecommand{\BIBdecl}{\relax}
\BIBdecl

\bibitem{P2010Cascaded}
P.~Dollár, P.~Welinder, and P.~Perona, ``Cascaded pose regression,''
  \emph{Proceedings of the IEEE conference on computer vision and pattern
  recognition}, 2010.

\bibitem{vongkulbhisal2017discriminative}
J.~Vongkulbhisal, F.~De~la Torre, and J.~P. Costeira, ``Discriminative
  optimization: Theory and applications to point cloud registration,'' in
  \emph{Proceedings of the IEEE conference on computer vision and pattern
  recognition}, 2017, pp. 4104--4112.

\bibitem{2013Superviseddescentmethod}
D.~l. T.~F. Xiong~X, ``Supervised descent method and its applications to face
  alignment,'' in \emph{Proceedings of the IEEE conference on computer vision
  and pattern recognition}, 2013, pp. 532--539.

\bibitem{2020guo3Dclouds}
Y.~Guo, H.~Wang, Q.~Hu, H.~Liu, and M.~Bennamoun, ``Deep learning for 3d point
  clouds: A survey,'' \emph{Transactions on Pattern Analysis and Machine
  Intelligence}, vol.~PP, no.~99, pp. 1--1, 2020.

\bibitem{1992ICP}
P.~J. Besl and H.~Mckay, ``A method for registration of 3-d shapes,''
  \emph{IEEE Transactions on Pattern Analysis and Machine Intelligence},
  vol.~14, no.~2, pp. 239--256, 1992.

\bibitem{rusinkiewicz2001ICP}
S.~Rusinkiewicz and M.~Levoy, ``Efficient variants of the icp algorithm,''
  \emph{Proceedings Third International Conference on 3-D Digital Imaging and
  Modeling}, 2001.

\bibitem{fitzgibbon2003robust}
A.~W. Fitzgibbon, ``Robust registration of 2d and 3d point sets,'' \emph{Image
  and vision computing}, vol.~21, no. 13-14, pp. 1145--1153, 2003.

\bibitem{2010CPD}
A.~Myronenko and X.~Song, ``Point set registration: Coherent point drift,''
  \emph{IEEE Transactions on Pattern Analysis and Machine Intelligence},
  vol.~32, no.~12, pp. 2262--2275, 2010.

\bibitem{2004KC_kernel}
Y.~Tsin and T.~Kanade, ``A correlation-based approach to robust point set
  registration,'' in \emph{European Conference on Computer Vision}, 2004.

\bibitem{RPM1998}
S.~Gold, A.~Rangarajan, C.~P. Lu, S.~Pappu, and E.~Mjolsness, ``New algorithms
  for 2d and 3d point matching: pose estimation and correspondence,''
  \emph{Pattern recognition}, vol.~31, no.~8, pp. 1019--1031, 1998.

\bibitem{bergstrom2014IRLS}
P.~Bergstr{\"o}m and O.~Edlund, ``Robust registration of point sets using
  iteratively reweighted least squares,'' \emph{Computational optimization and
  applications}, vol.~58, no.~3, pp. 543--561, 2014.

\bibitem{GMMReg}
J.~B. and V.~B. C., ``Robust point set registration using gaussian mixture
  models,'' \emph{IEEE transactions on pattern analysis and machine
  intelligence}, vol.~33, no.~8, pp. 1633--1645, 2010.

\bibitem{2016supportVecotrRegress}
D.~Campbell and L.~Petersson, ``An adaptive data representation for robust
  point-set registration and merging,'' in \emph{IEEE International Conference
  on Computer Vision}, 2016.

\bibitem{2016Gravitational}
V.~Golyanik, S.~Aziz, and D.~Stricker, ``Gravitational approach for point set
  registration,'' in \emph{International Conference on Computer Vision and
  Pattern Recognition}, 2016.

\bibitem{1998AAM}
T.~F. Cootes, G.~J. Edwards, and C.~J. Taylor, ``Active appearance models,'' in
  \emph{Proceedings of the 5th European Conference on Computer Vision}, 1998.

\bibitem{HyperplaneTempMatch}
F.~Jurie and M.~Dhome., ``Boosted regression active shape models,'' \emph{IEEE
  transactions on pattern analysis and machine intelligence}, vol.~24, no.~7,
  p. 996–1000, 2002.

\bibitem{2007GentleBoost}
D.~Cristinacce and T.~F. Cootes., ``Active appearance models,'' in
  \emph{Proceedings of British Machine Vision Conf.}, 2007.

\bibitem{2004Lietemptracking}
E.~Bayro-Corrochano and J.~Ortegon-Aguilar, ``Lie algebra template tracking,''
  in \emph{International Conference on Pattern Recognition}, 2004.

\bibitem{2007Lie3Dtracking}
E.~B. Corrochano and J.~O. Aguilar, ``Lie algebra approach for tracking and 3d
  motion estimation using monocular vision,'' \emph{Image and Vision
  Computing}, vol.~25, no.~6, pp. 907--921, 2007.

\bibitem{2008HOGfeatureLie}
O.~Tuzel, F.~Porikli, and P.~Meer., ``Learning on lie groups for invariant
  detection and tracking,'' in \emph{International Conference on Computer
  Vision and Pattern Recognition}, 2008.

\bibitem{2006IEBM}
J.~Saragih and R.~Goecke, ``Iterative error bound minimisation for aam
  alignment,'' in \emph{International Conference on Pattern Recognition},
  vol.~2.\hskip 1em plus 0.5em minus 0.4em\relax IEEE, 2006, pp. 1196--1195.

\bibitem{2002Vsupport}
C.~C. Chang and C.~J. Lin, ``Training v-support vector regression: Theory and
  algorithms,'' \emph{Neural Computation}, vol.~14, no.~8, pp. p.1959--1977,
  2002.

\bibitem{cao2014faceboost}
X.~Cao, Y.~Wei, F.~Wen, and J.~Sun, ``Face alignment by explicit shape
  regression,'' \emph{International Journal of Computer Vision}, vol. 107,
  no.~2, pp. 177--190, 2014.

\bibitem{2013DeepCascadefacial}
S.~Yi, X.~Wang, and X.~Tang, ``Deep convolutional network cascade for facial
  point detection,'' in \emph{International Conference on Computer Vision and
  Pattern Recognition}.\hskip 1em plus 0.5em minus 0.4em\relax IEEE, 2013.

\bibitem{xiong2015global}
X.~Xiong and F.~De~la Torre, ``Global supervised descent method,'' in
  \emph{Proceedings of the IEEE Conference on Computer Vision and Pattern
  Recognition}, 2015, pp. 2664--2673.

\bibitem{2009OptimalSequence}
K.~Zimmermann, J.~Matas, and T.~Svoboda, ``Tracking by an optimal sequence of
  linear predictors,'' \emph{Transactions on Pattern Analysis and Machine
  Intelligence}, vol.~31, no.~4, pp. 677--692, 2009.

\bibitem{Qi2017PointNet}
C.~R. Qi, L.~Yi, H.~Su, and L.~J. Guibas, ``Pointnet: Deep learning on point
  sets for 3d classification and segmentation,'' in \emph{Conference on
  Computer Vision and Pattern Recognition}, 2017, pp. 652--660.

\bibitem{Qi2017PointNetPP}
------, ``Pointnet++: Deep hierarchical feature learning on point sets in a
  metric space,'' in \emph{Conference on Neural Information Processing
  Systems}, 2017, pp. 5105--5114.

\bibitem{aoki2019pointnetlk}
Y.~Aoki, H.~Goforth, R.~A. Srivatsan, and S.~Lucey, ``Pointnetlk: Robust and
  efficient point cloud registration using pointnet,'' in \emph{Conference on
  Computer Vision and Pattern Recognition}, 2019, pp. 7163--7172.

\bibitem{huang2020feature}
X.~Huang, G.~Mei, and J.~Zhang, ``Feature-metric registration: A fast
  semi-supervised approach for robust point cloud registration without
  correspondences,'' in \emph{Conference on Computer Vision and Pattern
  Recognition}, 2020, pp. 11\,366--11\,374.

\bibitem{ao2020SpinNet}
S.~Ao, Q.~Hu, B.~Yang, A.~Markham, and Y.~Guo, ``Spinnet: Learning a general
  surface descriptor for 3d point cloud registration,'' in \emph{Proceedings of
  the IEEE/CVF Conference on Computer Vision and Pattern Recognition}, 2021.

\bibitem{choy2020deep}
C.~Choy, W.~Dong, and V.~Koltun, ``Deep global registration,'' in
  \emph{Proceedings of the IEEE/CVF Conference on Computer Vision and Pattern
  Recognition}, 2020, pp. 2514--2523.

\bibitem{gower1975generalized}
J.~C. Gower, ``Generalized procrustes analysis,'' \emph{Psychometrika},
  vol.~40, no.~1, pp. 33--51, 1975.

\bibitem{elbaz20173d}
G.~Elbaz, T.~Avraham, and A.~Fischer, ``3d point cloud registration for
  localization using a deep neural network auto-encoder,'' in \emph{Proceedings
  of the IEEE/CVF Conference on Computer Vision and Pattern Recognition}, 2017,
  pp. 4631--4640.

\bibitem{gojcic2020learning}
Z.~Gojcic, C.~Zhou, J.~D. Wegner, L.~J. Guibas, and T.~Birdal, ``Learning
  multiview 3d point cloud registration,'' in \emph{Proceedings of the IEEE/CVF
  Conference on Computer Vision and Pattern Recognition}, 2020, pp. 1759--1769.

\bibitem{lu2019deepvcp}
W.~Lu, G.~Wan, Y.~Zhou, X.~Fu, P.~Yuan, and S.~Song, ``Deepvcp: An end-to-end
  deep neural network for point cloud registration,'' in \emph{Proceedings of
  the IEEE/CVF International Conference on Computer Vision}, 2019, pp. 12--21.

\bibitem{2015LiegroupAlgebras}
B.~Hall, ``Lie groups, lie algebras, and representations: an elementary
  introduction,'' \emph{Springer}, vol. 222, 2015.

\bibitem{KCCE2011}
P.~Chen, ``A novel kernel correlation model with the correspondence
  estimation,'' \emph{Journal of Mathematical Imaging and Vision}, vol.~39,
  no.~82, pp. 100--120, 2011.

\bibitem{HiroseBCPD}
O.~Hirose, ``A bayesian formulation of coherent point drift,''
  \emph{Transactions on Pattern Analysis and Machine Intelligence}, 2020.

\bibitem{stanfordbunny}
S.~U. C.~G. Laboratory, ``The stanford 3d scanning repository,''
  \url{https://graphics.stanford.edu/data/3Dscanrep/}.

\bibitem{Qingyong:2021dataset}
Q.~Hu, B.~Yang, and Khalid, ``Towards semantic segmentation of urban-scale 3d
  point clouds: A dataset, benchmarks and challenges,'' \emph{Proceedings of
  the IEEE/CVF Conference on Computer Vision and Pattern Recognition}, 2021.

\end{thebibliography}

%
%
%
%
%

\end{document}